\documentclass{article}

\usepackage[preprint, nonatbib]{neurips_2025}

\usepackage[utf8]{inputenc} 
\usepackage[T1]{fontenc}    
\usepackage[hidelinks]{hyperref}       
\usepackage{url}            
\usepackage{booktabs}       
\usepackage{amsfonts}       
\usepackage{nicefrac}       
\usepackage{microtype}      
\usepackage{xcolor}         

\usepackage{algorithm}
\usepackage{algpseudocode}
\usepackage{multirow}
\usepackage{caption}
\usepackage{subcaption}
\usepackage{float}
\usepackage{soul, color}
\usepackage{eqparbox}
\usepackage{graphicx}
\usepackage{amsmath}

\algnewcommand{\LeftComment}[1]{\hfill\eqparbox{COMMENT}{\(\triangleright\) #1}}

\AtBeginDocument{%
  }

\algdef{SE}
[METHOD]
{Method}
{EndMethod}
[2]
{\textbf{method} \textsc{#1} (\textsc{#2})}
{\textbf{end method}}

\title{LSH-DynED: A Dynamic Ensemble Framework with LSH-Based Undersampling for Evolving Multi-Class Imbalanced Classification}
\runningtitle{LSH-DynED}

\author{%
  \href{https://orcid.org/0000-0002-2980-4251}{Soheil Abadifard} \thanks{This research was primarily conducted while the author was affiliated with Bilkent University.} \\
  Department of Computer Science\\
  Kansas State University\\
  Manhattan, KS, USA \\
  \texttt{abadifard@k-state.edu} \\
  \And
   \href{https://orcid.org/0000-0003-0016-4278}{Fazli Can} \\
   Department of Computer Engineering \\
   Bilkent University\\
   Ankara, Turkey \\
  \texttt{canf@cs.bilkent.edu.tr} \\
}

\begin{document}

\maketitle

\begin{abstract}
The classification of imbalanced data streams, which have unequal class distributions, is a key difficulty in machine learning, especially when dealing with multiple classes. While binary imbalanced data stream classification tasks have received considerable attention, only a few studies have focused on multi-class imbalanced data streams. Effectively managing the dynamic imbalance ratio is a key challenge in this domain. This study introduces a novel, robust, and resilient approach to address these challenges by integrating Locality Sensitive Hashing with Random Hyperplane Projections (LSH-RHP) into the Dynamic Ensemble Diversification (DynED) framework. To the best of our knowledge, we present the first application of LSH-RHP for undersampling in the context of imbalanced non-stationary data streams. The proposed method undersamples the majority classes by utilizing LSH-RHP, provides a balanced training set, and improves the ensemble's prediction performance. We conduct comprehensive experiments on 23 real-world and ten semi-synthetic datasets and compare LSH-DynED with 15 state-of-the-art methods. The results reveal that LSH-DynED outperforms other approaches in terms of both Kappa and mG-Mean effectiveness measures, demonstrating its capability in dealing with multi-class imbalanced non-stationary data streams. Notably, LSH-DynED performs well in large-scale, high-dimensional datasets with considerable class imbalances and demonstrates adaptation and robustness in real-world circumstances. To motivate our design, we review existing methods for imbalanced data streams, outline key challenges, and offer guidance for future work.\footnote{For the reproducibility of our results, we have made our implementation available on \href{https://github.com/soheilabadifard/LSH-DynED}{GitHub}.}
\end{abstract}

\textit{\textbf{keywords}}  Data stream, Classification, Class imbalance, Ensemble learning, Locality sensitive hashing

\section{Introduction}\label{sec: introduction}
The volume of data generated by various sources has grown explosively, resulting in continuous high-velocity streams that require real-time processing and analysis \cite{read2023learning}. Data streams are characterized by their continuous flow, unbounded nature, high velocity, one-pass processing, and dynamic changes in content over time. These properties are different from static data, which has all data available while processing. This trend poses new hurdles for traditional machine learning techniques designed initially to handle static data. In the context of data streams, data samples are processed incrementally using methods appropriate for real-time or nearly real-time analysis \cite{jinjie2025noise}, which is crucial in industries such as banking, healthcare, and transportation. Additionally, some of this data may also come from sensors, social media feeds, or Internet of Things devices, all of which produce big data streams since they produce large amounts of data not collected based on predetermined data characteristics. For instance, IoT devices are expected to grow from 26 billion connected units in 2019 to nearly 80 billion by 2025 \cite{bahri2021data}. This necessitates using specialized algorithms that can manage the high throughput and adjust to the dynamic nature of the data stream. Consequently, this task has become a focus of research, developing algorithms capable of performing real-time tasks such as incremental clustering and classification. Current research in unsupervised and supervised learning within data streams explores methods like density-micro clustering and classification algorithms \cite{kokate2018data}, respectively. The study reported in \cite{read2011streaming} focuses on multi-class data streams where each data sample is associated with a single label, as opposed to multi-label data streams, in which each data sample is linked to a combination of labels \cite{berkay2022multi}. 

The core challenge addressed in this paper is the classification of non-stationary, multi-class, imbalanced data streams, particularly those characterized by dynamically fluctuating imbalance ratios in the presence of concept drift. Addressing these combined complexities is crucial for developing robust, resilient, and adaptive learning systems. The imbalance ratio is defined as the ratio between the number of instances in the most frequent class and that in the least frequent class. In the rest of this section, we first introduce imbalanced data streams and concept drift, then outline the motivations and contributions of this study.

\textit{\textbf{Imbalanced data streams and concept drift:}} One of the most difficult challenges in the data stream classification task is effectively handling class imbalance in data. \textit{Imbalanced data streams} refer to situations when there is a significant difference in the number of instances between various classes \cite{aguiar2023survey, lipska2022influence}. These streams exhibit a notable difference in the representation of different classes, with some classes being referred to as minority classes and others as majority classes. The difference in the distribution of classes poses significant difficulties for any machine learning algorithm, as they may display a bias towards the dominant class and often overlook the less common but crucial information included within the minority class \cite{li2020incremental}. This difference might result in poor predictive accuracy, particularly when identifying instances from the minority classes. On the other hand, the ever-changing characteristics of data streams result in a fluctuating imbalance ratio, which can reverse the roles of majority and minority classes as the stream progresses.

In addition to the challenge of imbalance, another significant obstacle in machine learning is dealing with how data streams constantly change, a phenomenon called concept drift \cite{gozuaccik2021concept}. As data distributions change over time, it is crucial to have a learning system that can adapt. Concept drift happens when the model's statistical properties undergo an unexpected change in the target variable. This can significantly lower the accuracy of models, making it hard to keep their performance high. To handle concept drift effectively, one needs methods to detect it and adjust the learning process, such as updating the model with new data, using algorithms to detect drift, or resetting the model periodically \cite{bakhshi2023broad}. 
The interplay of concept drift with class imbalance, especially in multi-class scenarios with dynamic ratios and overlapping class boundaries, further generates complicated classification tasks \cite{aguiar2023survey}. A recently proposed method, DynED \cite{abadifard2023dyned}, handles the concept drift challenge by introducing diversity among the components of an ensemble, as a greater diversity of ensemble components is known to enhance prediction accuracy in data stream classification tasks \cite{bonab2019less, bonab2018goowe, juan2021dyn}. 

Therefore, a combination of concept drift and class imbalance introduces complex dynamics where class roles may shift, new classes may emerge, or existing classes might disappear, requiring continual adaptation of the classifiers. Handling these challenges effectively demands specialized techniques and algorithms, such as ensemble methods, feature selection, resampling techniques, and cost-sensitive learning. These approaches either alter the underlying dataset to balance the class distribution or modify the training approaches to make classifiers robust against these imbalances \cite{aguiar2023survey}. Hence, effectively handling an imbalanced data stream necessitates employing advanced tactics to guarantee equitable and precise model performance across all classes while acknowledging the dynamic and intricate characteristics of the data at hand.

Recently, the integration of resampling techniques with online ensemble learning to handle imbalanced data streams has attracted significant interest. Resampling is a typical technique for dealing with class imbalance. This technique involves either increasing instances in the minority class (oversampling) or decreasing instances in the majority class (undersampling) \cite{aguiar2023survey} to manage imbalanced data distribution. Although numerous sampling algorithms have been created so far, many of these methods are influenced by the traditional SMOTE algorithm \cite{zhu2017synthetic}. It lowers class imbalance by employing linear interpolation to generate new instances of the minority class from existing minority class samples. Although this technique effectively reduces class imbalance, it may provide noisy and false instances that do not truly belong to the minority class, potentially leading to over-generalizing the minority class into the majority class region and lowering the overall prediction performance \cite{tarawneh2022stop}.  

The imbalance issue in the data stream classification task can also be addressed by utilizing ensemble classifiers \cite{gomes2017survey, Wang_Minku_Yao_2015}, which combine several weak classifiers to form a stronger one. By combining multiple classifiers, the ensemble utilizes the advantage of each one's unique capabilities to achieve a better performance. Furthermore, resampling can be used in these techniques to balance class distribution during training, provide equal representation of all classes, and enhance overall accuracy and minority class handling \cite{li2020incremental}.

\textit{\textbf{Motivations and Contributions: }}The primary motivation behind this research is that the classification of imbalanced binary data streams has been extensively studied; however, multi-class imbalanced data streams have received less attention and remain an open research challenge, according to recent surveys and research \cite{bernardo2020c, gulowaty2019smote, Palli_Jaafar_Gilal_Alsughayyir_Gomes_Alshanqiti_Omar_2024, wang2018systematic}. While some existing approaches address multi-class scenarios, they often exhibit limitations in areas such as adaptability to rapid distributional changes, or the representativeness of sampling in highly dynamic and skewed environments, issues detailed further in Section \ref{sec: Related_Works}. Techniques such as VFC-SMOTE and RebalanceStream have addressed binary class imbalances by adapting to evolving data streams, and concept drifts \cite{bernardo2021vfc, bernardo2020incremental}. Additionally, SMOTE-OB has demonstrated the effectiveness of combining synthetic minority oversampling with online bagging to continuously rebalance class distributions in evolving streams \cite{bernardo2021smote}. The Adaptive Chunk-Based Dynamic Weighted Majority (ACDWM) method further enriches this landscape by using an ensemble of classifiers dynamically weighted based on their recent error rates to handle both concept drift and class imbalance, demonstrating significant advancements in the field of online learning with imbalanced data streams \cite{lu2019adaptive}.

These binary classification advancements underline the urgent need for robust solutions in multi-class settings, particularly when faced with non-stationary environments where class distributions and relationships dynamically change. Furthermore, the problem becomes more complicated when class proportions fluctuate dynamically \cite{aguiar2023survey} and may even reverse, particularly when concept drift is involved. These challenges can have a negative influence on classification model performance. These necessitate a more robust and resilient method toward the above-mentioned challenges \cite{vardi2020efficiency}. While resilience is about recovering and adapting after disruptions, robustness is about withstanding and resisting them \cite{wang2012resilience}.  

This study presents a novel approach to address the issues of multi-class imbalance in non-stationary data streams and aims to introduce a robust and resilient approach. Our contributions are detailed below. We
\begin{itemize}\label{sec: contributions}
    \item Propose the first application of LSH-RHP \cite{liu2016multilinear} for undersampling in the context of multi-class imbalanced non-stationary data stream classification;
    \item Extend and modify the DynED \cite{abadifard2023dyned}, our recently introduced data stream classification method, to also overcome multi-class imbalanced data stream tasks, especially those with dynamic imbalance ratios;
    \item Present comprehensive experiments comparing approaches proposed for this task over the last decade on multiple real-world and semi-synthetic datasets and show that our method outperforms other approaches in terms of both Kappa and mG-Mean effectiveness measures;
    \item Provide our implementation with all experimental details to make our approach open to new improvements in future research and available as a baseline;
    \item Offer a detailed review of current methods for imbalanced data stream classification, highlighting key challenges and trends relevant to our proposed method.
\end{itemize}

While our main technical novelty lies in managing class imbalance via LSH-RHP, the backbone of our approach, DynED, integrates explicit concept drift detection through ADWIN and dynamically adapts the ensemble, ensuring responsiveness to evolving distributions over time \cite{abadifard2023dyned, al2024incremental, barboza2025inca}.

This study is organized as follows: in section \ref{sec: Related_Works}, we provide an in-depth review of relevant literature; in section \ref{sec: Proposed Approach LSH-DynED}, we present our proposed method; and in section \ref{sec: Experimental Evaluation}, we present our experiments and provide a complete and comprehensive discussion of the findings. Lastly, we conclude with Section \ref{sec: Conclusion}.

\section{Related Works}\label{sec: Related_Works}

This section begins with a detailed review of current methods for imbalanced data stream classification \ref{sec: Stream Classification Methods}, followed by a short overview of sampling methods \ref{sec: sampling}.

\subsection{Data Stream Classification Methods for Imbalanced Data Streams}\label{sec: Stream Classification Methods}

Data stream classification methods can be categorized into two primary groups: General-Purpose Methods (GPM) such as \textbf{OBA} \cite{bifet2009new}, \textbf{LB} \cite{bifet2010leveraging}, \textbf{ARF} \cite{gomes2017adaptive}, \textbf{BELS} \cite{bakhshi2023broad}, \textbf{DynED} \cite{abadifard2023dyned}, \textbf{KUE} \cite{cano2020kappa}, \textbf{SRP} \cite{gomes2019streaming}, and Imbalance-Specific Methods (ISM). This section analyzes several methods developed only for classifying imbalanced data streams, using various strategies, and presents their strengths and weaknesses. The summary of various approaches can be seen in Table \ref{table: methods}. We explain them in the following.

The \textbf{Hellinger Distance Very Fast Decision Tree} (\textbf{HD-VFDT}) method \cite{cieslak2008learning} is specifically developed to tackle the issue of learning from imbalanced data streams, a prevalent difficulty in various supervised learning applications. In such cases, conventional algorithms favor larger and more dominant classes. This paper \cite{rao1995review} presents and assesses the Hellinger distance as a new, robust splitting criterion for decision tree learning unaffected by skewness. The Hellinger distance is a reliable metric for quantifying the divergence between feature value distributions. Unlike traditional criteria such as information gain \cite{quinlan1986induction} and the Gini index \cite{breiman2017classification}, it is unaffected by any skewness in the class distribution. This technique aims to enhance the precision and equity of decision tree models, offering a resilient solution to the difficulties presented by imbalanced data streams.

The HD-VFDT algorithm greatly enhances the management of imbalanced data streams, namely by improving the recall rates for the minority class. However, it has difficulties regarding memory efficiency and computing performance while handling large-scale data streams. These constraints can impede its practical implementation in real-time applications where fast processing is crucial. The \textbf{Gaussian Hellinger Very Fast Decision Tree} (\textbf{GH-VFDT}) \cite{lyon2014hellinger} addresses restrictions by integrating an enhanced variation of the Hellinger distance, which measures the dissimilarity between normal distributions, resulting in better computational efficiency. The GH-VFDT is ideal for high-speed data streams in which memory consumption and processing performance are critical. The GH-VFDT, similar to its previous version, prioritizes enhancing recall rates for the minority class while minimizing false positives. The algorithm successfully balances sensitivity towards the minority class and overall accuracy by utilizing statistical metrics of data distribution to make educated decisions at each node in the tree. Furthermore, the GH-VFDT algorithm improves the management of imbalanced class distributions and consistently achieves high G-Mean values in diverse levels of class imbalance, demonstrating its resilience in various streaming settings. This enhancement in memory usage does not undermine the classifier's capacity to handle the difficulties presented by imbalanced learning in data streams.
\begin{table}
\centering
\caption{\textbf{Summary of data stream methods used in this study for the classification of data streams.} (Related to Sec. \ref{sec: Related_Works}, and Sec. \ref{sec: Experimental Evaluation}.)}\label{table: methods}
\resizebox{1\textwidth}{!}{
\begin{tabular}{lrllll} 
\toprule
     Method &  Proposed Year & Domain &                   Imbalance Handling Approach & Learning Method & Concept Drift \\
\midrule
        OBA \cite{bifet2009new}        &        2009 &     GPM &                                               None &          Online &      Explicit \\
        LB \cite{bifet2010leveraging}  &        2010 &     GPM &                                               None &          Online &      Explicit \\
        ARF \cite{gomes2017adaptive}   &        2017 &     GPM &                                               None &          Online &      Explicit \\
        SRP \cite{gomes2019streaming}  &        2019 &     GPM &                                               None &          Online &      Explicit \\
        KUE \cite{cano2020kappa}       &        2020 &     GPM &                                               None &           Chunk &      Implicit \\
       BELS \cite{bakhshi2023broad}    &        2023 &     GPM &                                               None &           Chunk &      Implicit \\
      DynED \cite{abadifard2023dyned}  &        2023 &     GPM &                                               None &          Online &      Explicit \\
     HD-VFDT \cite{cieslak2008learning} &       2008 &     ISM &              Hellinger Distance Splitting Criteria &          Online &      Implicit \\
     GH-VFDT \cite{lyon2014hellinger}   &       2014 &     ISM &              Gaussian Hellinger Splitting Criteria &          Online &      Implicit \\
       MUOB \cite{wang2016dealing}     &        2016 &     ISM &                                      Undersampling &          Online &      Implicit \\
       MOOB \cite{wang2016dealing}     &        2016 &     ISM &                                       Oversampling &          Online &      Implicit \\
       ARFR \cite{ferreira2019adaptive}&        2019 &     ISM &                                         Resampling &          Online &      Explicit \\
      CSARF \cite{loezer2020cost}      &        2020 &     ISM &                             Misclassification Cost &          Online &      Explicit \\
     CALMID \cite{liu2021comprehensiveF}&       2021 &     ISM &              Sample Weight \& Uncertainty Strategy &          Online &      Explicit \\
       ROSE \cite{cano2022rose}        &        2022 &     ISM & Self-adjusting Bagging \& Per Class Sliding Window &          Online &      Explicit \\
    MicFoal \cite{liu2023multiclass}   &        2023 &     ISM & Sample Weight \& Uncertain Label Request Strategy  &          Hybrid &      Implicit \\
    LSH-DynED [This Study]             &        2024 &     ISM &                   Locality Sensitive Undersampling &          Online &      Explicit \\
\bottomrule
\end{tabular}}
\caption*{GPM: General-Purpose Method, ISM: Imbalance-Specific Method, Hybrid: Online $+$ Chunk, Explicit: Drift Detector used to Identify Concept Drifts, Implicit: Concept Drifts Identified by Method Itself. Implementation links for all these methods are available on the GitHub page associated with this paper.}
\end{table}

Previous methods in imbalanced streams, particularly for multi-class scenarios, have struggled with issues like dependency on initial samples for class decomposition and a lack of adaptivity to dynamic class distributions. This often leads to less effective handling of skewed class distributions in real-time online learning environments. Two new ensemble learning methods, \textbf{Multi-class Oversampling-based Online Bagging} (\textbf{MOOB}) and \textbf{Multi-class Undersampling-based Online Bagging} (\textbf{MUOB}) \cite{wang2016dealing}, are introduced to overcome these issues. Both approaches are intended to process multi-class data streams directly and adaptively without the use of class decomposition or initial samples. To successfully balance class distributions, they employ adaptive resampling techniques \cite{wang2016dealing}.

MOOB and MUOB differ in their approach to managing class imbalances: MOOB focuses on generating additional instances of the minority classes (oversampling), whereas MUOB reduces the presence of majority class instances (undersampling). This enables both strategies to maintain updated and balanced class representations while adjusting to changes in class distribution as they occur. They also provide a time-decayed class size metric that represents the current state of class imbalances, allowing for adaptive sampling rates. In comparison, MOOB performs better and more consistently across numerous dynamic circumstances, particularly in sustaining stronger recall for minority classes without affecting overall accuracy. This is most likely owing to its strategy of increasing the presence of minority classes rather than just diminishing the majority class, which can be vital in contexts where minority class samples are especially valuable or predictive of critical outcomes.

The \textbf{Adaptive Random Forest with Resampling} (\textbf{ARFR}) \cite{ferreira2019adaptive} is a method developed to address the constraints of imbalanced streams by improving upon the conventional ARF \cite{gomes2017adaptive} approach. The ARFR method presents a resampling strategy that modifies instances' weights during training according to the current class distribution. By utilizing adaptive resampling, the importance of the minority class is enhanced, resulting in improved visibility and the classifier's enhanced accuracy in predicting it. Each instance in the data stream is dynamically weighted, increasing the representation of minority classes and ensuring continual adaptation to changes in class distribution. Moreover, ARFR has strategies to manage concept drift, which is essential for preserving model relevance in dynamic situations. The approach also demonstrates enhanced computational efficiency by reducing CPU time compared to classical ARF. Efficiency is crucial for applications that necessitate real-time data processing and decision-making.

Although the ARF and ARFR adapt to data streams, they usually struggle to adequately handle class imbalances, particularly in circumstances where misclassification of minority classes could have serious consequences. ARF prioritizes responding to changes in data streams over expressly considering class imbalance. At the same time, ARFR tries to address imbalance issues by employing resampling algorithms that do not directly incorporate the cost of misclassification into the learning process.

The \textbf{Cost-sensitive Adaptive Random Forest} (\textbf{CSARF}) \cite{loezer2020cost} addresses these constraints by introducing misclassification costs into the learning process, with a specific focus on the difficulties posed by imbalanced streams. CSARF improves on the standard ARF framework \cite{gomes2017adaptive} by using the MCC (Matthews Correlation Coefficient) \cite{chicco2020advantages} to provide weight to trees. This approach offers a more refined evaluation of classifier performance, which is particularly advantageous in situations when there is an imbalance in the stream. Furthermore, CSARF has a sliding window method to consistently monitor changes in class distribution, guaranteeing that the training process is adaptable to the ever-changing nature of data streams and the inclusion of minority class samples.

Previous techniques, such as \textbf{KUE} \cite{cano2020kappa}, have made considerable progress in dealing with concept drift in data streams. However, they often struggle to handle non-stationary class imbalances effectively. KUE prioritizes the utilization of the Kappa statistic for weighting and selecting classifiers. Although this strategy is effective at adjusting to changes in data drift, it may not always provide the flexibility required to respond swiftly to sudden shifts in class distribution and imbalance. 

The \textbf{Robust Online Self-Adjusting Ensemble} (\textbf{ROSE}) method \cite{cano2022rose} is a novel online ensemble classifier designed to address imbalanced and drifting data streams, attempting to overcome these limitations. ROSE seeks to address the deficiencies of prior approaches by improving adaptation and resilience to changes in concepts and class distribution imbalances, all without human intervention. The system includes innovative technical improvements, such as variable-size random feature subspaces for each classifier, which increase diversity and resilience to drift. Furthermore, ROSE utilizes a background ensemble to facilitate rapid adaptation by monitoring and replacing underperforming classifiers based on their response to drift.

Furthermore, ROSE implements a distinctive self-adjusting bagging method that adapts its strategy according to the existing class distribution, specifically enhancing the representation and classification precision of underrepresented classes. This is aided by using per-class sliding windows, which store a current selection of cases for each class. This guarantees that the classifier is constantly updated with the most relevant data. ROSE has a collection of capabilities that allow it to handle fast changes in data streams properly. It is able to retain strong performance even when faced with limited availability of labels and significant imbalances between classes.

A new approach, \textbf{Comprehensive Active Learning Method for Multi-class Imbalanced Data Streams with Concept Drift} (\textbf{CALMID}) \cite{liu2021comprehensiveF}, has been developed to address the limitations mentioned earlier such as effectively managing concurrent concept drift and severe class imbalance, adapting to sudden distributional shifts, and reducing the need for fully labeled data streams. The CALMID method uses a resilient framework to improve the management of imbalances in multi-class scenarios while reducing concept drift. The sample weight formula used in this method considers the dynamic class distribution and the specific difficulties associated with each class. In addition, CALMID employs an asymmetric margin threshold matrix for its uncertainty strategy, enabling more accurate and adaptable management of classification uncertainties across various classes. In addition, CALMID implements a novel initialization procedure for newly introduced basic classifiers, guaranteeing their enhanced readiness to handle the dynamic data streams right from the beginning. This strategy enhances the resilience of the learning process against changes in data patterns and also maximizes performance within limited resources for labeling.

Although methods such as CALMID and ARFR are creative in tackling class imbalance and concept drift in data streams, they have limitations, especially in situations where the dynamics are very unpredictable. CALMID, for example, may not adequately adjust to situations when there is a sudden change in class dominance or the emergence of new classes. ARFR primarily addresses imbalanced data by utilizing resampling techniques. However, it may not effectively respond to quick changes in class features or the development of new classes. The \textbf{MicFoal} framework \cite{liu2023multiclass} is designed to overcome these limitations by offering a reliable solution for imbalanced stream classification that efficiently handles multiclass imbalance and concept drift. MicFoal is a system specifically designed to manage the ever-changing nature of data streams. It is capable of adapting to situations where previous classes may no longer exist, and new ones may form. MicFoal includes a configurable supervised learner and uses a hybrid label request method, which provides flexibility in the process of active learning. This technique is backed by an adaptive mechanism that modifies the budget allocated for label costs according to current performance evaluations, guaranteeing the most efficient application of resources.

In addition, MicFoal presents a new and uncertain approach to requesting labels, utilizing a changeable least confidence threshold vector. This strategy is particularly effective in dealing with the fluctuating number of classes over time. MicFoal can adapt to changes in the data stream, ensuring that it stays accurate and efficient. MicFoal has greater flexibility and adaptability than CALMID and ARFR. It allows for easy modification and tuning of its components to accommodate the complex nature of real-world data streams effectively.

\textbf{\textit{Motivational Observations.}} The literature review we presented highlights critical challenges and limitations associated with current data stream classification and sampling techniques. The methods, such as HD-VFDT \cite{cieslak2008learning}, GH-VFDT \cite{lyon2014hellinger}, and ARFR \cite{ferreira2019adaptive}, specifically address key issues like class imbalance and responsiveness to evolving data streams. They underscore practical limitations, including the inability to adapt to rapidly changing class distributions. Our proposed method addresses this challenge through dynamic undersampling and ensemble adaptation. Approaches like MOOB \cite{wang2016dealing} and MUOB \cite{wang2016dealing} emphasize adaptive sampling methods for multi-class imbalance but reveal limitations such as dependency solely on adaptive resampling, potentially leading to ineffective responses during sudden class distribution shifts.

These constraints motivated our development of a more robust undersampling technique using Locality Sensitive Hashing (LSH-RHP), aimed at achieving enhanced responsiveness and representativeness in selected data subsets. Similarly, ensemble strategies such as CSARF \cite{loezer2020cost} and ROSE \cite{cano2022rose} manage concept drift and imbalance but possess complexity and explicit ensemble weighting dependencies that may not sufficiently address the scalability required for real-time applications. Additionally, foundational ensemble methods like ARFR\cite{ferreira2019adaptive}, CSARF\cite{loezer2020cost}, and ROSE \cite{cano2022rose} establish essential strategies for dynamic and adaptive classification. Our LSH-DynED framework enhances ensemble methods by addressing their diversity-related limitations.

\subsection{Sampling Techniques}\label{sec: sampling}
Sampling techniques try to balance data samples before using them to train classifiers. Several sampling techniques have been proposed to either oversample or increase data samples in the minority class(es) or undersample to eliminate some data samples from the majority class(es).

\subsubsection{\textbf{Oversampling Techniques}}
The most straightforward oversampling technique randomly reproduces data items from the minority class. A more complex technique, \textbf{Synthetic Minority Oversampling Technique} (\textbf{SMOTE}) \cite{chawla2002smote}, aims to generate synthetic examples rather than simple replication of minority class instances by interpolating between the two samples, ensuring the minority class decision region is more generalized. \textbf{SMOTEBoost} \cite{chawla2003smoteboost} is proposed to further improve prediction for the minority class in imbalanced datasets by integrating SMOTE with the boosting algorithm to create synthetic minority class examples during each boosting iteration. \textbf{Borderline-SMOTE} \cite{han2005borderline} is proposed to improve the predictive performance for the minority class by only oversampling the minority class samples near the decision boundary. \textbf{Geometric SMOTE} \cite{douzas2017geometric} improves the oversampling by generating synthetic samples within a geometric region around each selected minority sample, allowing the region to be a hyper-sphere, hyper-spheroid, or line segment.

\subsubsection{\textbf{Undersampling Techniques}}

The simplest undersampling technique is to eliminate samples from the majority classes randomly. \textbf{One-Sided Selection} (\textbf{OSS}) \cite{kubat1997addressing}, a more complex technique, removes noisy and borderline majority class samples using Tomek links \cite{tomek1976two} to perform undersampling tasks. The \textbf{Neighborhood Cleaning Rule} (\textbf{NCL}) \cite{laurikkala2001improving} is a technique designed to enhance the identification of small and difficult-to-classify classes within imbalanced datasets. NCL emphasizes data cleaning over data reduction by incorporating Wilson’s \textbf{Edited Nearest Neighbor} (\textbf{ENN}) \cite{tomek1976experiment} rule to remove noisy and borderline instances from the majority class. \textbf{Cluster-based Undersampling} \cite{yen2009cluster} clusters the entire dataset and selectively undersamples the majority class based on cluster characteristics, ensuring a more representative subset of the data. \textbf{Topics Oriented Directed Undersampling} (\textbf{TODUS}) \cite{santhiappan2018novel} leverages topic modeling to estimate the latent data distribution and generate a balanced training set based on their estimated importance, directing the undersampling process to minimize information loss.

\textbf{\textit{Motivational Observations.}} The sampling methods such as SMOTE \cite{chawla2002smote} and various undersampling techniques like OSS \cite{kubat1997addressing}, NCL \cite{laurikkala2001improving}, TODUS \cite{santhiappan2018novel}, and Cluster-based Undersampling \cite{yen2009cluster} provided insights and foundations for our approach. While traditional techniques suffer from drawbacks like overfitting in oversampling or excessive information loss in random undersampling \cite{nguyen2012comparative, tarawneh2022stop}, the LSH-RHP method we introduce strategically avoids these issues through randomized yet targeted subset selection, ensuring both diversity and representativeness with minimal information loss. 

Thus, the relevance of the reviewed methods lies in identifying and clarifying these critical limitations—inefficient handling of rapid class distribution changes and restricted adaptability—directly motivating our proposed method LSH-DynED.

\section{Proposed Approach: LSH-DynED}\label{sec: Proposed Approach LSH-DynED}

Building upon the analysis in Section \ref{sec: Related_Works}, which highlighted critical limitations in current methods, such as ineffective handling of rapid class distribution changes and restricted adaptability; this section introduces our novel method, LSH-DynED. It is specifically designed to overcome the shortcomings of existing methods in addressing the challenges posed by multi-class imbalanced data in non-stationary data streams.

We begin with the motivation and expected benefits of our approach (Section \ref{sec: Benefits of the Approach}), followed by a summary of the original DynED framework (Section \ref{sec: dyned-overview}). We then introduce Locality Sensitive Hashing with Random Hyperplane Projections (LSH-RHP) (Section \ref{sec: Locality Sensitive Hashing with Random Projections}) and explain how it is integrated into DynED to address class imbalance ( Section \ref{sec: LSH-DynED}). Finally, we analyze the time complexity of the full method in Section \ref{sec: Time Complexity Analysis}.

\subsection{Motivations and Benefits}\label{sec: Benefits of the Approach}

This integration aims to develop a novel method to manage class imbalances and make it resilient to dynamic class ratios. By combining LSH-RHP with DynED, we can better manage the selection of ensemble components, introducing greater diversity while addressing the challenges of imbalanced, multi-class real-time data streams. Our approach has two main parts. First, it chooses dynamic components that strike a balance between accuracy and diversity (DynED). Second, it constantly updates the ensemble's training data to ensure that the components are trained on a balanced yet diverse set of data (LSH-RHP). 

The key feature of our approach is the targeted undersampling of the majority classes using the LSH-RHP technique. By dividing the data into hash-based partitions, we can target and reduce the representation of over-represented classes without losing critical information, resulting in a balanced training dataset. Additionally, repeatedly performing LSH-RHP provides a diverse set of samples from the majority classes, leading to higher component diversity. This targeted undersampling is critical for ensuring that minority classes are adequately represented and plays a crucial role in robustness to class imbalance challenges, resilience to dynamic imbalance ratios, and improving the ensemble's overall predictive accuracy.

\subsection{DynED Overview}\label{sec: dyned-overview}
The DynED \cite{abadifard2023dyned} is a dynamic ensemble method that we introduced in our recent research, designed to mitigate the challenges posed by concept drift in balanced data stream classification; its effectiveness of DynED on handling the concept drift in non-stationary data streams is studied in several research papers \cite{abadifard2023dyned, al2024incremental, barboza2025inca}. DynED operates in three stages: (1) Prediction and Training, (2) Drift Detection and Adaptation, and (3) Component Selection.

\noindent\textbf{Prediction and Training:} At this stage, DynED employs a subset of its ensemble components, known as the ``selected components," to predict the label of the incoming data sample. This prediction is made via a majority voting. Following the prediction, the ``selected components" are trained on the new data instance.

\noindent\textbf{Drift Detection and Adaptation:} This stage focuses on detecting and adapting to concept drift. The drift detector, ADWIN \cite{bifet2007learning}, is constantly updated with the system's correct/incorrect predictions. If a drift is detected, a new component is added and trained on the most recent data samples stored in a sliding window. This new component is then added to a pool of ``reserved components." If it is added or the number of processed samples exceeds the $\mathbf{\theta}$ threshold, then a parameter $\mathbf{\lambda}$, which controls the balance between diversity and accuracy in the ensemble, is dynamically adjusted based on the intensity of changes in the accuracy of the ensemble over time, and stage 3 is activated to update the ``selected components."  
\begin{figure}[t]
    \centering
    \resizebox{\textwidth}{!}{
    \begin{tabular}{c}
        \includegraphics{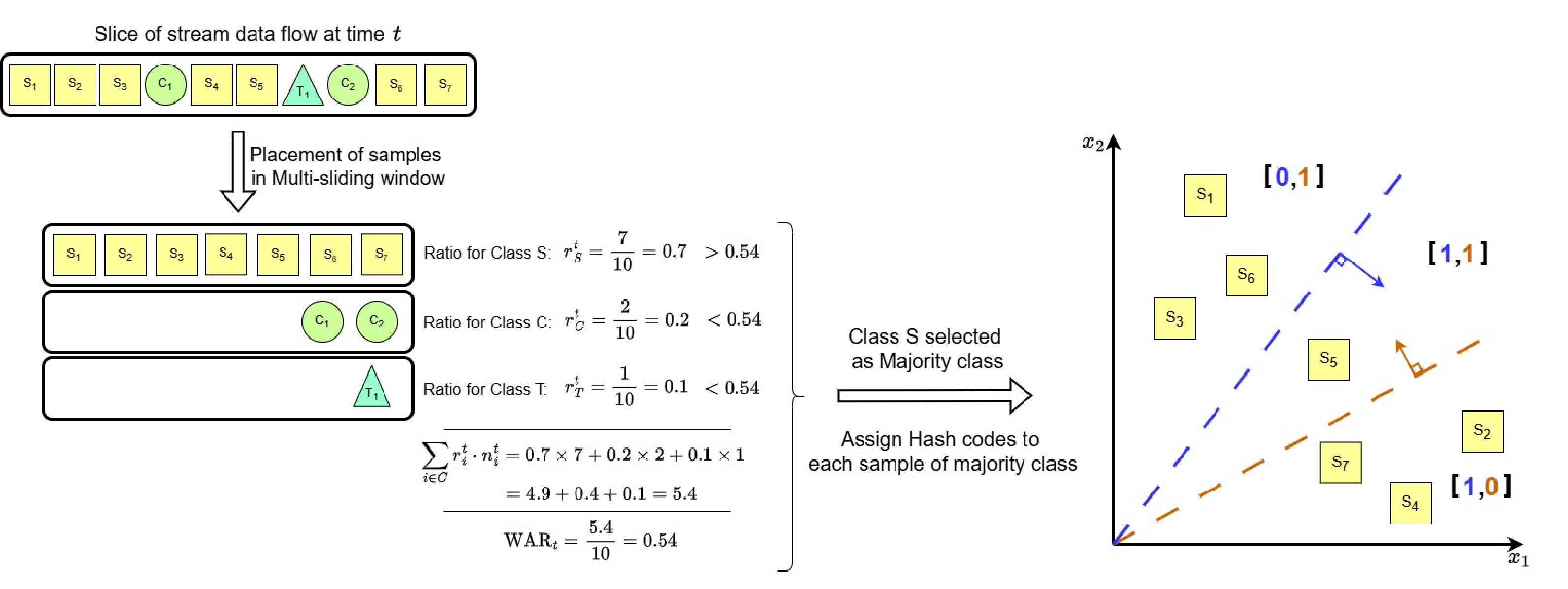}
    \end{tabular}}
    \caption{\textbf{Simple example of performing Locality Sensitive Hashing with Random Hyperplane Projection.} We assumed this stream contains two features, $x_1$ and $x_2$, and three classes: Square (S), Circle (C), and Triangle (T). The Square class is the majority class; thus, the LSH-RHP is performed on it. In the case of multiple majority classes, this process is applied to all of them. (Related to Sec. \ref{sec: Locality Sensitive Hashing with Random Projections}.)}
    \label{fig: lsh}
\end{figure}

\noindent\textbf{Component Selection:} The purpose of this stage is to update the ensemble's components to maintain a balance between diversity and accuracy. The selected components from the previous stage are combined with the reserved components. This combined pool is then sorted based on component accuracy, and any extra components that exceed a predetermined pool size are deleted. The remaining components are clustered according to their prediction errors, and a predetermined number of high-performing components are chosen from each cluster. Finally, a modified version of the MMR (Maximal Marginal Relevance) algorithm \cite{abadifard2023dyned} is used in the final selection of components. 

This technique chooses a new set of components by considering both accuracy and diversity and ensures that the ensemble remains accurate while adapting to changes in the data stream. Newly picked components are designated as ``selected components"  to replace the active components, resuming the prediction stage, while the remaining components are returned to the ``reserved pool." Table \ref{tab: Symbols} shows the size limitations for these pools.

\subsection{Locality Sensitive Hashing with Random Hyperplane Projections}\label{sec: Locality Sensitive Hashing with Random Projections}

Locality Sensitive Hashing (LSH) \cite{liu2016multilinear} with Random Hyperplane Projections (RHP) is an effective method initially designed for approximate nearest-neighbor searches in high-dimensional data spaces. LSH-RHP leverages hash functions known as Multilinear Hash Functions, computing binary hash codes through multiple linear projection vectors \cite{liu2016multilinear}. These codes create buckets, grouping similar data points into locally dense regions within the feature space \cite{carraher2016random}.

The theoretical foundation of LSH-RHP is its ability to preserve angular similarity between points during the projection from high-dimensional to lower-dimensional spaces \cite{liu2016multilinear}. Mathematically, LSH-RHP utilizes vectors randomly sampled from a normal distribution:

\begin{equation}\label{eq: lsh}
\mathbf{h}_m(\mathbf{x}) = \text{sgn}(\mathbf{u}_1^T\mathbf{x}, \dots, \mathbf{u}_m^T\mathbf{x})
\end{equation}
with
\begin{equation*}
\text{sgn}(\mathbf{u}_l^T \mathbf{x}) = \begin{cases}
1, & \text{if } \mathbf{u}_l^T \mathbf{x} > 0 \\
0, & \text{otherwise}
\end{cases}
\end{equation*}

Each projection vector $\mathbf{u}_l$ is \textit{independently drawn from a standard Gaussian distribution} $N(0, I_{d \times d})$, where $I_{d \times d}$ is the $d$-dimensional identity matrix, $d$ is the original dimensionality of the data space, and $m$ is the number of projection vectors. This method ensures the maintenance of critical angular relationships between points after dimensionality reduction \cite{liu2016multilinear}.

The inspiration for applying LSH-RHP in undersampling for imbalanced streaming data arises from its inherent efficiency and adaptability \cite{liu2016multilinear}. In dynamic streaming environments, where the underlying class distribution frequently changes, traditional methods often struggle to adapt efficiently. Unlike random undersampling, which can inadvertently discard essential majority class information, or oversampling techniques (e.g., SMOTE \cite{chawla2002smote}), which risk overfitting and complexity, LSH-RHP offers a balanced approach. It efficiently groups similar \cite{johnson2019billion, 10.4028/www.scientific.net/amm.321-324.804} majority class instances, allowing informed and targeted undersampling while preserving crucial data instances \cite{wing2022hashing}. 

Moreover, the randomized and data-independent nature of LSH-RHP ensures that it remains effective even under evolving data distributions, eliminating the need for frequent reconfiguration or retraining of the hashing functions themselves. This characteristic significantly enhances its suitability for real-time processing of streaming data, aligning with the demands of dynamic scenarios \cite{jiang2019incremental}.

Recent research underscores that hashing-based methods, such as the Hashing-Based Undersampling Ensemble (HUE), effectively create diversified and representative training subsets by segmenting majority class data into meaningful subspaces \cite{wing2022hashing}. This strategy boosts ensemble performance by simultaneously increasing diversity and representativeness. Additionally, compared to traditional clustering approaches, LSH-RHP achieves comparable effectiveness with significantly reduced computational overhead due to its non-iterative, randomized design.

The advantages of LSH-RHP in the context of streaming data include:
\begin{itemize}
    \item \textbf{Computational Efficiency:} Sub-linear complexity enables rapid similarity assessments and partitioning \cite{10.1145/3274895.3274943};
    \item \textbf{Scalability:} Handles high-dimensional data effectively, maintaining efficiency regardless of data complexity \cite{10.1142/9789811204746_0018};
    \item \textbf{Adaptive Diversity:} Random projections naturally introduce variability in sampled subsets, enhancing ensemble robustness and generalization capability \cite{10.4028/www.scientific.net/amm.321-324.804, 10.4028/www.scientific.net/amm.556-562.3804};
    \item \textbf{Adaptability to Dynamic Class Ratios:} Capable of continuously adjusting to shifts in class proportions without extensive computational cost \cite{10.1109/cec.2014.6900653}.
\end{itemize}

\begin{figure}[t]
    \centering
    \resizebox{\textwidth}{!}{
    \begin{tabular}{c}
        \includegraphics{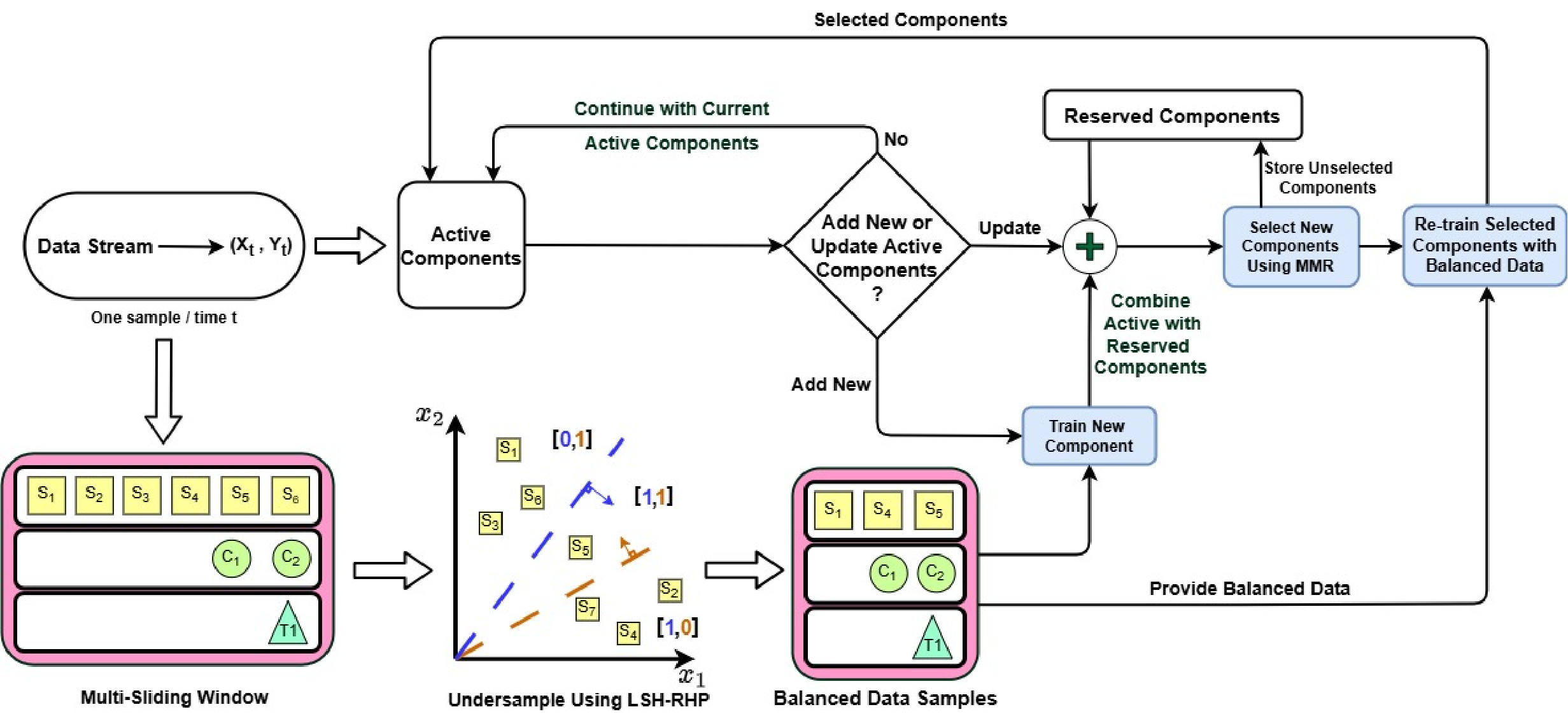}
    \end{tabular}} 
    \caption{\textbf{Working Principle of Integration of Locality Sensitive Hashing (LSH) with Random Hyperplane Projection (RHP) into DynED.} We assume this stream contains two features, $x_1$ and $x_2$, and three classes: Square (S), Circle (C), and Triangle (T). Active components predict the label of data samples at time $t$, while multi-sliding windows retain recent class-specific samples. Upon ensemble expansion, new components are trained on balanced datasets generated through undersampling. Component selection employs the third stage of the DynED approach. Selected components undergo retraining on balanced datasets before active components are replaced. (Related to Sec. \ref{sec: LSH-DynED}.)}
    \label{fig: lsh-dyned}
\end{figure}

\begin{table}[t]
    \centering  
    \caption{\textbf{LSH-DynED symbols and applicable default hyperparameter values.} (Related to Sec. \ref{sec: dyned-overview}, Sec. \ref{sec: experimental setting}, and Sec. \ref{sec: Hyperparameter Selection}.)}\label{tab: Symbols}
    \resizebox{1\textwidth}{!}{
    \begin{tabular}{llclll}
        \toprule
        Symbol & Meaning & Value & Symbol & Meaning & Value \\
        \midrule 
    $\xi$                  & Set of reserved components                        &N/A &
    $\theta$               & Threshold for the count of processed samples      & $1000^{\text{*}}$ \\ 

    $\xi_{slc}$            & Set of selected components                        & N/A & 
    $\Delta_{\lambda}$     & Variation in $\lambda$ parameter                  & $0.1^{\text{*}}$ \\ 
    
    $i$                    & Index of a sample in sliding window               & N/A & 
    $\mathit{S_p}$         & Maximum size of the component pool                & $500^{\text{*}}$ \\ 

    $Id_c$                 & Class identifier                                  & N/A & 
    $\mathit{S_{cls}}$     & \# of components to select from each cluster      & $10^{\text{*}}$ \\ 
    
    $\sigma$               & A data sample                                     & N/A & 
    $n_{train}$            & \# of train samples                               & 20 \\ 
    
    $\mathit{S_i}$         & Sample of indice $i$                              & N/A & 
    $\lambda$              & Controlling parameter of similarity and accuracy  & $0.6^{\text{*}}$ \\ 
    
    $D$                    & Data stream                                       & N/A & 
    $\mathit{S_w}$         & Window size                                       & 1000 \\ 
    
    $MW$                   & Multi-sliding window                              & N/A & 
    $n_v$                  & Number of vectors                                 & 5 \\ 

    $DD$                   & Drift detector                                    & N/A & 
    $\mathit{S_{slc}}$     & \# of active components for classification        & 10 \\ 

    $cls_{slc}$            & Set of selected components based on clustering    & N/A & 
    $n_{test}$             & \# of test samples                                & 50 \\ 
    
    $\mathit{S_c}$         & Sample of each class                              & N/A & \\

    $D_n$                  & A set of data samples from all classes            & N/A & \\
        \bottomrule
    \end{tabular}}
    \caption*{\text{*} These values are provided by DynED \cite{abadifard2023dyned} based on grid search, $\mathit{S_w}$ is set to 1000 \cite{aguiar2023survey}, with the aim of conducting a fair comparison across all methods, N/A: Not Applicable (No value is needed).}
\end{table}

Thus, integrating LSH-RHP into undersampling for imbalanced data streams effectively addresses crucial challenges, making it a reliable choice for maintaining model performance under dynamic conditions. Figure \ref{fig: lsh} illustrates the effectiveness of LSH-RHP in partitioning the feature space for targeted and informed downsampling.

\subsection{Integrating LSH-RHP into DynED}\label{sec: LSH-DynED}
This section details how we integrate LSH-RHP into DynED to form our proposed method.

To effectively address the challenges posed by imbalanced data streams, DynED's process must be extended, and the advantages of LSH-RHP must be incorporated to enable a more informed undersampling approach for the majority classes. It is worth noting that DynED is a General-Purpose method (GPM) without specific mechanisms for handling imbalanced data streams. This integration aims for the ensemble's training to include evenly distributed samples, effectively reducing the bias towards the majority classes in the stream. The goal is to leverage the strengths of both methods to create a more robust and scalable classification method. It is important to note that while DynED directly addresses concept drift through its dynamic ensemble updating mechanism, the integration of LSH-RHP is primarily intended to manage class imbalance. As a result, LSH-RHP indirectly enhances ensemble robustness to evolving class distributions rather than explicitly handling concept drift. The integrated method continuously monitors ensemble accuracy using the ADWIN \cite{bifet2007learning} drift detector within DynED.

\begin{algorithm}
\captionsetup{name=Class}
    \caption{LSH-DynED: MultiSlidingWindow-Undersampling Class.}
    \label{alg:slidingwindowlsh}
    \begin{algorithmic}[1]
        \State \textbf{Class}: Multi-SlidingWindow-Undersampling
        \State \textbf{Properties}:
        \State \quad \textit{window\_size}: Integer  \LeftComment{Maximum $\mathit{S_w}$ per class}
        \State \quad \textit{class\_list}: List      \LeftComment{Identifiers $Id_c$ for each class}
        \State \quad \textit{window}: Dictionary     \LeftComment{Stores $\mathit{S_c}$ for each class}
        
        \Method{Initialize}{$w_{size}$, $Id_c$}
            \State $\textit{window\_size} \leftarrow \mathit{S_w}$
            \State $\textit{class\_list} \leftarrow Id_c$
            \State $\textit{window} \leftarrow \text{Initialize dictionary with keys in } Id_c$
        \EndMethod
        \Method{AddToWindow}{$\sigma$, $class\_id$}
            \If {$\text{window}[class\_id]$ is $>$ $\mathit{S_w}$} 
                \State $\text{Remove oldest sample from } \text{window}[class\_id]$
            \EndIf
            \State $\text{Add } \sigma \text{ to } \text{window}[class\_id]$
        \EndMethod
        \Method{GetDataForPrediction}{$n_{test}$}       \LeftComment{Retrieve test samples}
            \State $\text{Prepare } \mathcal{D}_N \text{ by selecting the last } N_{test} \text{ samples from each class}$
            \State $\text{Shuffle and return } \mathcal{D}_N \text{ and corresponding labels}$
        \EndMethod
        \Method{FetchDataPoints}{$class\_id$, $i$}
            \State Retrieve specific samples $S_i$ based on indices $i$ from window[$class\_id$]
            \State $\text{return } \mathit{S_i}$
        \EndMethod
        \Method{UndersampleDataForTraining}{$n_{train}$} \LeftComment{Perform LSH-RHP on majority classes}
            \State Identify the majority classes ($class\_id$) by calculating the $WAR_t$ using Eq. \ref{Eq: wightedration}
            \For {Each Majority class} 
                \State Initialize the $n_v$ vectors of the hashing function
                \State Get the buckets using the LSH-RHP strategy
                \State $\text{Specify } n_{train} \text{ indices } i \text{ from buckets randomly}$
                \State $\text{Fetch the data by} \textsc{FetchDataPoints} \text{ method and put into } \mathcal{D}_N$
            \EndFor
            \State $\text{Shuffle and return }\mathcal{D}_N \text{ the selected data and labels}$ \LeftComment{Retrieve balanced train samples}
        \EndMethod
    \end{algorithmic}
\end{algorithm}

Current methods in the literature often utilize a single sliding window as a buffer, containing a fixed-size sample for training the components of an ensemble. However, due to inherent imbalances in the data stream, this approach inadvertently introduces a bias towards the majority classes, as highlighted by Cano et al. \cite{cano2022rose}. The first adaptation of our approach is the use of multi-sliding windows described in Class \ref{alg:slidingwindowlsh}, which is required by the LSH-RHP technique. Specifically, each class allocates its own sliding window to buffer the most recent data samples of a fixed size (Class \ref{alg:slidingwindowlsh}, lines 6-16).

\addtocounter{algorithm}{-1}
\captionsetup[algorithm]{name=Algorithm}
\begin{algorithm}
\captionsetup{name=Algorithm}
    \caption{LSH-DynED: Main flow.}
    \centering
    \label{alg: main}
    \begin{algorithmic}[1]
        \Require $D$, $MW$, $DD$, $\mathit{S_{slc}}$, $\mathit{S_w}$, $\mathit{S_p}$, $\Delta_{\lambda}$, $n_{train}$
        \Ensure $\hat{y_k}$: Prediction of each sample at time $t$
        \State $\xi_{slc} \leftarrow \text{Initialize } slc_s \text{ components }$
        \While{$D$ has more samples}
            \State $sample \leftarrow \text{Get next sample from } D$
            \State $\hat{y_k} \leftarrow \text{Use }\xi_{slc}\text{ to predict using majority voting}$
            \State $MW.\textsc{AddToWindow}(sample)$ \Comment{ Add sample to the sliding window}
            \State $DD\leftarrow \text{ Update drift detector}$
            \State Train $\xi_{slc}$ on $sample$
            \If{$DD$ detects drift}
                \State $Data, Labels \leftarrow MW.\textsc{UndersampleDataForTraining}(n_{train})$
                \State $\xi \leftarrow$ Generate and train a new component on $Data, Labels$, add to the pool
            \EndIf
            \If{$\text{passed samples count} \ge \theta$ or new component is added}
                \State Update $\lambda$ with $\Delta_{\lambda}$ step size 
                \State $\xi_{slc}$, $\xi \leftarrow \text{Update } \xi_{slc} \text{ and } \xi \text{ using Algorithm }\ref{alg: component Selection}$
            \EndIf
        \EndWhile
    \end{algorithmic}
\end{algorithm}

This adaptation addresses the challenge of availability; classes with fewer samples will see their respective windows refreshed less frequently compared to the majority classes. As a result, some data samples from minority classes are consistently available. To correctly choose the majority classes, we continuously calculate the weighted average of class ratios (Class \ref{alg:slidingwindowlsh}, line 26) using the following equation:
\begin{equation}\label{Eq: wightedration}
\centering
    \text{\textbf{W}eighted\ \textbf{A}verage\ of \textbf{R}atios } (WAR_t) = \frac{\sum_{i \in C} r_i^t \cdot n_i^t}{\sum_{i \in C} n_i^t}
\end{equation}
Where $C$ represents the set of all classes in the stream, with $r_i^t$ denoting the ratio and $n_i^t$ the sample count for class $i$ up to time $t$, and they are subject to change dynamically over time $t$. The $WAR_t$ can range in value from $0$ to $1$. A class whose ratio exceeds the $WAR_t$ is considered a majority class and a candidate for undersampling. As shown in Figure \ref{fig: data_dynamic}, the $WAR_t$ tends to converge and approach a steady state. This trend suggests that initial fluctuations in class distribution become less volatile as more data is processed and ensures our undersampling technique is dynamically adjusted to prevent minority classes from being underrepresented, improving model performance. 

After determining the majority classes, we set up the hash function with a certain number of vectors $n_v$. Next, the buckets are created. The samples are then drawn randomly from each bucket based on the number of samples in each. However, the total number of samples for each majority class does not surpass a certain preset limit $n_{train}$. By selectively undersampling the majority classes (Class \ref{alg:slidingwindowlsh}, line 27-32), we focus on a subset of samples that encompasses the feature space of those classes, resulting in more balanced training data for the ensemble components.

\begin{algorithm}[h]
    \caption{LSH-DynED: Component selection.}
    \centering
    \label{alg: component Selection}
    \begin{algorithmic}[1]
        
        \Require $MW$, $\xi$, $\xi_{slc}$, $\mathit{S_{slc}}$, $\mathit{S_p}$, $\mathit{S_{cls}}$, $n_{train}$, $n_{test}$
        \Ensure $\xi$, $\xi_{slc}$
        \State $P \leftarrow \xi + \xi_{slc}$
        \State $\text{Sort } P \text{ by Kappa statistic and limit to } \mathit{S_p}$
        \State $Data, Labels \leftarrow MW.\textsc{GetDataForPrediction}(n_{test})$
        \State $PE \leftarrow$ Calculate prediction error for each component in $P$ using $Data$ and $Labels$

        \State Perform KMeans clustering on $PE$ with two clusters
        \State $cls \leftarrow$ Labels from KMeans clustering
        \State $cls_{slc} \leftarrow$ Select top $\mathit{S_{cls}}$ components from each cluster based on highest accuracy
        \State $Diversity \leftarrow$ Calculate Kappa metric for components in $cls_{slc}$
        \State $\xi_{slc} \leftarrow$ Apply MMR on $cls_{slc}$ to select $\mathit{S_{slc}}$ based on diversity and accuracy
        \State $\xi\leftarrow P \text{ excluding } \xi_{slc}$ \Comment{All other components remain in reserve}
        \State $Data, Labels \leftarrow MW.\textsc{UndersampleDataForTraining}(n_{train})$
        \State $\xi_{slc} \leftarrow$ Re-train selected components on $Data$ and $Labels$
        \Return $\xi$, $\xi_{slc}$
    \end{algorithmic}
\end{algorithm}

In the next step, we incorporate the previously introduced undersampling method into the second stage—Drift Detection and Adaptation. Thus, this stage functions as follows:

Initially, as detailed in Section \ref{sec: dyned-overview}, the drift detector is updated (Algorithm \ref{alg: main}, line 6). Upon detecting drift, a new component is created and trained with a balanced and fixed sample size (Algorithm \ref{alg: main}, lines 8-10), using the method outlined in Section \ref{sec: Locality Sensitive Hashing with Random Projections}. This procedure ensures the new component is trained on balanced samples, eliminating bias towards any class. Besides, in the case of several consecutive runs, this technique generates an entirely different set of balanced training data to help increase diversity among the ensemble's components. Following this, the component is added to a pool of ``reserved components."

If a new component is trained or if the number of processed samples surpasses the $\mathbf{\theta}$ threshold, then the parameter $\mathbf{\lambda}$ is dynamically adjusted (Algorithm \ref{alg: main}, lines 13). This adjustment is based on changes in the ensemble's accuracy over time, leading to the activation of the third stage—Component Selection.
In the updated third stage, we introduced another adaptation. DynED traditionally utilizes accuracy and double-fault (df) diversity measures \cite{kuncheva2003measures} to select components. To address the imbalance problem more effectively, we replaced the previously used diversity measure with the Kappa statistic \cite{TSYMBAL200583} (Algorithm \ref{alg: component Selection}, lines 8-10). Additionally, suppose the component count exceeds the predefined limit within the pool. In that case, they are sorted by their Kappa statistic values, and the least performing component is removed (Algorithm \ref{alg: component Selection}, line 2).

A crucial aspect of the DynED workflow is the online training of ``selected components" for incoming samples. Given the imbalanced nature of the data stream, this training typically introduces a bias towards the majority class. To mitigate this bias, we enhanced this stage by incorporating additional training sessions for newly ``selected components" using balanced data samples obtained through our undersampling method, an element absent in the original DynED framework (Algorithm \ref{alg: component Selection}, lines 11-12). Figure \ref{fig: lsh-dyned} provides an abstract representation of the proposed method's workflow.


\subsection{Time Complexity Analysis}\label{sec: Time Complexity Analysis}
As noted in \cite{abadifard2023dyned}, the time complexity of the DynED is approximated as $\mathcal{O}(n^2)$, where the $n$ stands for the number of components in the ensemble. The time complexity of the LSH-RHP involves the following components: we only utilize LSH-RHP to build buckets and divide the feature space of each class into portions, and we skip the nearest neighbor finding step.

Thus, the time complexity of generating a hyperplane is $\mathcal{O}(d)$ if the dataset includes a $d$-dimensional feature space and it repeats this process $k$ times to generate $k$ hyperplanes. So far, the time complexity has become $\mathcal{O}(kd)$. For the worst-case scenario, we assume that the majority class's sliding window is full and contains $v$ samples. 

Therefore, the time complexity of obtaining the hash code for samples of that class is $\mathcal{O}(vkd)$. This step is repeated for $c$ times, where $c$ represents the number of majority classes determined by $WAR_t$ (Eq. \ref{Eq: wightedration}), and the time complexity becomes $\mathcal{O}(cvkd)$. Thus, the overall time complexity of the method is $\mathcal{O}(n^2\times cvkd)$. 

\section{Experimental Evaluation}\label{sec: Experimental Evaluation}
In this section, we present our experimental evaluation. In Section \ref{sec: Experimental Environment}, we describe the experimental environment, including the baselines (\ref{sec: baselines}), datasets (\ref{sec: datasets}), parameter settings (\ref{sec: experimental setting}), evaluation metrics (\ref{sec: effective measures}), and hyperparameter selection (\ref{sec: Hyperparameter Selection}). Section \ref{sec: effective analysis} reports the effectiveness results and statistical comparisons, with detailed outcomes in \ref{sec: General Comparison} and significance tests in \ref{sec: statistical analysis}. In Section \ref{Sec: impact of class characteristics}, we analyze the impact of class characteristics, including the number of classes (\ref{sec: Impact of Dataset Characteristics}) and dynamic imbalance ratios (\ref{sec: Impact of Dynamic Imbalance Ratios}). Section \ref{sec: ablation study} provides an ablation analysis, and Section \ref{sec: overall} summarizes key findings.

\subsection{Experimental Environment} \label{sec: Experimental Environment}
This section introduces the experimental environment of our study, which encompasses the baselines, datasets, experimental settings, and the effectiveness measures employed for comparison.


\subsubsection{\textbf{Baselines}}\label{sec: baselines}

The experiments use 15 classification methods as baselines capable of handling multi-class non-stationary data streams. Six of them are high-performing general-purpose methods (GPM), while the remaining nine are specifically designed to address the issues of imbalanced data streams (ISM). The reason for including GPM methods in our evaluation is that although they are not explicitly equipped with imbalance handling strategies, they demonstrate robust performance in such environments. Among these six GPM methods, BELS is the only neural network-based method. The characteristics of these methods are detailed in Table \ref{table: methods}. We employed the publicly accessible source code on GitHub\footnote{\href{https://github.com/canoalberto/imbalanced-streams}{Imbalanced-Streams}}, which is created by Aguiar et al. \cite{aguiar2023survey} on top of the MOA framework \cite{bifet2010moa} to run our experiments. Additionally, the Python implementation of BELS \cite{bakhshi2023broad} is available publicly on the author's GitHub\footnote{\href{https://github.com/sepehrbakhshi/BELS}{BELS}}. Our proposed method is implemented in Python 3.11.7 using the River library 0.21.1 \cite{montiel2021river} and the Faiss library 1.7.4 \cite{douze2024faiss}, with a Hoeffding Tree as the base classifier. The source code for our method is also publicly available on GitHub\footnote{\href{https://github.com/soheilabadifard/LSH-DynED}{LSH-DynED}} with all necessary details to ensure the reproducibility of our experiments.
\begin{table}[t]
\centering
\caption{\textbf{Datasets (all imbalanced) and their characteristics.} (Related to Sec. \ref{sec: datasets}, Sec. \ref{sec: Impact of Dataset Characteristics}, Sec. \ref{sec: Impact of Dynamic Imbalance Ratios}, and Sec. \ref{sec: ablation study}.)}\label{table: datasets}
\resizebox{\textwidth}{!}{
\begin{tabular}{lrrrlrrcc}
\toprule
Dataset               & Instances  & Features  & Classes   & Type & Max Ratio & Min Ratio & Dynamic Ratio Behavior & Drift \\
\midrule
Activity              & 5,418      & 45        & 6         & real & 0.384090 & 0.045404 & ---        & \checkmark \\
Balance               & 625        & 4         & 3         & real & 0.460800 & 0.078400 & \checkmark & Unknown \\
Connect-4             & 67,557     & 42        & 3         & real & 0.658303 & 0.095460 & ---        & Unknown \\
Contraceptive         & 1,473      & 9         & 3         & real & 0.427020 & 0.226069 & \checkmark & Unknown  \\
Covtype               & 581,012    & 54        & 7         & real & 0.487599 & 0.004728 & ---        & Unknown  \\
Crimes                & 878,049    & 3         & 39        & real & 0.199192 & 0.000007 & ---        & Unknown  \\
Ecoli                 & 336        & 7         & 8         & real & 0.425595 & 0.005952 & \checkmark & Unknown  \\
Fars                  & 100,968    & 29        & 8         & real & 0.417122 & 0.000089 & ---        & Unknown  \\
Gas                   & 13,910     & 128       & 6         & real & 0.216319 & 0.118973 & ---        & \checkmark \\
Glass                 & 214        & 9         & 6         & real & 0.355140 & 0.042056 & \checkmark & Unknown  \\
Hayes-roth            & 132        & 4         & 3         & real & 0.386364 & 0.227273 & \checkmark & Unknown  \\
Kr-vs-k               & 28,056     & 6         & 18        & real & 0.162283 & 0.000962 & \checkmark & \checkmark \\
New-thyroid           & 215        & 5         & 3         & real & 0.697674 & 0.139535 & ---        & Unknown  \\
Olympic               & 271,116    & 7         & 4         & real & 0.853262 & 0.048378 & ---        & Unknown  \\
Pageblocks            & 548        & 10        & 5         & real & 0.897810 & 0.005474 & ---        & Unknown  \\
Poker                 & 829,201    & 10        & 10        & real & 0.501116 & 0.000002 & ---        & \checkmark  \\
Shuttle               & 57,999     & 9         & 7         & real & 0.785979 & 0.000172 & ---        & \checkmark  \\
Tags                  & 164,860    & 4         & 11        & real & 0.330462 & 0.008377 & ---        & Unknown  \\
Thyroid-s             & 720        & 21        & 3         & real & 0.925000 & 0.023611 & ---        & Unknown  \\
Thyroid-l             & 7,200      & 21        & 3         & real & 0.925833 & 0.023056 & ---        & Unknown  \\
Wine                  & 178        & 13        & 3         & real & 0.398876 & 0.269663 & \checkmark & Unknown  \\
Yeast                 & 1,484      & 8         & 10        & real & 0.311995 & 0.003369 & ---        & Unknown  \\
Zoo                   & 1,000,000  & 17        & 7         & real & 0.396504 & 0.042792 & ---        & \checkmark  \\
ACTIVITY-D1           & 10,853     & 43        & 4         & semi-syn & 0.368193 & 0.125495 & \checkmark & \checkmark  \\
CONNECT4-D1           & 67,557     & 42        & 3         & semi-syn & 0.343340 & 0.313409 & \checkmark & \checkmark  \\
COVERTYPE-D1          & 581,012    & 54        & 4         & semi-syn & 0.474603 & 0.031383 & \checkmark & \checkmark  \\
CRIMES-D1             & 878,049    & 3         & 4         & semi-syn & 0.329192 & 0.149404 & \checkmark & \checkmark  \\
DJ30-D1               & 138,166    & 7         & 4         & semi-syn & 0.357078 & 0.107537 & \checkmark & \checkmark  \\
GAS-D1                & 13,910     & 128       & 3         & semi-syn & 0.488497 & 0.131416 & \checkmark & \checkmark  \\
OLYMPIC-D1            & 271,116    & 7         & 3         & semi-syn & 0.353432 & 0.321840 & \checkmark & \checkmark  \\
POKER-D1              & 829,201    & 10        & 4         & semi-syn & 0.340320 & 0.024806 & \checkmark & \checkmark  \\
SENSOR-D1             & 2,219,802  & 5         & 4         & semi-syn & 0.372768 & 0.090036 & \checkmark & \checkmark  \\
TAGS-D1               & 164,860    & 4         & 4         & semi-syn & 0.349945 & 0.092327 & \checkmark & \checkmark  \\
\bottomrule
\end{tabular}}
\end{table}

\begin{figure}[t]
    \centering

    \begin{subfigure}[b]{0.32\textwidth}
        \includegraphics[width=\textwidth]{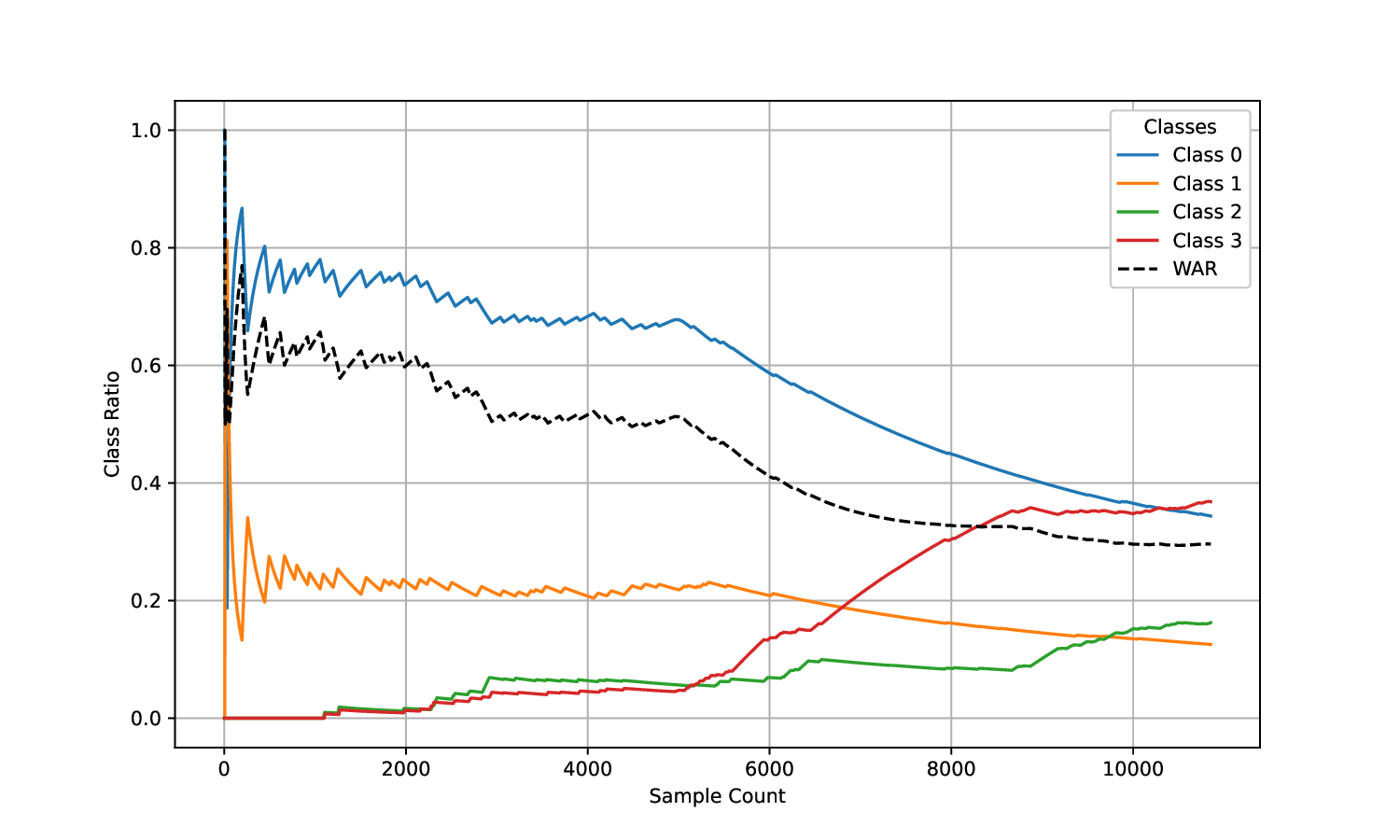}
        \caption{ACTIVITY-D1.}
        \label{fig:activity-d1}
    \end{subfigure}
    \hfill
    \begin{subfigure}[b]{0.32\textwidth}
        \includegraphics[width=\textwidth]{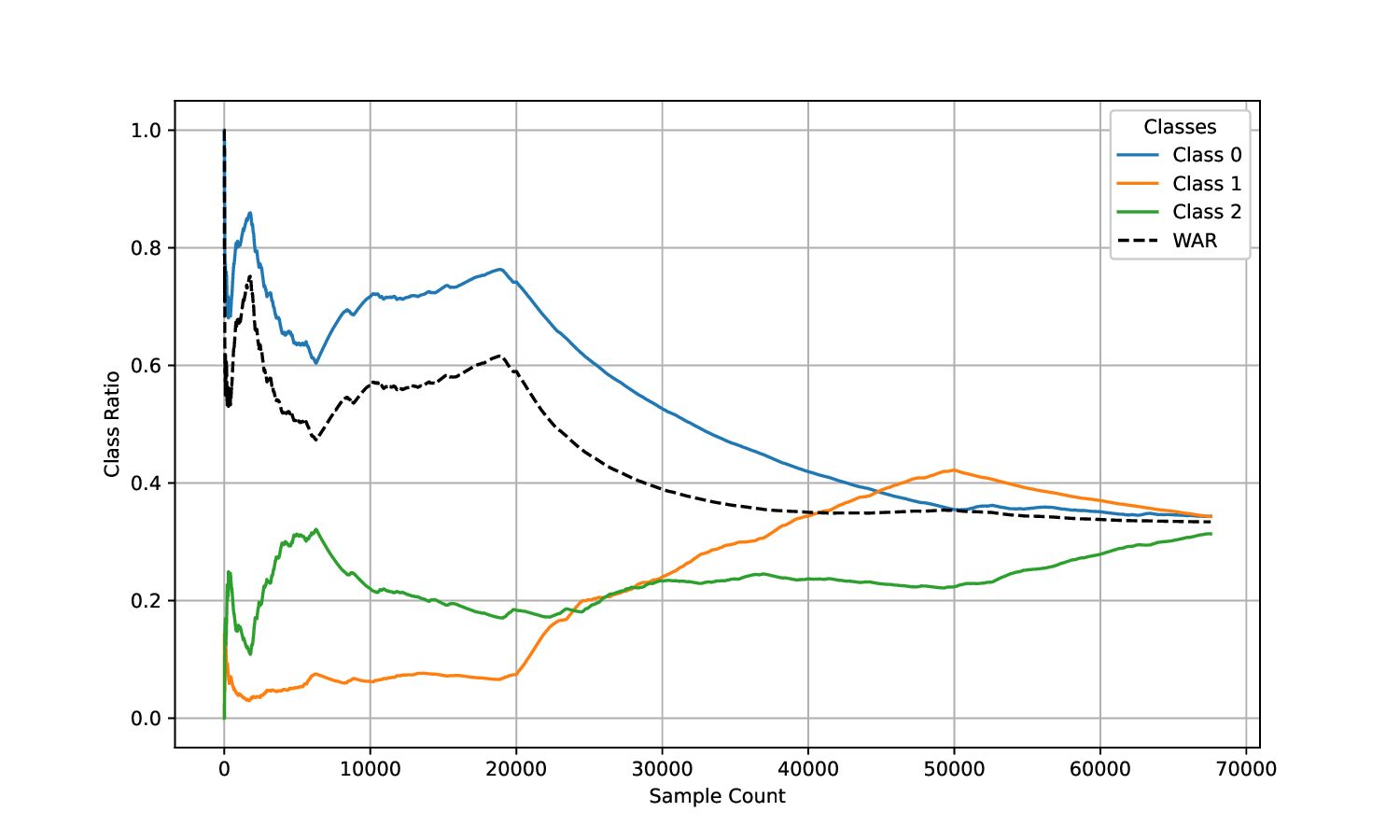}
        \caption{CONNECT4-D1.}
        \label{fig:connect4-d1}
    \end{subfigure}
    \hfill
    \begin{subfigure}[b]{0.32\textwidth}
        \includegraphics[width=\textwidth]{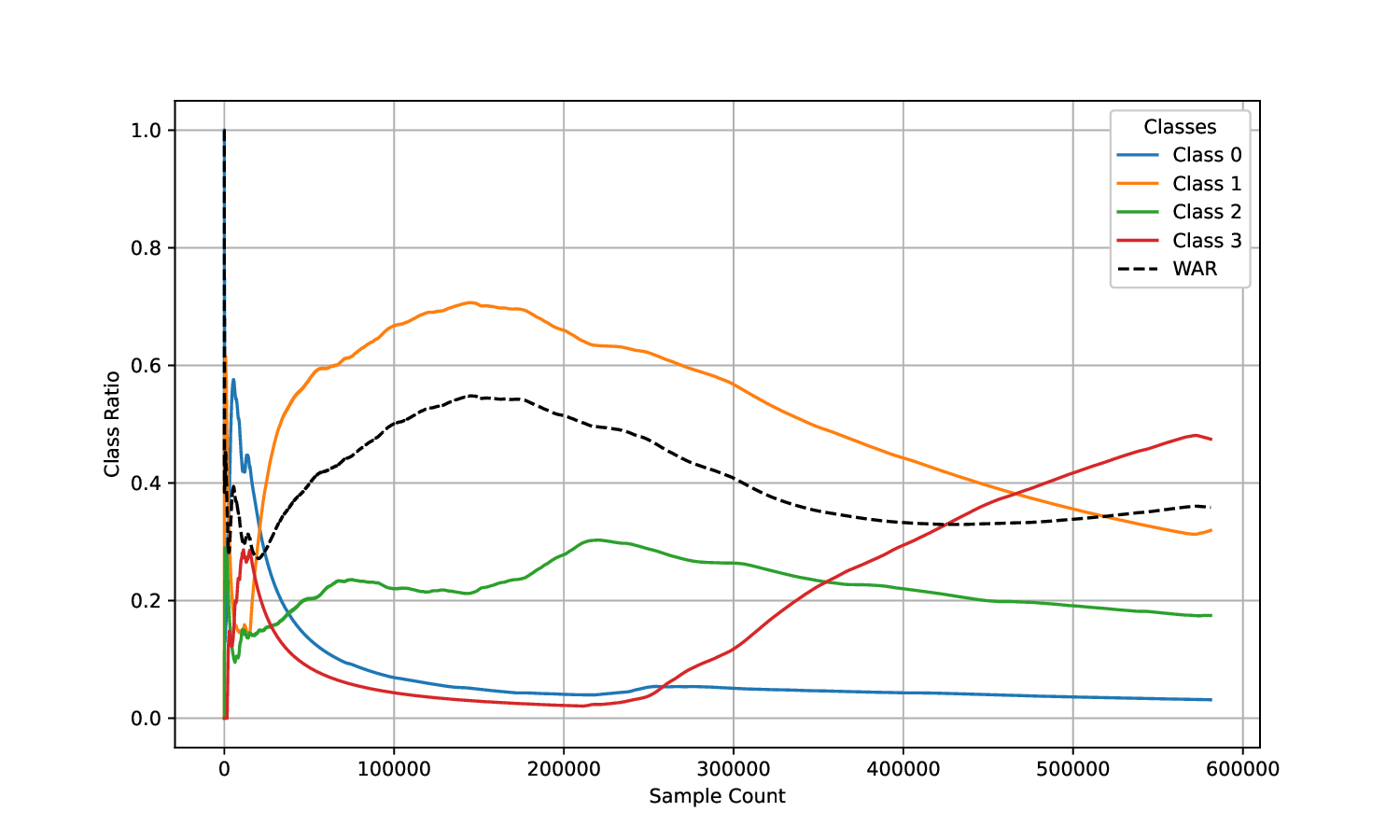}
        \caption{COVERTYPE-D1.}
        \label{fig:covertype-d1}
    \end{subfigure}



    \begin{subfigure}[b]{0.32\textwidth}
        \includegraphics[width=\textwidth]{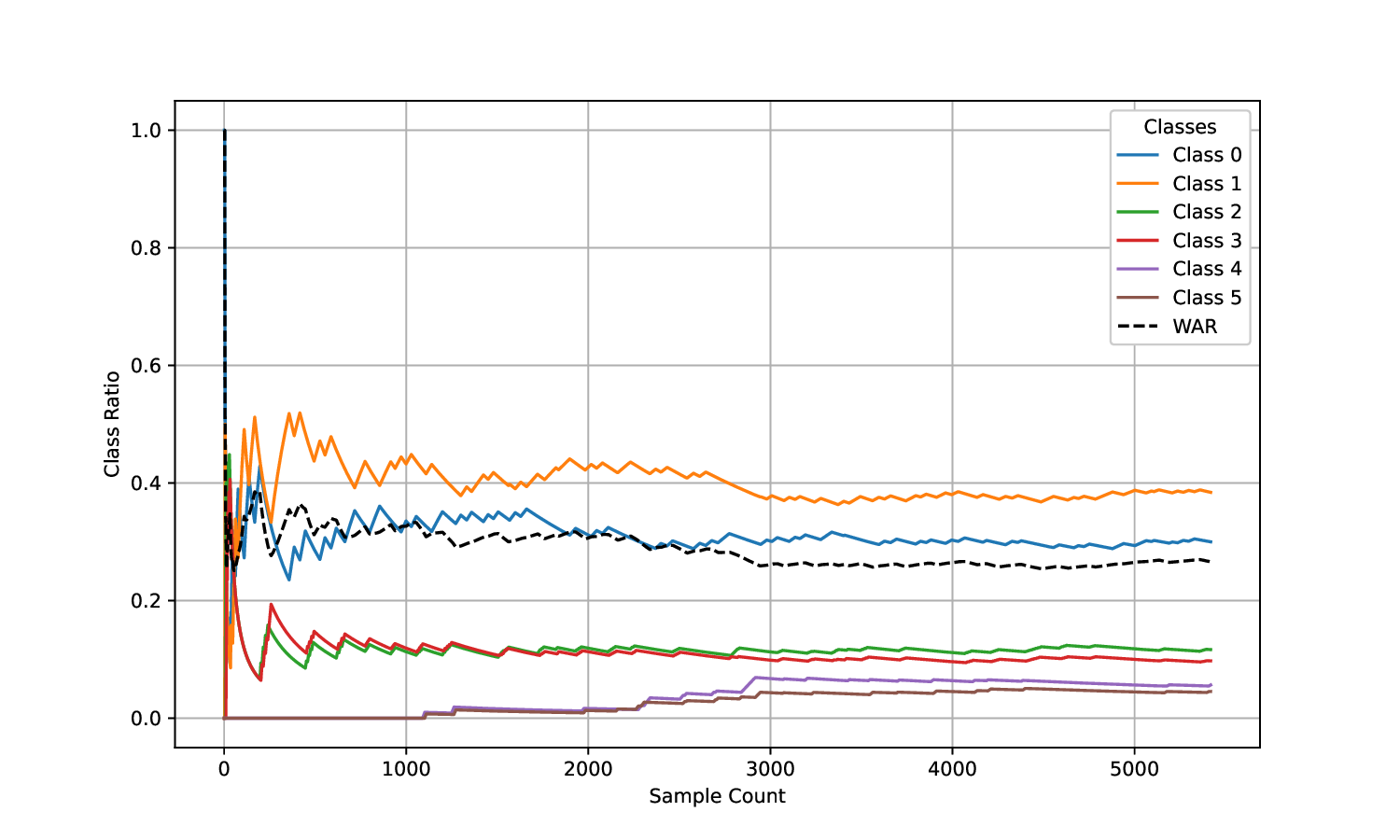}
        \caption{Activity.}
        \label{fig:Activity_normal}
    \end{subfigure}
    \hfill
    \begin{subfigure}[b]{0.32\textwidth}
        \includegraphics[width=\textwidth]{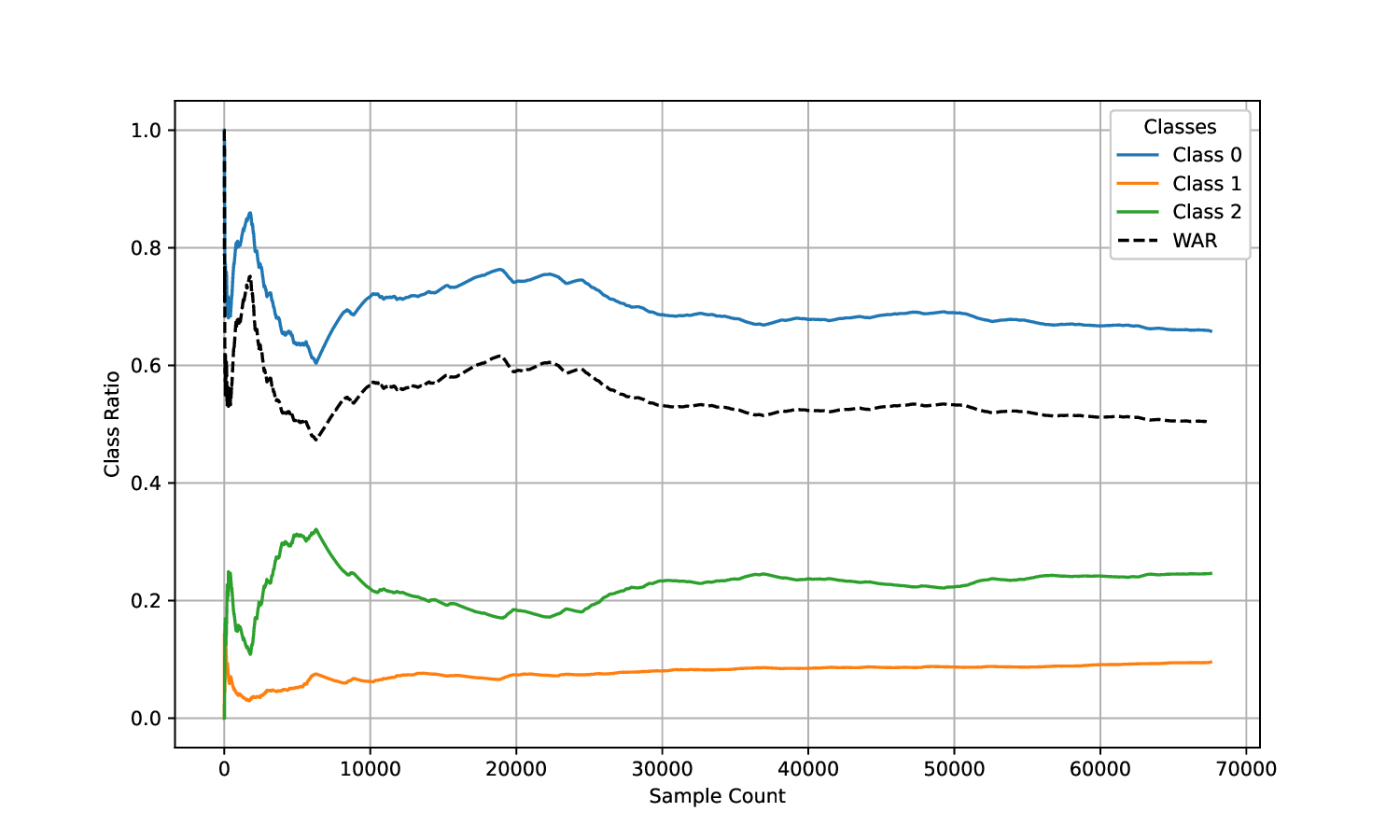}
        \caption{Connect-4.}
        \label{fig:Connect_normal}
    \end{subfigure}
    \hfill
    \begin{subfigure}[b]{0.32\textwidth}
        \includegraphics[width=\textwidth]{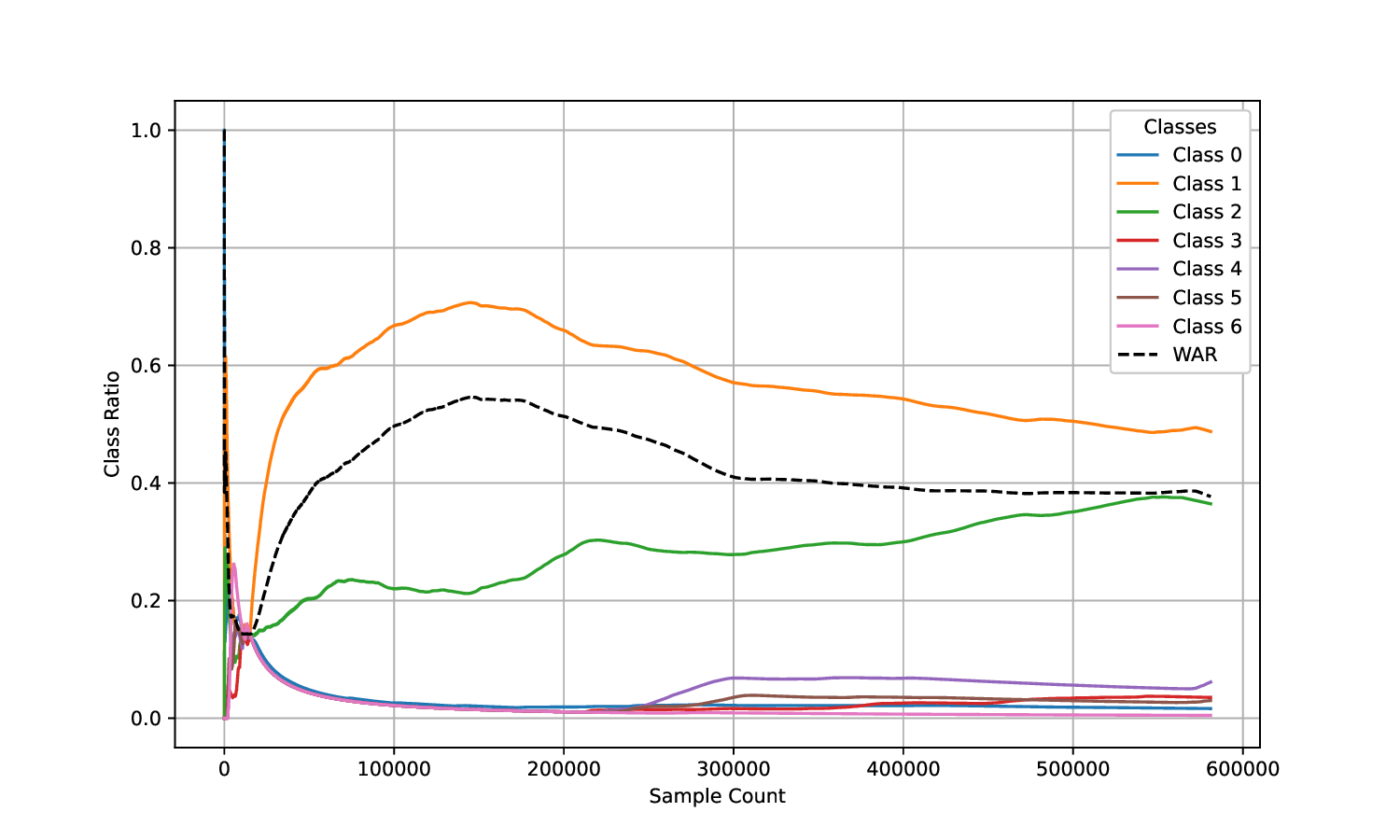}
        \caption{Covtype.}
        \label{fig:Covtype_normal}
    \end{subfigure}
    
    \caption{\textbf{Plots of semi-synthetic datasets with dynamic imbalance ratios (a, b, c) and datasets with no noticeable imbalance variations (d, e, f).} $WAR_t$, as presented in Eq. \ref{Eq: wightedration}, is also shown in all plots. (Related to Sec. \ref{sec: LSH-DynED}, Sec. \ref{sec: datasets}, and Sec. \ref{sec: Impact of Dynamic Imbalance Ratios}.)}
    \label{fig: data_dynamic}
\end{figure}

\subsubsection{\textbf{Datasets}}\label{sec: datasets}

To assess the effectiveness, robustness, and resilience of the proposed method, relative to established baselines, we select 33 imbalanced datasets, detailed in Table \ref{table: datasets}, which include 23 real datasets commonly used in the data stream domain and publicly accessible from UCI repository \cite{kelly2023uci}, KEEL repository \cite{derrac2015keel} and Kaggle competitions and ten semi-synthetic data streams as described by Korycki et al. \cite{korycki2020online}. For some datasets, the presence of concept drift is explicitly noted, whereas those marked as "Unknown" should not be assumed to be stationary or devoid of concept drift. The semi-synthetic streams are specifically designed for scenarios where class ratios fluctuate considerably, as depicted in Figure \ref{fig: data_dynamic}, making them ideal for testing methods' resilience toward dynamic imbalance ratios in uneven multi-class streams. Concurrently, certain real-world datasets also display the dynamic imbalance ratio behavior. This observation is denoted by the "\checkmark" symbol in the Dynamic Ratio Behavior column of Table \ref{table: datasets}. We chose to examine both real-world and semi-synthetic datasets because they introduce different challenging situations, such as confusing inter-class interactions. These conditions are crucial for understanding classifier performance under challenging scenarios. Furthermore, these datasets are curated to replicate specific behaviors and lack the apparent probabilistic features commonly found in stream generators \cite{aguiar2023active}.

\subsubsection{\textbf{Experimental Settings}}\label{sec: experimental setting}

All methods are evaluated in a single pass and online using the interleaved-test-then-train approach \cite{gama2009issues}, which is critical for determining the adaptability of models to new data as it arrives. In machine learning, choosing the right hyperparameters is critical since they significantly impact model performance. To guarantee a fair comparison across all methods, we used the suggested hyperparameters for each model as specified by the original authors. We hard-coded the chunk size to one in the BELS model to replicate the behavior of online models. We determined values after conducting tests with various hyperparameters based on the grid search method as described in Sec. \ref{sec: Hyperparameter Selection}. Table \ref{tab: Symbols} details the hyperparameters for our proposed technique, which are not tuned to any specific dataset to preserve its broad applicability. The experiments are carried out on a server with an Intel(R) Xeon(R) Gold 5118 CPU @ 2.30 GHz, 128 GB RAM, and Ubuntu 18.04.4 LTS.

Although our method is designed with computational efficiency in mind, we do not report direct execution time comparisons. In the experiments, all baseline methods were run using the original code provided by their respective authors. Due to the fundamental differences between the frameworks used (Java vs. Python), a direct comparison of computational performance would be unreliable. Such comparisons would mix algorithmic efficiency with platform-dependent factors, preventing meaningful comparisons of execution time and memory usage \cite{hamaamin2024java, Jain_2023}. Instead, we emphasize the theoretical time complexity analysis presented in Section \ref{sec: Time Complexity Analysis}, which offers a more reliable basis for evaluating efficiency in this context.
\begin{figure}[t]
\centering
    \includegraphics[width=0.8\textwidth, keepaspectratio]{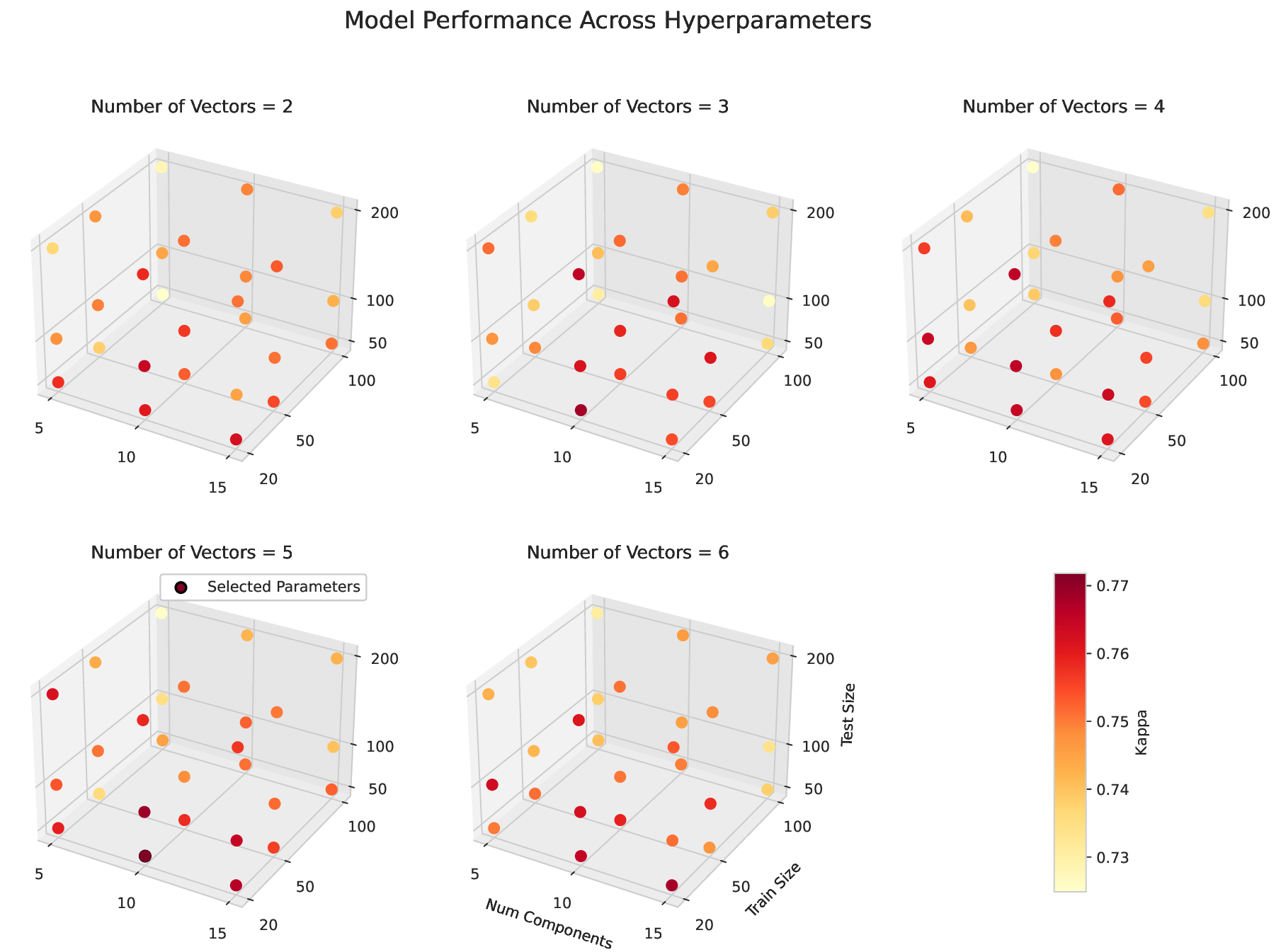}
    \caption{\textbf{Hyperparameter Selection.} (Related to Sec. \ref{sec: Hyperparameter Selection}.)}
    \caption*{ $x-axis$ [5-15]: Number of Active Components ($\mathit{S_{slc}}$), $y-axis$ [20-100]: Number of Train Samples ($n_{train}$), and $z-axis$ [50-200]: Number of Test Samples ($n_{test}$), The selected hyperparameters are observed in the cube with Number of Vectors = 5. Selected Default Parameters: $n_{train}= 20$, $n_v= 5$, $\mathit{S_{slc}}= 10$, and $n_{test}= 50$.}
    \label{fig: hyper-param-selec}
\end{figure}

\subsubsection{\textbf{Effectiveness Measures and Justification of Their Selection}}\label{sec: effective measures} 
To evaluate the methods, we utilized the Kappa \cite{aguiar2023survey} and mG-Mean (multi-class G-Mean) \cite{espindola2005extending, wang2020auc, wang2016dealing} metrics. They are both introduced to measure the classification effectiveness of multi-class classification methods in imbalanced environments. 

The Kappa measures inter-rater reliability and is adjusted for chance agreement. It is calculated as shown in Eq. \ref{eq: kappa}:
\begin{equation}\label{eq: kappa}
    \text{Kappa} = \frac{n \sum_{i=1}^{c} x_{ii} - \sum_{i=1}^{c} x_{i.} x_{.i}}{n^2 - \sum_{i=1}^{c} x_{i.} x_{.i}}
\end{equation}

Here, $n$ is the total number of observations, and $x_{ii}$ represents the instances where both raters agree on category $i$. The terms $x_{.i}$ and $x_{i.}$ denote the columns and row total classifications each rater makes into category $i$, respectively. The product of these terms provides the expected frequency of agreements by chance for each category. The numerator of Kappa calculates the excess of observed agreements over those expected by chance, indicating the actual agreement adjusted for randomness. The denominator represents the maximum possible excess of agreements over chance. The resulting value ranges from -1 to 1, where 1 indicates perfect agreement, 0 suggests chance-level agreement, and negative values indicate less agreement than expected by chance. 

This makes Kappa especially advantageous in imbalanced contexts as it penalizes homogeneous predictions and effectively reflects performance changes and shifts in class distributions across multiple classes over time within multi-class imbalanced data streams \cite{aguiar2023survey}.

Additionally, we report the mG-Mean (multi-class G-Mean), which is an essential metric for evaluating classification model performance over imbalanced class distributions: 
\begin{align} \label{pmauc}
    \text{mG-Mean} &= \sqrt[n]{\prod_{i=1}^{n} \text{sensitivity}_i}\\
    \text{sensitivity}_i = \text{recall}_i &= \frac{{TP}_i}{{TP}_i + {FN}_i}
\end{align}

Where $\text{sensitivity}_i$ indicates the class-wise sensitivity. The mG-Mean provides a comprehensive picture of model performance by equally accounting for all classes, thus maintaining a balanced evaluation in multi-class imbalanced environments, preventing majority classes from dominating results, and offering robustness against biases arising from uneven class distributions \cite{espindola2005extending, wang2016dealing}. All of these metrics are calculated across a 500-instance window. 

While other metrics, such as Accuracy, WMAUC, EWMAUC, and PMAUC \cite{wang2020auc} are commonly considered, their applicability is limited in dynamic multi-class imbalanced streams. Accuracy can be misleading, heavily favoring majority classes and masking minority class performance; WMAUC and EWMAUC, although incorporating dynamic weighting, suffer from interpretability issues and deteriorating robustness when class distributions shift continuously; and PMAUC, despite its usefulness in stable multi-class scenarios, fails to account for dynamic class imbalance and is thus insensitive to dynamic class distributions. Consequently, it often does not capture the complexities of multi-class imbalance, limiting both its informativeness and practicality in real-time streaming contexts \cite{10.1016/j.neunet.2018.07.011, 10.17485/ijst/v16i16.146, luo2023multiclass, pinto2023imbalance}. Given these considerations, we prioritized Kappa and mG-Mean as the primary evaluation metrics in this study, as they are designed or robustly validated for dynamic, imbalanced multi-class scenarios.



\subsubsection{\textbf{Hyperparameter Selection}}\label{sec: Hyperparameter Selection}

\begin{table}
\centering
\caption{\textbf{Kappa and mG-Mean metrics comparison, and the Ranks of the methods for each dataset.} (Related to Sec. \ref{sec: effective analysis}.)}
\label{tab:comprehensive-metrics}
\resizebox{\textwidth}{!}{
\begin{tabular}{llrrrrrrrrrrrrrrrr}
\toprule
\textbf{Dataset} & & \textbf{LSH-DynED} & \textbf{ARF} & \textbf{ARFR} & \textbf{BELS} & \textbf{CALMID} & \textbf{CSARF} & \textbf{GH-VFDT} & \textbf{HD-VFDT} & \textbf{KUE} & \textbf{LB} & \textbf{MicFoal} & \textbf{MOOB} & \textbf{MUOB} & \textbf{OBA} & \textbf{ROSE} & \textbf{SRP} \\ \bottomrule
\multirow{2}{*}{Activity}
& $Kappa$ & \textbf{77.49} & 59.94 & 61.37 & 0.21 & 35.09 & 58.92 & 51.12 & 51.12 & 32.66 & 48.05 & 30.27 & 52.52 & 0.00 & 54.49 & 25.34 & 59.93 \\
& $mG-Mean$ & \textbf{88.64} & 83.44 & 82.46 & 0.00 & 66.84 & 80.34 & 57.96 & 57.96 & 57.77 & 77.14 & 60.96 & 60.23 & 85.07 & 52.13 & 41.45 & 84.30 \\
\midrule
\multirow{2}{*}{Balance}
& $Kappa$ & 38.09 & 61.69 & 64.01 & 47.07 & 68.76 & 48.12 & \textbf{69.57} & \textbf{69.57} & 0.00 & 68.80 & 65.56 & 50.34 & 28.26 & 68.67 & 61.92 & 64.75 \\
& $mG-Mean$ & 71.03 & 0.00 & 0.00 & 19.87 & 23.17 & 34.08 & 21.34 & 21.34 & \textbf{79.31} & 23.17 & 0.00 & 11.65 & 33.85 & 0.00  & 16.01 & 0.00 \\
\midrule
\multirow{2}{*}{Connect-4}
& $Kappa$ & 26.19 & 32.23 & 35.46 & 23.39 & 34.79 & 36.98 & 26.08 & 21.10 & 19.69 & 33.68 & 30.53 & 37.90 & 20.02 & 25.84 & \textbf{41.70} & 33.44 \\
& $mG-Mean$ & 51.05 & 61.47 & 53.55 & 22.84 & 52.45 & 47.02 & 41.69 & 49.43 & 50.41 & 55.72 & 51.44 & 47.44 & 36.22 & 56.32 & 48.90  & \textbf{63.75} \\
\midrule
\multirow{2}{*}{Contraceptive}
& $Kappa$ & 37.52 & 27.30 & 24.17 & \textbf{93.38} & 0.00 & 28.81 & 7.03 & 7.03 & 6.09 & 22.71 & 11.86 & 9.38 & 0.79 & 18.98 & 23.73 & 42.34 \\
& $mG-Mean$ & 71.72 & 43.69 & 38.36 & \textbf{96.74} & 73.45 & 44.37 & 52.75 & 52.75 & 67.44 & 61.27 & 50.39 & 54.59 & 73.60 & 59.50 & 43.54 & 72.48 \\
\midrule
\multirow{2}{*}{Covtype}
& $Kappa$ & \textbf{91.13} & 51.98 & 52.28 & 78.69 & 77.96 & 47.69 & 22.83 & 12.03 & 34.99 & 44.10 & 61.44 & 37.89 & 14.02 & 42.56 & 40.57 & 52.65 \\
& $mG-Mean$ & \textbf{95.24} & 85.82 & 84.59 & 82.04 & 82.94 & 62.22 & 57.99 & 70.94 & 75.39 & 79.85 & 85.18 & 66.75 & 4.78 & 76.53 & 80.19 & 85.81 \\
\midrule
\multirow{2}{*}{Crimes}
& $Kappa$ & \textbf{10.88} & 0.36 & 0.36 & 3.54 & 1.29 & 0.14 & 1.14 & 1.82 & 0.99 & 2.74 & 0.02 & 4.35 & 0.00 & 2.27 & 1.52 & 4.40 \\
& $mG-Mean$ & 34.64 & 80.74 & 80.00 & 0.00 & 34.78 & 80.57 & 51.20 & 43.40 & 43.41 & 47.21 & 83.26 & 27.94 & \textbf{92.42} & 53.03 & 41.24 & 16.63 \\
\midrule
\multirow{2}{*}{Ecoli}
& $Kappa$ & 46.38 & 45.66 & 46.08 & \textbf{94.03} & 44.49 & 54.93 & 39.37 & 39.37 & 0.00 & 44.52 & 50.29 & 43.56 & 0.00 & 42.05 & 40.66 & 45.21 \\
& $mG-Mean$ & 86.50 & 0.00 & 0.00 & 0.00 & 0.00 & 0.00 & 0.00 & 0.00 & \textbf{89.87} & 0.00 & 0.00 & 0.00 & \textbf{89.87} & 0.00 & 0.00 & 0.00 \\
\midrule
\multirow{2}{*}{Fars}
& $Kappa$ & 69.41 & 65.60 & \textbf{70.39} & 20.05 & 66.51 & 48.73 & 45.57 & 49.33 & 63.89 & 66.70 & 61.04 & 50.83 & 0.00 & 63.19 & 42.81 & 69.13 \\
& $mG-Mean$ & \textbf{84.01} & 49.48 & 46.95 & 10.84 & 41.95 & 1.31 & 28.73 & 15.55 & 57.43 & 38.93 & 46.58 & 46.61 & 81.50 & 18.72 & 41.85 & 55.96 \\
\midrule
\multirow{2}{*}{Gas}
& $Kappa$ & \textbf{92.03} & 69.70 & 62.52 & 84.34 & 55.25 & 72.34 & 40.55 & 40.55 & 30.42 & 70.68 & 79.39 & 39.12 & 0.00 & 60.21 & 43.97 & 82.98 \\
& $mG-Mean$ & \textbf{96.74} & 36.65 & 27.48 & 87.24 & 16.35 & 36.55 & 9.71 & 9.71 & 22.72 & 27.10 & 37.87 & 8.41 & 51.83 & 29.95 & 19.93 & 39.11 \\
\midrule
\multirow{2}{*}{Glass}
& $Kappa$ & 34.05 & 25.46 & 21.18 & \textbf{89.43} & 21.87 & 26.46 & 21.55 & 21.55 & 0.00 & 21.87 & 35.66 & 25.00 & 0.00 & 21.70 & 30.56 & 29.45 \\
& $mG-Mean$ & 64.52 & 0.00 & 0.00 & 83.40 & 0.00 & 47.39 & 0.00 & 0.00 & \textbf{85.25} & 0.00 & 61.26 & 0.00 & \textbf{85.25} & 0.00 & 47.78 & 58.48 \\
\midrule
\multirow{2}{*}{Hayes-roth}
& $Kappa$ & 10.44 & 18.80 & 23.65 & 6.30 & 37.16 & 25.19 & \textbf{40.59} & \textbf{40.59} & 0.00 & 37.16 & 33.85 & 35.67 & 33.37 & 35.31 & 35.78 & 36.81 \\
& $mG-Mean$ & 61.30 & 48.46 & 53.12 & 27.26 & 59.27 & 50.59 & 61.27 & 61.27 & \textbf{72.83} & 59.27 & 60.53 & 58.44 & 56.42 & 58.85 & 58.13 & 60.28 \\
\midrule
\multirow{2}{*}{Kr-vs-k}
& $Kappa$ & 5.64 & 10.00 & 9.97 & \textbf{94.83} & 10.08 & 10.19 & 8.13 & 8.13 & 8.73 & 10.21 & 9.85 & 8.28 & 8.77 & 10.23 & 9.74 & 10.13 \\
& $mG-Mean$ & 13.22 & 10.48 & 10.48 & \textbf{98.18} & 10.48 & 10.48 & 10.48 & 10.48 & 10.48 & 10.48 & 10.48 & 10.48 & 10.48 & 10.48 & 10.48 & 10.48 \\
\midrule
\multirow{2}{*}{New-thyroid}
& $Kappa$ & 75.58 & 49.43 & 49.43 & \textbf{85.97} & 47.70 & 54.41 & 47.70 & 47.70 & 0.00 & 47.70 & 48.18 & 47.70 & 0.00 & 47.70 & 54.41 & 54.41 \\
& $mG-Mean$ & 78.50 & 0.00 & 0.00 & \textbf{95.97} & 92.26 & 0.00 & 92.26 & 92.26 & 88.69 & 92.26 & 0.00 & 92.26 & 88.69 & 92.26 & 0.00 & 0.00 \\
\midrule
\multirow{2}{*}{Olympic}
& $Kappa$ & 9.89 & 32.27 & 35.57 & 7.35 & 30.31 & 26.16 & 18.24 & 7.11 & 26.05 & 34.93 & 30.37 & 27.03 & 10.74 & 24.17 & \textbf{37.29} & 7.07 \\
& $mG-Mean$ & 22.68 & \textbf{61.94} & 45.89 & 3.70 & 44.89 & 27.80 & 31.56 & 35.33 & 40.52 & 53.39 & 44.78 & 26.06 & 13.16 & 44.42 & 42.47 & 58.32 \\
\midrule
\multirow{2}{*}{Pageblocks}
& $Kappa$ & 21.00 & 31.37 & 25.92 & \textbf{48.48} & 29.07 & 28.91 & 28.14 & 28.14 & 0.00 & 29.07 & 13.53 & 28.74 & 12.68 & 28.72 & 28.16 & 32.71 \\
& $mG-Mean$ & 80.92 & 0.00 & 27.91 & 0.00 & 0.00 & 19.60 & 0.00 & 0.00 & 98.00 & 0.00 & 0.00 & 0.00 & \textbf{98.14} & 0.00 & 0.00 & 0.00 \\
\midrule
\multirow{2}{*}{Poker}
& $Kappa$ & \textbf{95.07} & 0.64 & 0.72 & 49.38 & 51.27 & 0.41 & 0.19 & 0.66 & 2.55 & 1.10 & 1.40 & 5.26 & 0.00 & 0.12 & 9.36 & 0.29 \\
& $mG-Mean$ & \textbf{97.36} & 82.29 & 82.34 & 29.35 & 83.43 & 80.16 & 81.20 & 81.23 & 81.90 & 81.97 & 81.71 & 81.42 & 82.35 & 82.14 & 80.83 & 82.21 \\
\midrule
\multirow{2}{*}{Shuttle}
& $Kappa$ & \textbf{98.82} & 2.72 & 12.88 & 55.31 & 72.78 & -3.80 & 56.89 & 37.72 & 28.20 & 0.43 & 16.10 & 18.78 & 0.00 & 15.26 & 0.86 & 37.39 \\
& $mG-Mean$ & \textbf{99.59} & 95.83 & 92.62 & 15.08 & 64.49 & 7.16 & 20.17 & 29.99 & 92.82 & 91.46 & 95.98 & 70.58 & 96.61 & 78.62 & 84.58 & 96.50 \\
\midrule
\multirow{2}{*}{Tags}
& $Kappa$ & 47.99 & 1.94 & 1.90 & \textbf{85.83} & 51.70 & 1.81 & 22.09 & 23.86 & 23.99 & 18.09 & 35.04 & 29.70 & 0.00 & 16.32 & 10.62 & 7.79 \\
& $mG-Mean$ & 73.11 & 55.71 & 56.43 & \textbf{96.33} & 11.65 & 54.91 & 2.52 & 2.22 & 2.08 & 41.33 & 36.84 & 2.10 & 62.61 & 43.54 & 45.01 & 56.93 \\
\midrule
\multirow{2}{*}{Thyroid-l}
& $Kappa$ & 87.73 & 86.20 & 78.83 & 1.51 & 79.48 & 73.92 & 60.58 & 61.44 & 76.93 & 87.91 & 78.84 & 83.83 & 58.59 & 79.34 & 85.09 & \textbf{87.81} \\
& $mG-Mean$ & \textbf{95.53} & 87.85 & 82.00 & 0.00 & 86.24 & 76.82 & 79.55 & 82.63 & 82.64 & 90.38 & 87.49 & 85.07 & 64.68 & 87.21 & 86.06 & 92.49 \\
\midrule
\multirow{2}{*}{Thyroid-s}
& $Kappa$ & \textbf{66.76} & 41.07 & 29.62 & -2.95 & 18.21 & 42.15 & -0.26 & 0.49 & 0.00 & 16.39 & 64.35 & 28.98 & 34.43 & -0.26 & 54.87 & 38.93 \\
& $mG-Mean$ & \textbf{91.82} & 63.77 & 27.01 & 0.00 & 36.94 & 45.44 & 48.73 & 24.37 & 28.76 & 36.57 & 75.13 & 58.76 & 48.05 & 48.73 & 68.08 & 83.28 \\
\midrule
\multirow{2}{*}{Wine}
& $Kappa$ & 46.37 & 35.94 & 39.46 & \textbf{95.15} & 42.96 & 41.62 & 42.96 & 42.96 & 0.00 & 42.96 & 38.57 & 37.61 & -0.23 & 39.40 & 42.08 & 38.01 \\
& $mG-Mean$ & 79.24 & 57.06 & 59.87 & \textbf{96.38} & 66.82 & 70.78 & 66.82 & 66.82 & 69.21 & 66.82 & 58.58 & 57.72 & 0.00 & 60.21 & 64.90 & 58.12 \\
\midrule
\multirow{2}{*}{Yeast}
& $Kappa$ & 36.66 & 25.31 & 26.68 & 33.64 & 20.63 & 31.28 & 24.20 & 24.20 & 7.91 & 29.29 & 27.07 & 29.59 & 0.00 & 29.94 & 29.05 & \textbf{37.06} \\
& $mG-Mean$ & 66.34 & 39.69 & 54.39 & 0.00 & 42.82 & 0.00 & 42.05 & 42.05 & 56.94 & 43.42 & 39.60 & 63.05 & \textbf{82.63} & 42.88 & 44.16 & 43.78 \\
\midrule
\multirow{2}{*}{Zoo}
& $Kappa$ & \textbf{93.63} & 90.76 & 90.58 & 50.10 & 85.50 & 88.89 & 89.81 & 90.05 & 90.55 & 84.82 & 89.28 & 89.64 & 87.50 & 90.12 & 68.97 & 93.58 \\
& $mG-Mean$ & \textbf{97.19} & 86.41 & 85.91 & 25.72 & 76.90 & 81.48 & 83.92 & 84.12 & 85.62 & 75.66 & 83.61 & 83.01 & 80.20 & 84.38 & 80.15 & 91.41 \\
\midrule
\multirow{2}{*}{ACTIVITY-D1}
& $Kappa$ & \textbf{75.80} & 64.58 & 60.52 & 0.10 & 41.22 & 64.64 & 36.42 & 38.15 & 16.90 & 55.72 & 45.60 & 48.00 & 3.00 & 47.33 & 62.53 & 67.90 \\
& $mG-Mean$ & \textbf{87.95} & 69.36 & 70.87 & 1.58 & 64.13 & 70.46 & 57.14 & 53.34 & 46.43 & 65.10 & 47.39 & 60.41 & 55.51 & 64.91 & 66.19 & 76.03 \\
\midrule
\multirow{2}{*}{CONNECT4-D1}
& $Kappa$ & 24.01 & 32.12 & 30.75 & 23.88 & 33.14 & 37.68 & 22.59 & 22.05 & 15.72 & 33.64 & 29.85 & 31.56 & 16.14 & 25.83 & \textbf{41.45} & 33.61 \\
& $mG-Mean$ & 52.62 & \textbf{63.54} & 61.62 & 27.27 & 50.51 & 51.83 & 35.48 & 43.54 & 47.74 & 54.24 & 50.55 & 42.48 & 33.93 & 57.82 & 49.51 & 61.70 \\
\midrule
\multirow{2}{*}{COVERTYPE-D1}
& $Kappa$ & \textbf{88.56} & 57.12 & 56.23 & 78.03 & 75.23 & 45.28 & 37.07 & 36.09 & 51.84 & 54.52 & 63.23 & 59.33 & 8.43 & 45.25 & 57.82 & 60.49 \\
& $mG-Mean$ & \textbf{91.69} & 84.30 & 83.73 & 84.36 & 85.24 & 62.49 & 53.36 & 47.77 & 77.92 & 82.03 & 80.86 & 69.60 & 8.03 & 76.76 & 80.47 & 83.80 \\
\midrule
\multirow{2}{*}{CRIMES-D1}
& $Kappa$ & 6.91 & 2.21 & 2.48 & 2.84 & 1.85 & 11.58 & 1.41 & 0.27 & 4.22 & 0.59 & 5.32 & 4.61 & 3.81 & 0.01 & 10.87 & \textbf{14.31} \\
& $mG-Mean$ & 32.01 & 65.55 & 51.63 & 11.19 & 33.82 & 28.53 & 48.79 & 48.49 & 41.58 & 71.28 & 29.10 & 25.86 & 22.29 & \textbf{82.64} & 30.64 & 39.98 \\
\midrule
\multirow{2}{*}{DJ30-D1}
& $Kappa$ & 98.23 & 73.48 & 71.92 & 4.71 & 58.69 & 71.25 & 47.58 & 52.86 & 68.10 & 39.67 & 93.22 & 55.66 & 24.84 & 65.52 & 41.06 & \textbf{98.68} \\
& $mG-Mean$ & \textbf{99.06} & 88.66 & 87.66 & 5.82 & 82.76 & 86.76 & 61.68 & 55.04 & 83.82 & 78.17 & 98.07 & 65.90 & 41.63 & 87.93 & 78.51 & 99.03 \\
\midrule
\multirow{2}{*}{GAS-D1}
& $Kappa$ & 77.61 & \textbf{87.07} & 66.19 & 80.45 & 60.06 & 73.63 & 37.19 & 40.59 & 33.79 & 64.50 & 70.90 & 43.81 & 26.17 & 56.79 & 66.63 & 85.21 \\
& $mG-Mean$ & 81.61 & 74.42 & 56.01 & \textbf{88.07} & 43.69 & 52.80 & 53.74 & 63.08 & 44.08 & 54.89 & 57.49 & 66.53 & 51.46 & 53.60 & 60.19 & 74.13 \\
\midrule
\multirow{2}{*}{OLYMPIC-D1}
& $Kappa$ & 6.00 & 29.52 & 27.28 & 6.09 & 26.69 & 25.24 & 8.51 & 4.13 & 22.89 & 27.91& 29.76 & 14.16 & 9.13 & 20.77& \textbf{33.75} & 10.04  \\
& $mG-Mean$ & 24.78 & 61.18 & 55.92 & 13.55 & 50.62 & 33.62 & 30.70 & 25.00 & 49.75 & 54.63& 56.61 & 27.22 & 21.46 & 57.64 & 47.30 & \textbf{62.07} \\
\midrule
\multirow{2}{*}{POKER-D1}
& $Kappa$ & \textbf{57.15} & 22.73 & 16.71 & 29.45 & 56.92 & 15.66 & 12.18 & 16.81 & 37.14 & 28.25 & 49.94 & 40.64 & 2.04 & 10.94 & 36.48 & 28.74 \\
& $mG-Mean$ & 64.37 & 85.10 & 82.23 & 38.85 & 91.71 & 17.48 & 46.85 & 50.10 & 84.99 & 88.80 & 86.34 & 50.63 & 31.63 & \textbf{93.79} & 63.69 & 86.39 \\
\midrule
\multirow{2}{*}{SENSOR-D1}
& $Kappa$ & \textbf{92.35} & 56.64 & 57.15 & 31.00 & 59.55 & 52.49 & 28.51 & 28.66 & 45.80 & 55.17 & 60.62 & 56.46 & 1.84 & 28.72 & 50.39 & 55.79 \\
& $mG-Mean$ & \textbf{94.92} & 79.65 & 81.73 & 44.60 & 90.53 & 70.84 & 55.41 & 55.54 & 83.67 & 90.56 & 72.34 & 78.61 & 35.63 & 89.12 & 78.97 & 68.05 \\
\midrule
\multirow{2}{*}{TAGS-D1}
& $Kappa$ & 47.21 & 42.28 & 42.65 & \textbf{84.11} & 42.99 & 39.86 & 11.74 & 12.84 & 9.87 & 39.34 & 57.14 & 12.01 & 2.37 & 28.48 & 33.88 & 46.36 \\
& $mG-Mean$ & 70.50 & 56.28 & 55.13 & \textbf{92.77} & 40.36 & 48.42 & 7.27 & 7.96 & 20.65 & 60.39 & 49.71 & 23.03 & 27.21 & 62.68 & 47.02 & 59.54 \\
\bottomrule
\multirow{2}{*}{Avg. Mean}
& $Kappa$ & \textbf{54.32} & 40.43 & 39.24 & 45.03 & 43.61 & 38.84 & 30.52 & 29.67 & 23.03 & 38.28 & 42.97 & 36.00 & 12.32 & 34.73 & 37.56 & 44.47 \\
& $mG-Mean$ & \textbf{72.74} & 56.33 & 53.87 & 39.36 & 51.56 & 44.92 & 42.19 & 41.93 & 61.21 & 56.17 & 53.94 & 46.15 & 52.95 & 54.75 & 49.95 & 58.21 \\
\bottomrule
\multirow{2}{*}{Rank}
& $Kappa$ & \textbf{4.70} & 7.17 & 7.74 & 7.61 & 7.08 & 7.67 & 11.26 & 11.09 & 11.83 & 7.56 & 6.88 & 8.38 & 14.44 & 9.70 & 7.58 & 5.33 \\
& $mG-Mean$ & \textbf{3.88} & 6.65 & 7.59 & 10.62 & 8.30 & 10.38 & 10.98 & 10.91 & 7.97 & 7.48 & 8.15 & 10.68 & 9.09& 7.76 & 9.79 & 5.76 \\
  \bottomrule
\end{tabular}}
\end{table}

\begin{figure}
    \centering
    \begin{subfigure}[b]{0.49\textwidth} 
    \centering
        \includegraphics[width=\textwidth]{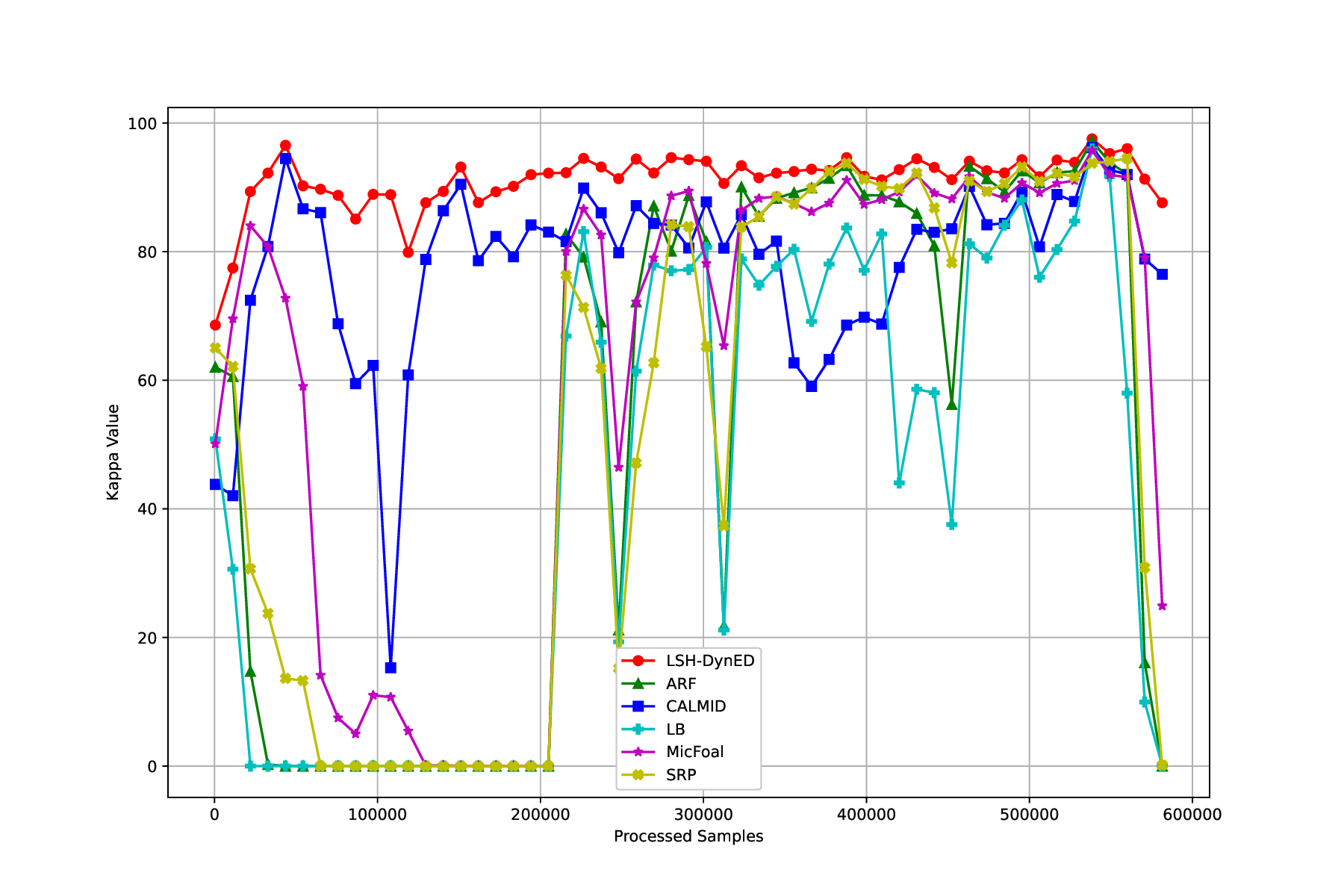}
        \caption{Covtype.}
        \label{fig:covtype}
    \end{subfigure}
    \hfill
    \begin{subfigure}[b]{0.49\textwidth} 
    \centering
        \includegraphics[width=\textwidth]{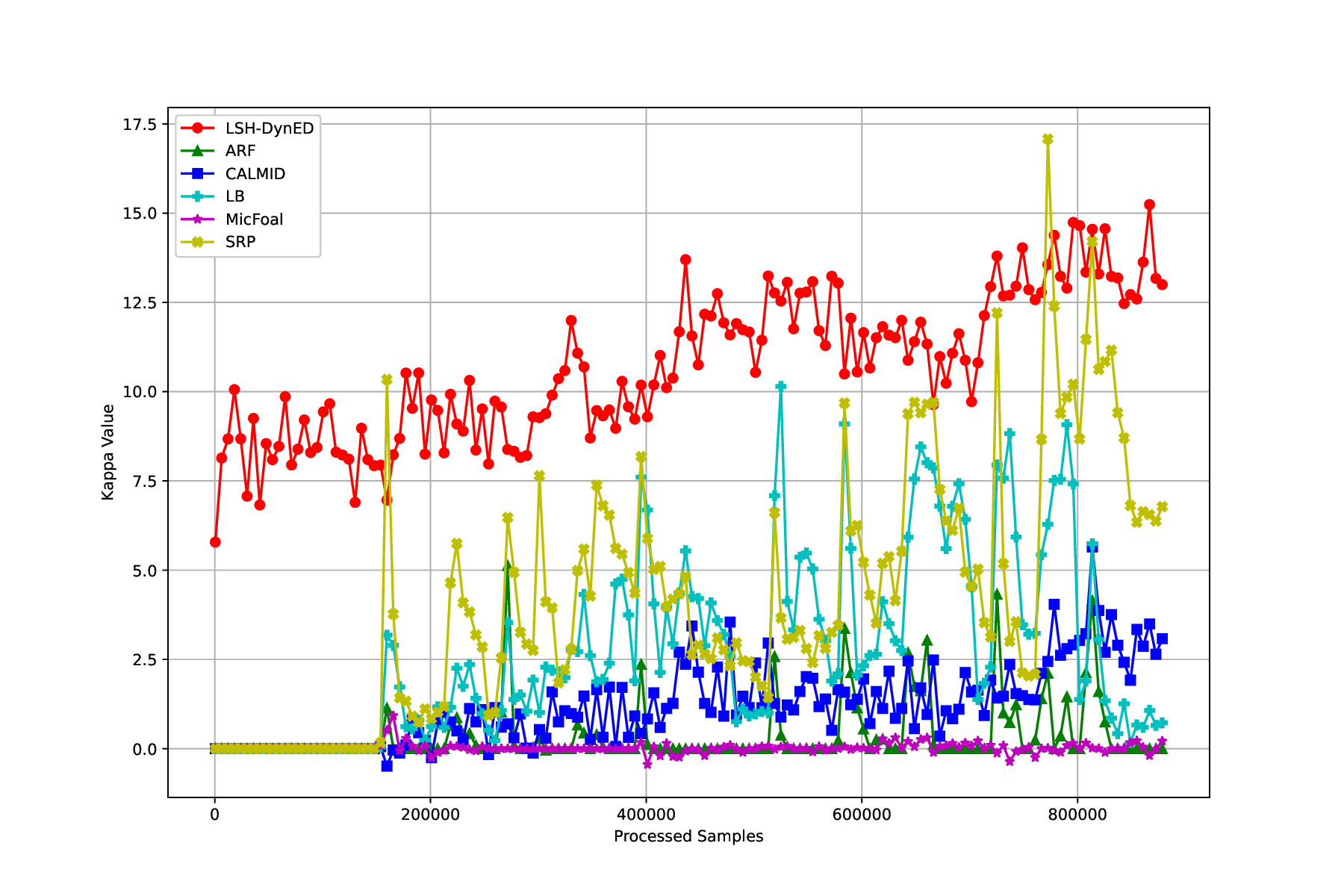}
        \caption{Crimes.}
        \label{fig:crimes}
    \end{subfigure}

    \vspace{\floatsep} 

    \begin{subfigure}[b]{0.49\textwidth} 
    \centering
        \includegraphics[width=\textwidth]{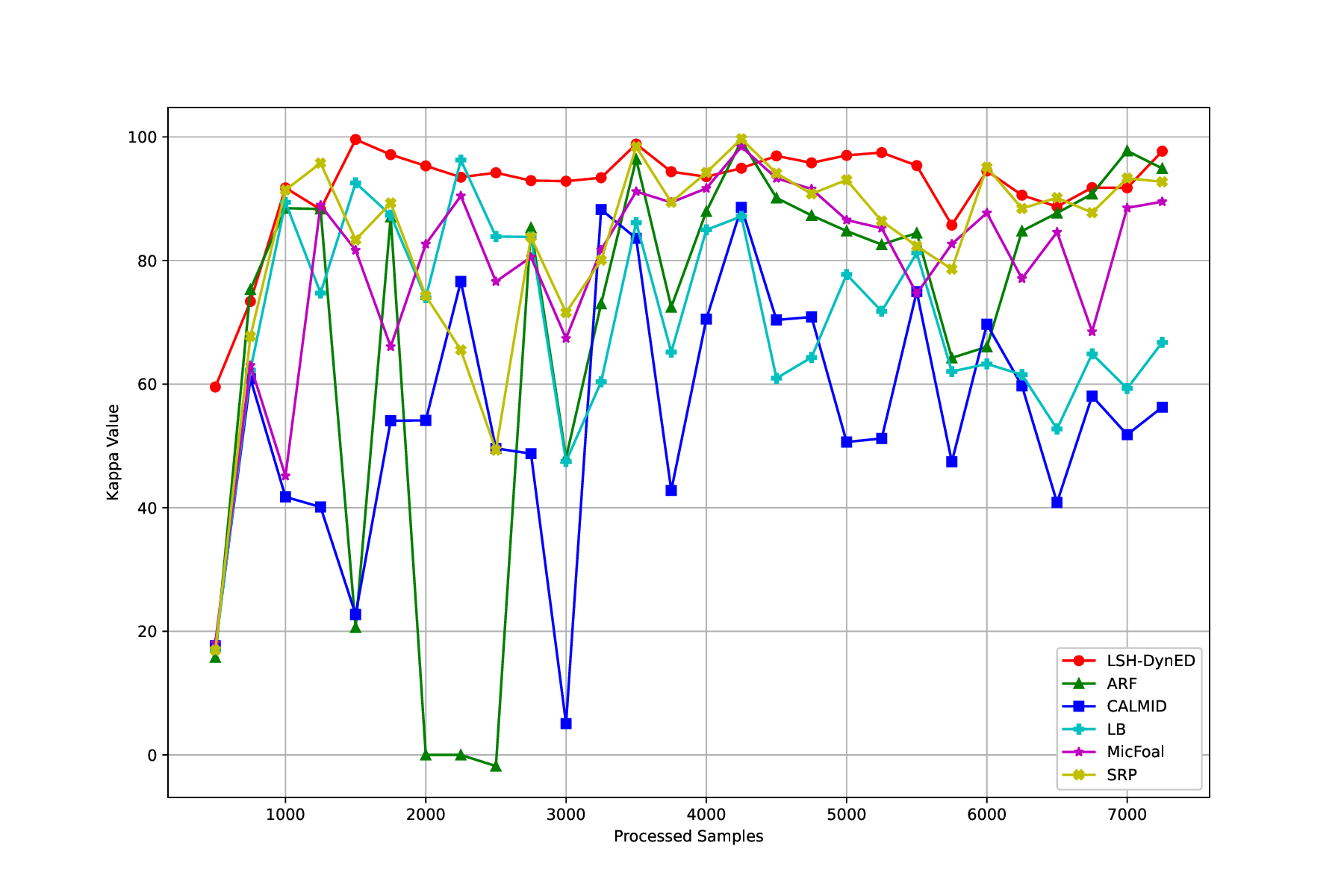}
        \caption{Gas.}
        \label{fig:gas}
    \end{subfigure}
    \hfill
    \begin{subfigure}[b]{0.49\textwidth} 
    \centering
        \includegraphics[width=\textwidth]{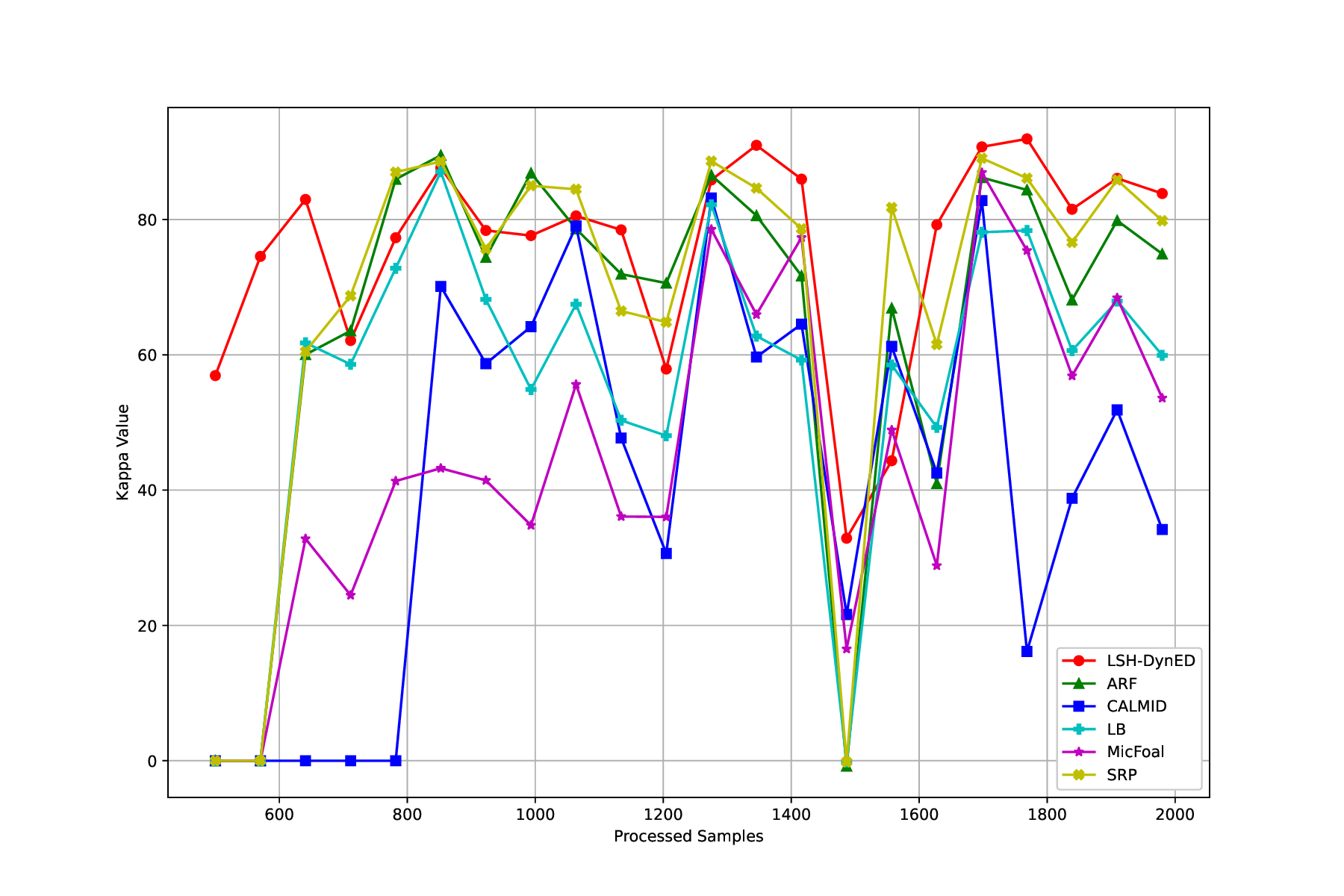}
        \caption{ACTIVITY-D1.}
        \label{fig:activity-d1-per}
    \end{subfigure}

    \vspace{\floatsep} 

    \begin{subfigure}[b]{0.49\textwidth} 
    \centering
        \includegraphics[width=\textwidth]{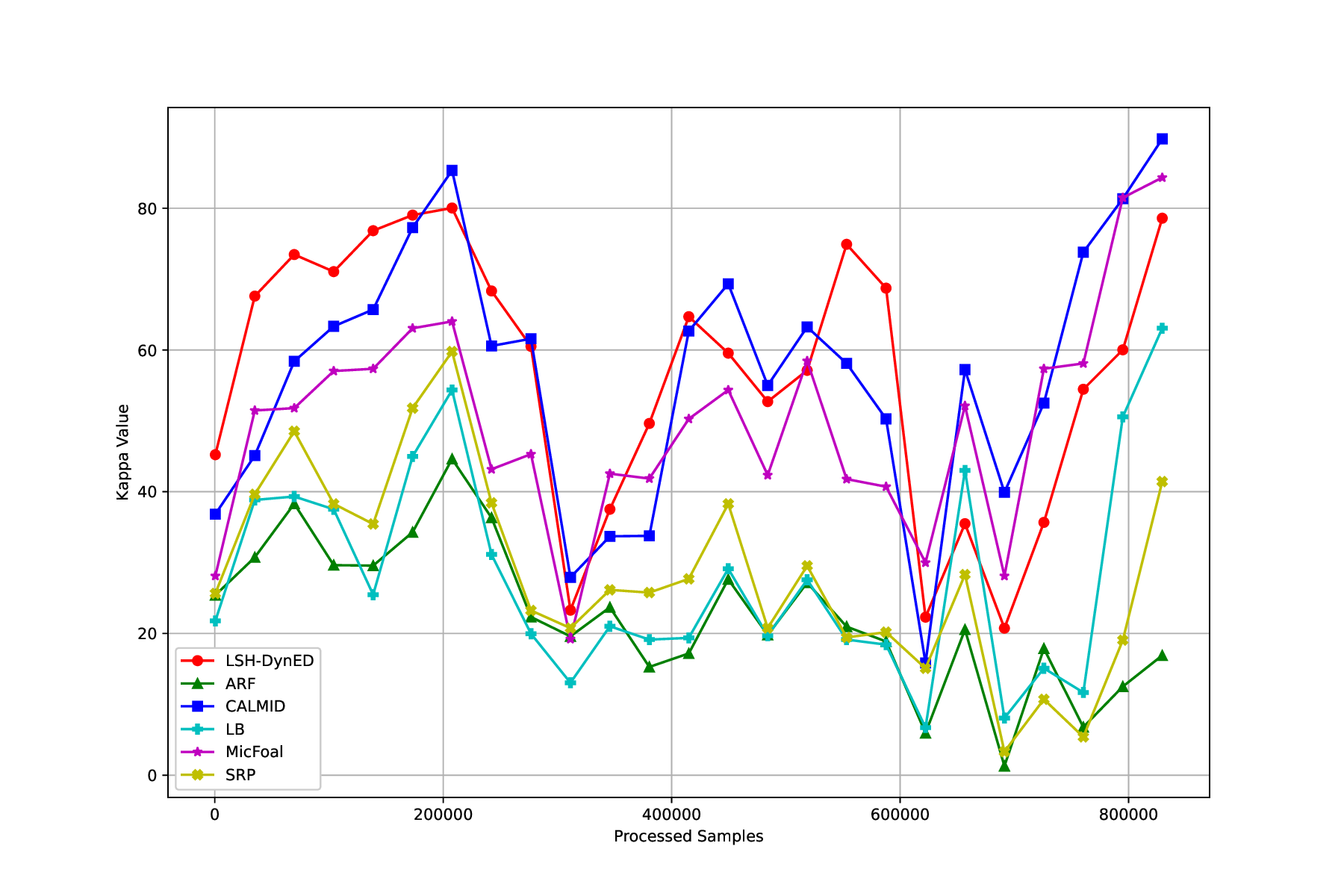}
        \caption{POKER-D1.}
        \label{fig:poker-d1-per}
    \end{subfigure}
    \hfill
    \begin{subfigure}[b]{0.49\textwidth} 
    \centering
        \includegraphics[width=\textwidth]{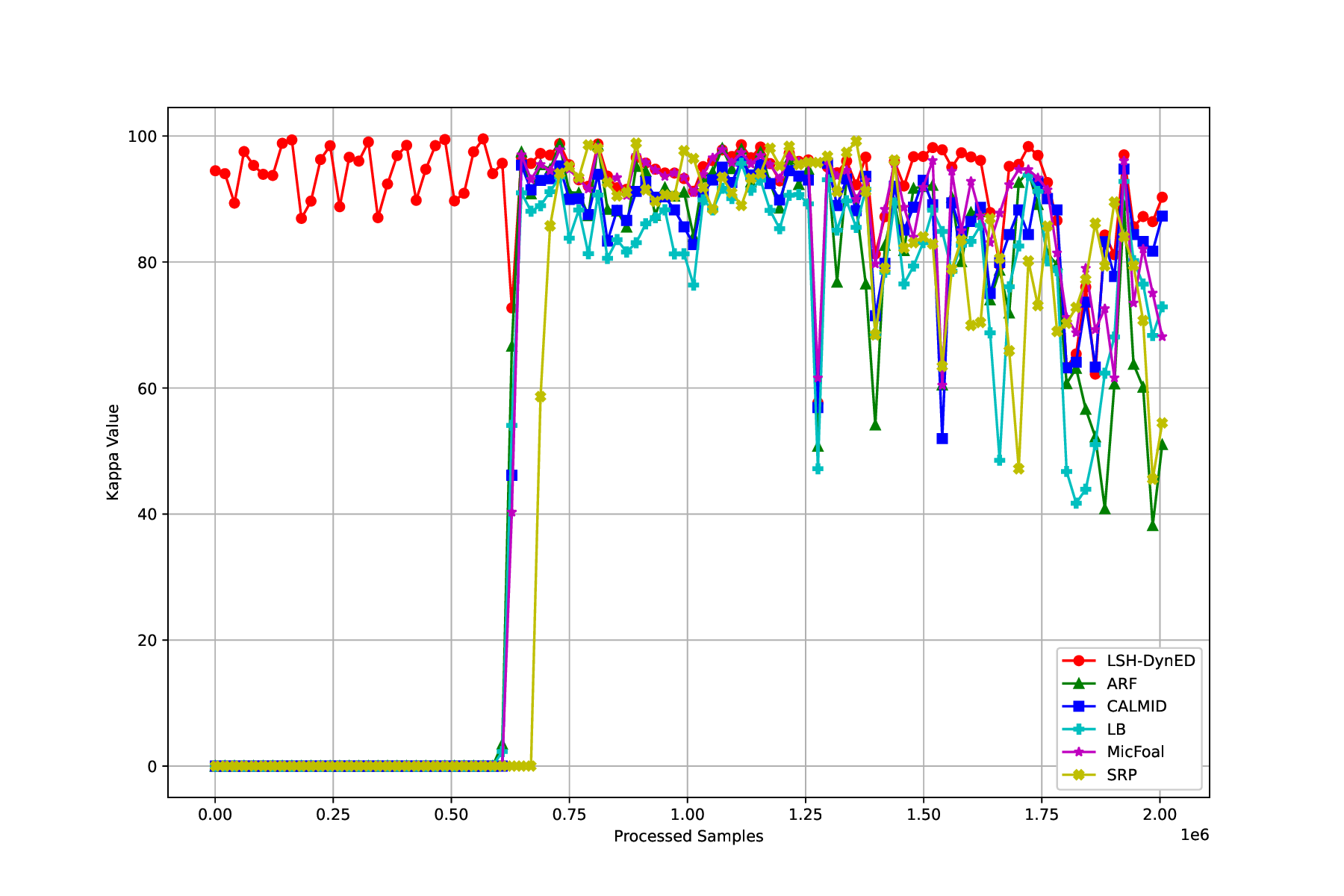}
        \caption{SENSOR-D1.}
        \label{fig:sensor-d1-per}
    \end{subfigure}

    \caption{\textbf{Kappa metric results of three real and three semi-synthetic datasets. The results are plotted among the top 5 ranked methods in the Kappa metric: LSH-DynED, ARF, CALMID, LB, MicFoal, and SRP.} (Related to Sec. \ref{sec: effective analysis}.)}
    \label{fig: kappa_fig}
\end{figure}
In our evaluation of the proposed method, we focus on four key parameters to assess how they impact overall performance using the Kappa metric. Note that we use the Kappa metric for this sensitivity analysis as it offers a stable and comprehensive view of performance across a wide range of parameter combinations. Unlike mG-Mean, which can swing sharply due to variations in a single class, Kappa provides a more balanced and dependable signal for identifying generally optimal settings \cite{aguiar2023survey, wang2020auc}. To investigate different values for each parameter, we conduct this analysis over four semi-synthetic datasets that have a sufficient number of samples and a variety of features: 'ACTIVITY-D1', 'DJ30-D1', 'GAS-D1', and 'TAGS-D1'. The parameters we examine are:

\begin{itemize}
    \item $\mathit{S_{slc}}$: The number of active components for classification, three values: 5, 10, and 15.
    \item $n_{train}$: The number of samples used for training, three values: 20, 50, and 100.
    \item $n_{test}$: The number of samples used for testing purposes, three values: 50, 100, and 200.
    \item $n_v$: The number of hyperplanes used in the undersampling process, five values: 2, 3, 4, 5, and 6.
\end{itemize}
By adjusting these parameters, we aim to understand their influence on the method's performance and identify optimal configurations. These values produced 135 possible combinations of parameters for each dataset, totaling 540 results across four datasets.


The evaluation results are presented in Figure \ref{fig: hyper-param-selec}, and full results for each set of combinations are available on GitHub \footnote{\href{https://github.com/soheilabadifard/LSH-DynED}{LSH-DynED}}. By closely analyzing the results, we determined that the optimal values for our key parameters are as follows:
\noindent Our proposed method tends to yield a higher Kappa score with ten active components, 20 training samples, 50 test samples, and five hyperplanes, marked as 'Selected Parameters' in Figure \ref{fig: hyper-param-selec} for the undersampling part with LSH-RHP. These values are also presented in Table \ref{tab: Symbols}.

\subsection{Effectiveness and Statistical Comparison}\label{sec: effective analysis}
Table \ref{tab:comprehensive-metrics} and Figure \ref{fig: kappa_fig} present the experimental evaluation results of various data stream classification models, focusing on their capability to effectively handle imbalanced multi-class non-stationary data streams. This included adapting to concept drift and maintaining accuracy across diverse class distributions. Compared to the other methods, the proposed approach has the best average performance in both the Kappa and mG-Mean metrics and achieves the lowest (best rank), as shown in the last two rows of Table \ref{tab:comprehensive-metrics}.

In the sections that follow, we first compare the methods based on Kappa and mG-Mean scores in Section \ref{sec: General Comparison}. Then, we statistically analyze the effectiveness of these methods in Section \ref{sec: statistical analysis}.


\subsubsection{\textbf{General-Purpose vs. Imbalance-Specific Methods.}} \label{sec: General Comparison}
\paragraph*{\textbf{General-Purpose Methods (GPM)}}
General-Purpose Methods (GPM), presented in Table \ref{table: methods}, are designed to handle balanced scenarios but are not explicitly tailored for imbalance. These models show varying degrees of success across different datasets. For example, in the 'Poker' dataset, OBA and LB struggle with high imbalance ratios, reflected in their lower Kappa (OBA: 0.12, LB: 1.10) and mG-Mean scores. This variability highlights GPM methods' challenges in highly imbalanced environments compared to specialized approaches. On the other hand, BELS and SRP perform effectively on the Kappa metric and KUE on the mG-Mean when compared to other GPM approaches and, in certain circumstances, outperform ISM methods.
\paragraph*{\textbf{Imbalance-Specific Methods (ISM)}}Imbalance-Specific Methods (ISM) are explicitly designed to address class imbalances. LSH-DynED is within this category and demonstrates superior performance across multiple datasets. This is particularly noticeable in its ability to manage multi-class imbalances and adapt to concept drift, as seen in its high average Kappa (58.23) and mG-Mean (72.78) scores. Its robust performance is evident in challenging datasets like 'Poker' (Kappa: 95.11) and 'Shuttle' (Kappa: 99.32), and it shows adaptability in datasets with concept drift such as 'ACTIVITY-D1' (Kappa: 77.68), 'COVERTYPE-D1' (Kappa: 89.40), and 'SENSOR-D1' (Kappa: 92.26). Furthermore, it is evident how well LSH-DynED performs on datasets like "Activity," "Crimes," and "Fars" that have high imbalance ratios, indicating its resilience in such challenging environments. Thus, the integration of LSH-RHP in the proposed method helps it effectively undersample the majority classes and maintain a balanced training set, enhancing its performance.

\begin{figure}[!t]
    \centering
    \begin{tabular}{cc}
        \includegraphics[width=0.45\textwidth]{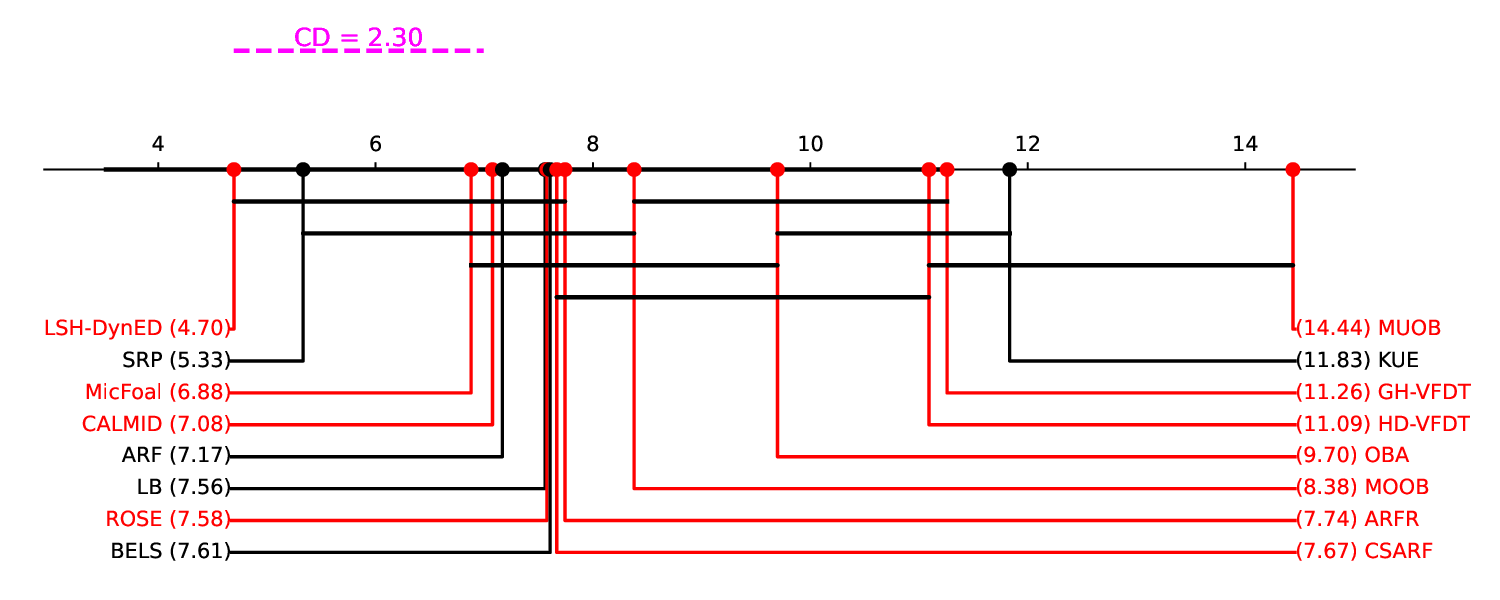} &
        \includegraphics[width=0.45\textwidth]{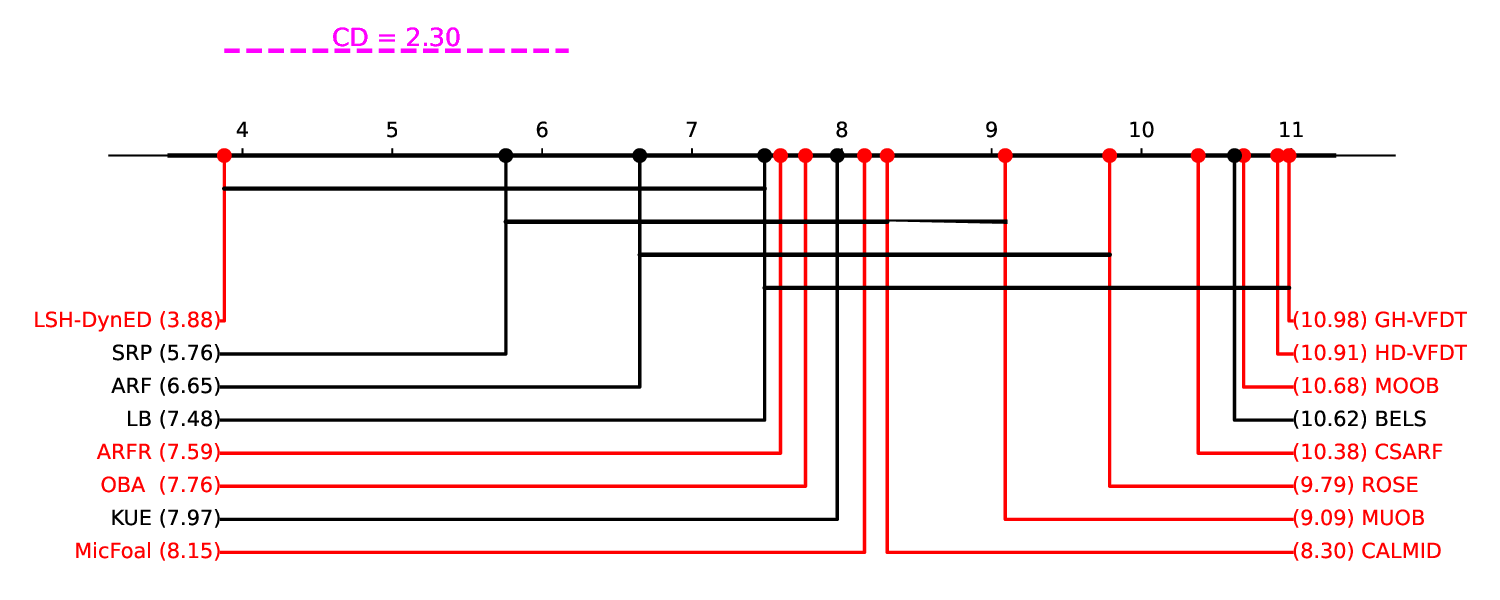} \\
        (a) Kappa. & (b) mG-Mean. \\
    \end{tabular}
    \caption{\textbf{Critical distance (CD) diagram based on the ranks of the methods (Table \ref{tab:comprehensive-metrics}) for each evaluation metric (CD = 2.30).} The color red is for ISM methods, and black is for GPM methods. (Related to Sec. \ref{sec: statistical analysis}.)}
    \label{fig: stat}
\end{figure}
\begin{figure}[t]
    \centering
    \includegraphics[width=0.6\textwidth]{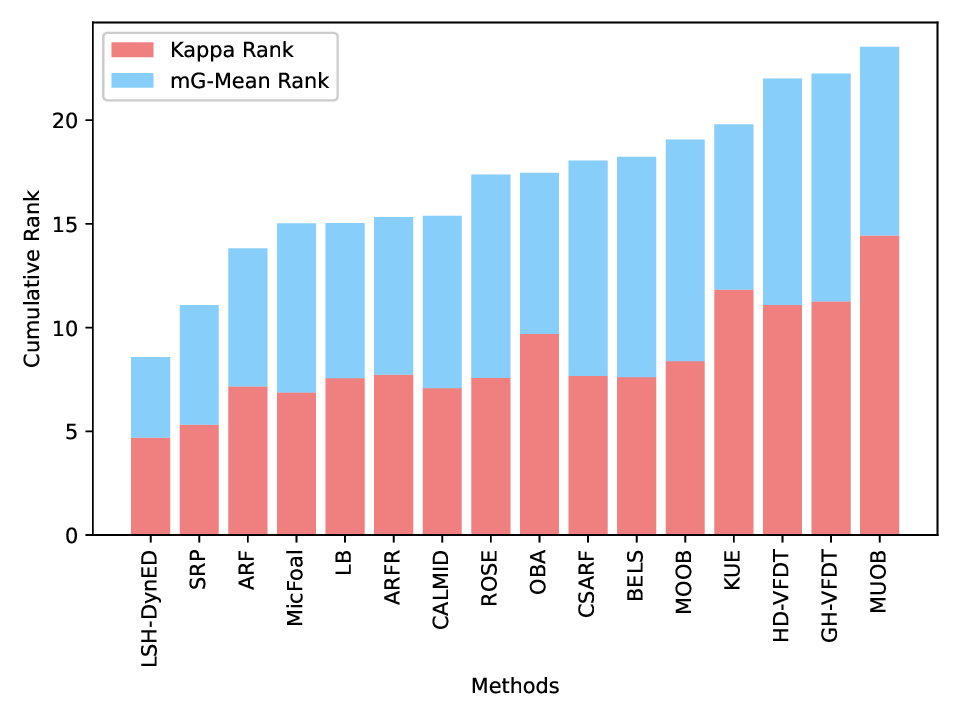}
    \caption{\textbf{Cumulative ranking across two metrics (lower is better), models sorted in ascending order.} (Related to Sec. \ref{sec: statistical analysis}.)}
    \label{fig:com-rank}
\end{figure}

\subsubsection{\textbf{Statistical Analysis of Effectiveness Performances}}\label{sec: statistical analysis}
We use the Friedman Test ($\alpha$ = 0.05) to dismiss the null hypothesis, followed by implementing the Nemenyi posthoc test (critical distance (CD) = 2.30) \cite{demvsar2006statistical}. The diagrams in Figure \ref{fig: stat} indicate that the proposed method obtains the best ranking in both two measures of effectiveness.

Regarding the Kappa metric, LSH-DynED is in the same critical distance as SRP and MicFoal and statistically significantly outperforms the rest of the methods. In the mG-Mean metric, LSH-DynED is in the same group as SRP, ARF, and LB and demonstrates statistically significant superiority over all other methods. The Figure \ref{fig:com-rank} shows the cumulative ranking of methods, with LSH-DynED having the best overall ranking. 

\subsection{Impact of Class Characteristics}\label{Sec: impact of class characteristics}

In this section, we analyze the effects of different factors. First, we look at how the number of classes affects the model's performance in Section \ref{sec: Impact of Dataset Characteristics}. Next, we examine the impact of dynamic imbalance ratios in Section \ref{sec: Impact of Dynamic Imbalance Ratios}. Finally, we conduct an ablation analysis in Section \ref{sec: ablation study}.

\begin{figure}[t]
\centering
    \begin{tabular}{c c}
         \includegraphics[width=0.45\textwidth]{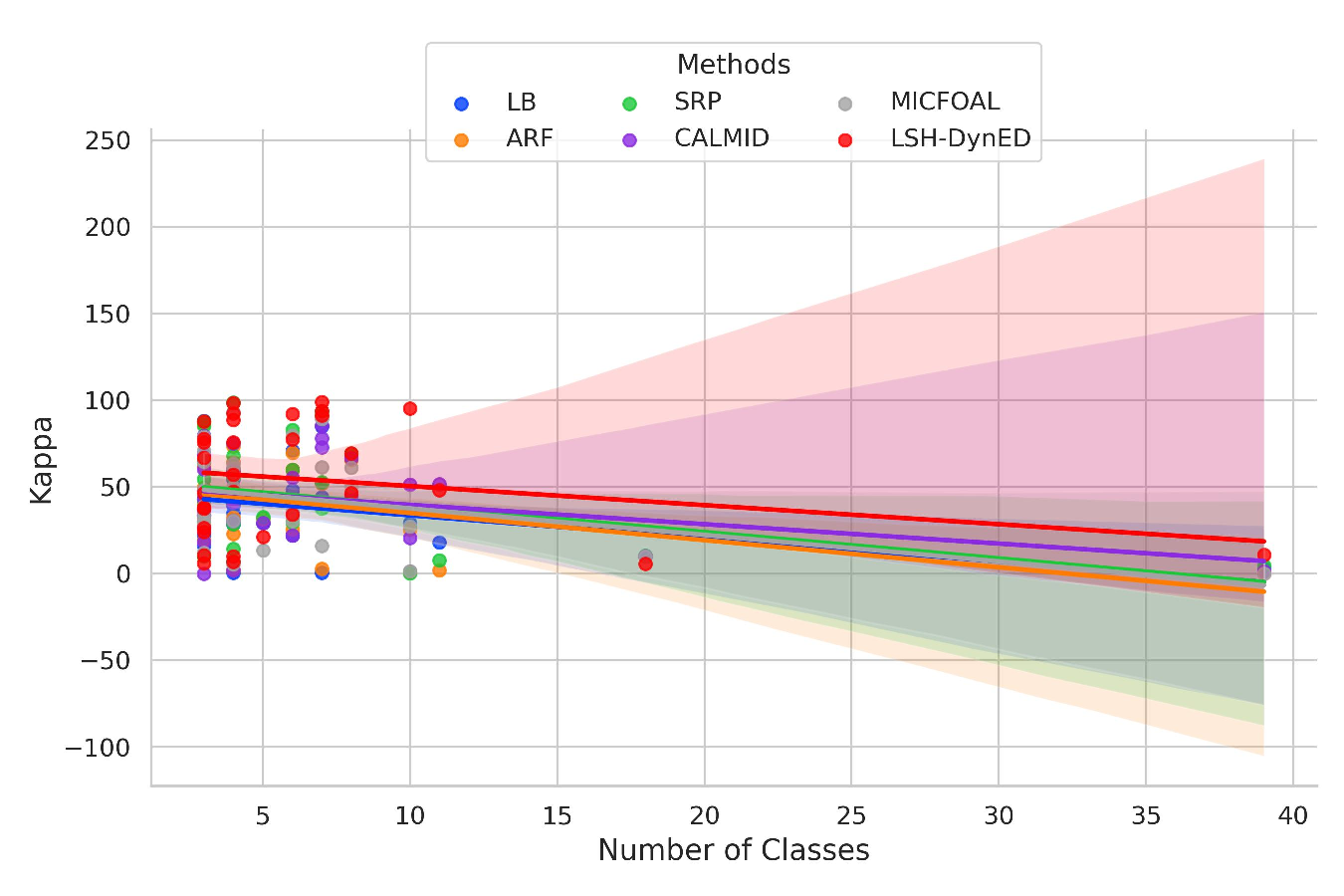} &
         \includegraphics[width=0.45\textwidth]{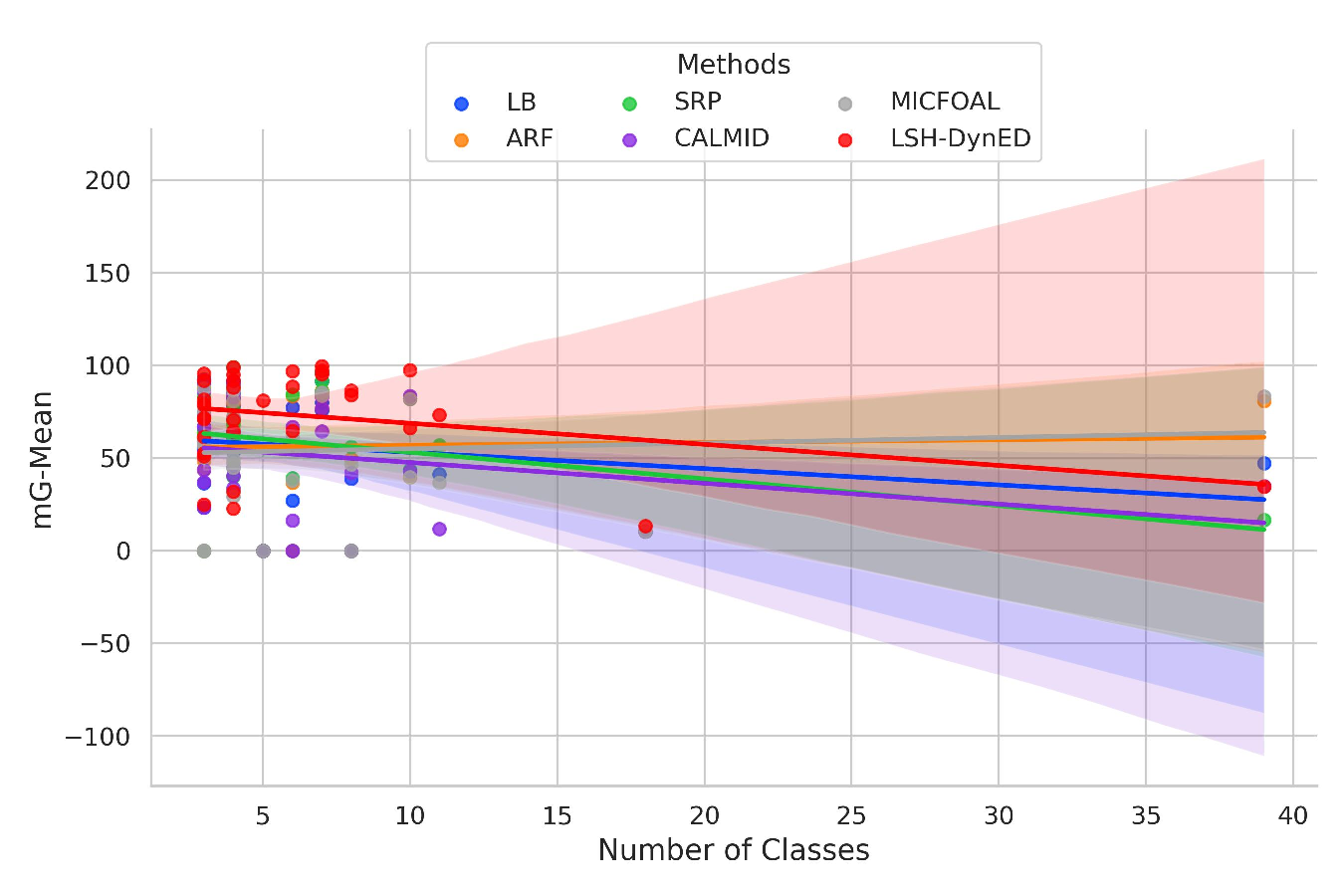}\\
         (a) Kappa. &
         (b) mG-Mean.\\
    \end{tabular}
    \caption{\textbf{Effect of the number of classes over the performance of top 5 ranked methods in terms of Kappa and mG-Mean metrics: LSH-DynED, ARF, CALMID, LB, MicFoal, and SRP.} (Related to Sec. \ref{sec: Impact of Dataset Characteristics}.)}
    \label{fig: class_effects}
\end{figure}
\begin{table}[t]
\centering
        \caption{\textbf{Correlation and p-value between the number of classes and both Kappa Metric and mG-Mean.} The case with $p<0.05$ is statistically significant. (Related to Sec. \ref{sec: Impact of Dataset Characteristics}.)}
        \begin{tabular}{lcccc}
            \toprule
            \multirow{2}{*}{\textbf{Methods}} & \multicolumn{2}{c}{\textbf{Kappa}} & \multicolumn{2}{c}{\textbf{mG-Mean}} \\
            \cmidrule(lr){2-3} \cmidrule(lr){4-5}
            & \textbf{Correlation} & \textbf{p-value} & \textbf{Correlation} & \textbf{p-value} \\
            \midrule
            LSH-DynED    & -0.23  & 0.20 & -0.31 & 0.08 \\
            ARF          & -0.40  & \textbf{0.02} & 0.03  & 0.86 \\
            ARFR         & -0.39  & \textbf{0.02} & 0.09  & 0.63 \\
            BELS         & -0.02  & 0.91 & -0.08 & 0.64 \\
            CALMID       & -0.32  & 0.07 & -0.26 & 0.14 \\
            CSARF        & -0.41  & \textbf{0.02} & 0.08  & 0.64 \\
            GH-VFDT       & -0.28  & 0.12 & -0.10 & 0.57 \\
            HD-VFDT       & -0.27  & 0.14 & -0.14 & 0.43 \\
            KUE          & -0.15  & 0.40 & -0.25 & 0.17 \\
            LB           & -0.38  & \textbf{0.03} & -0.21 & 0.23 \\
            MicFoal      & -0.40  & \textbf{0.02} & 0.07  & 0.70 \\
            MOOB          & -0.35  & \textbf{0.04} & -0.23 & 0.21 \\
            MUOB          & -0.19  & 0.29 & 0.28  & 0.12 \\
            OBA          & -0.31  & 0.08 & -0.17 & 0.34 \\
            ROSE         & -0.51  & \textbf{0.00} & -0.15 & 0.41 \\
            SRP          & -0.38  & \textbf{0.03} & -0.32 & 0.07 \\
            \bottomrule
        \end{tabular}
        \label{tab: correlation_table}
\end{table}

\begin{table}[t]
\centering
\caption{\textbf{The Kappa and mG-Mean scores of the top 5 ranked methods according to the Kappa metric for semi-synthetic datasets with dynamic imbalance ratios.} (Related to Sec. \ref{sec: Impact of Dynamic Imbalance Ratios}.)}
\label{tab: compare-dynamic}
\begin{tabular}{llrrrrrr}
\toprule
\textbf{Dataset} & & \textbf{LSH-DynED} & \textbf{ARF} & \textbf{CALMID} & \textbf{LB} & \textbf{MicFoal} & \textbf{SRP} \\ \bottomrule
\multirow{2}{*}{ACTIVITY-D1}
& $Kappa$ & \textbf{75.80} & 64.58 & 41.22 & 55.72 & 45.60 & 67.90 \\
& $mG-Mean$ & \textbf{87.95} & 69.36 & 64.13 & 65.10 & 47.39 & 76.03 \\
\midrule
\multirow{2}{*}{CONNECT4-D1}
& $Kappa$ & 24.01 & 32.12 & 33.14 & \textbf{33.64} & 29.85 & 33.61 \\
& $mG-Mean$ & 52.62 & \textbf{63.54} & 50.51 & 54.24 & 50.55 & 61.70 \\
\midrule
\multirow{2}{*}{COVERTYPE-D1}
& $Kappa$ & \textbf{88.56} & 57.12 & 75.23 & 54.52 & 63.23 & 60.49 \\
& $mG-Mean$ & \textbf{91.69} & 84.30 & 85.24 & 82.03 & 80.86 & 83.80 \\
\midrule
\multirow{2}{*}{CRIMES-D1}
& $Kappa$ & 6.91 & 2.21 & 1.85 & 0.59 & 5.32 & \textbf{14.31} \\
& $mG-Mean$ & 32.01 & 65.55 & 33.82 & \textbf{71.28} & 29.10 & 39.98 \\
\midrule
\multirow{2}{*}{DJ30-D1}
& $Kappa$ & 98.23 & 73.48 & 58.69 & 39.67 & 93.22 & \textbf{98.68} \\
& $mG-Mean$ & \textbf{99.06} & 88.66 & 82.76 & 78.17 & 98.07 & 99.03 \\
\midrule
\multirow{2}{*}{GAS-D1}
& $Kappa$ & 77.61 & \textbf{87.07} & 60.06 & 64.50 & 70.90 & 85.21 \\
& $mG-Mean$ & \textbf{81.61} & 74.42 & 43.69 & 54.89 & 57.49 & 74.13 \\
\midrule
\multirow{2}{*}{OLYMPIC-D1}
& $Kappa$ & 6.00 & 29.52 & 26.69 & 27.91 & 29.76 & \textbf{33.75} \\
& $mG-Mean$ & 24.78 & 61.18 & 50.62 & 54.63 & 56.61 & \textbf{62.07} \\
\midrule
\multirow{2}{*}{POKER-D1}
& $Kappa$ & \textbf{57.15} & 22.73 & 56.92 & 28.25 & 49.94 & 28.74 \\
& $mG-Mean$ & 64.37 & 85.10 & \textbf{91.71} & 88.80 & 86.34 & 86.39 \\
\midrule
\multirow{2}{*}{SENSOR-D1}
& $Kappa$ & \textbf{92.35} & 56.64 & 59.55 & 55.17 & 60.62 & 55.79 \\
& $mG-Mean$ & \textbf{94.92} & 79.65 & 90.53 & 90.56 & 72.34 & 68.05 \\
\midrule
\multirow{2}{*}{TAGS-D1}
& $Kappa$ & 47.21 & 42.28 & 42.99 & 39.34 & \textbf{57.14} & 46.36 \\
& $mG-Mean$ & \textbf{70.50} & 56.28 & 40.36 & 60.39 & 49.71 & 59.54 \\
\midrule
\multirow{2}{*}{Avg. Mean}
& $Kappa$ & \textbf{57.36} & 46.77 & 45.63 & 39.93 & 50.56 & 50.11 \\
& $mG-Mean$ & 69.95 & \textbf{72.80} & 63.33 & 70.01 & 62.84 & 71.07 \\
\bottomrule
\end{tabular}
\end{table}

\subsubsection{\textbf{Impact of Number of Classes}}\label{sec: Impact of Dataset Characteristics}

The characteristics of datasets in terms of the number of classes may significantly affect the performance of the models.
Table \ref{tab: correlation_table} and Figure \ref{fig: class_effects} present the correlation (Pearson correlation coefficient) \cite{cohen2009pearson} of each method's performance, Kappa, and mG-Mean metrics against the number of classes on datasets available in Table \ref{table: datasets}. Based on our observations on these datasets, all methods show negative correlations between Kappa and the number of classes, indicating that the Kappa value tends to decrease as the number of classes increases. Several methods, such as ARF, ARFR, CSARF, LB, MicFoal, MOOB, ROSE, and SRP, show a statistically significant negative correlation with the Kappa metric ($p-value <0.05$). Meanwhile, ROSE with $-0.51$ has the strongest and most significant negative correlation with the Kappa metric. On the other hand, the correlations between mG-Mean and the number of classes vary, with both positive and negative values, and most correlations with mG-Mean are not statistically significant.

This analysis shows that the number of classes significantly impacts Kappa, with all methods showing a negative correlation. In contrast, the impact of the number of classes on mG-Mean is mixed and generally not significant, indicating varying degrees of correlation across different methods. In the case of LSH-DynED, a negative correlation between Kappa and the number of classes suggests that as the number of classes increases, Kappa tends to decrease. However, the $p-value$ of $0.20$ indicates that this correlation is not statistically significant.

\subsubsection{\textbf{Impact of Dynamic Imbalance Ratio of Classes}}\label{sec: Impact of Dynamic Imbalance Ratios}
\begin{figure}[t]
    \centering
    \begin{subfigure}[b]{0.45\textwidth}
    \centering
        \includegraphics[width=\textwidth]{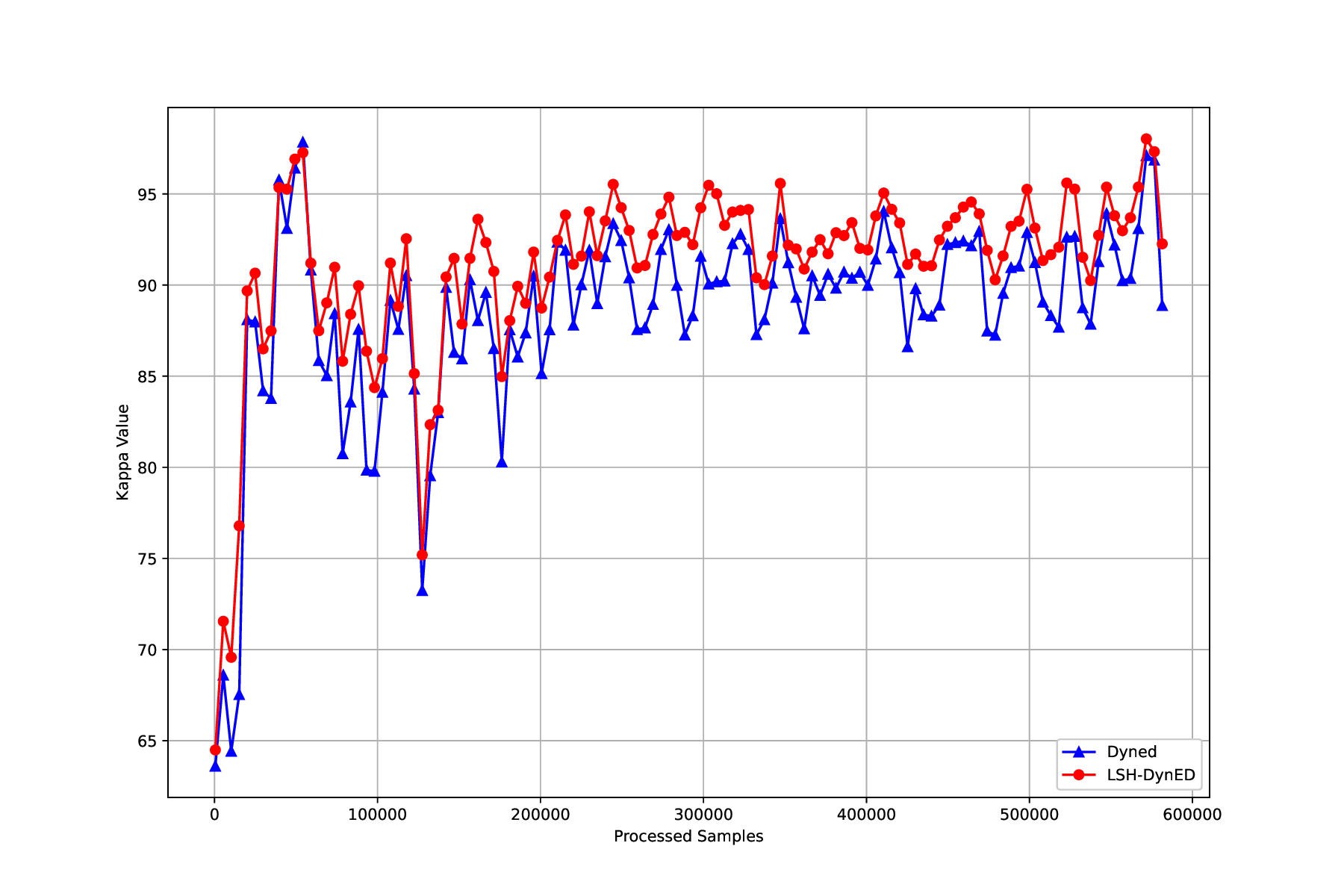}
        \caption{Covtype.}
        \label{fig:covtype-abl}
    \end{subfigure}
    \begin{subfigure}[b]{0.45\textwidth}
    \centering
        \includegraphics[width=\textwidth]{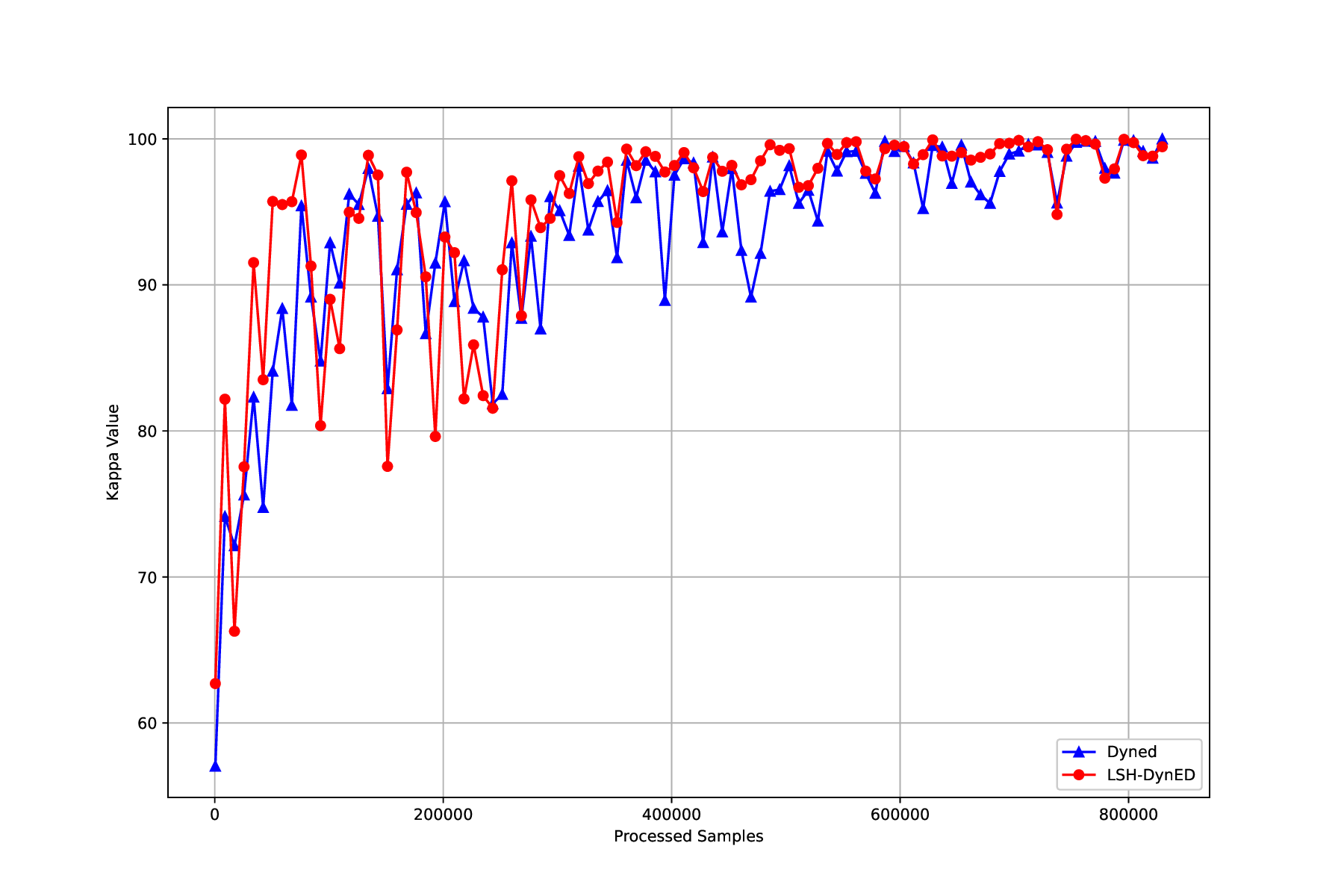}
        \caption{Poker.}
        \label{fig:poker-abl}
    \end{subfigure}

    \caption{\textbf{Ablation analysis: Kappa Metric results of  LSH-DynED vs. DynED.} (Related to Sec. \ref{sec: ablation study}.)}
    \label{fig: diag}
\end{figure}
\begin{table}
\centering
\caption{\textbf{Ablation analysis: LSH-DynED vs. DynED across various datasets.} (Related to Sec. \ref{sec: ablation study}.)}
\label{tab: ablation}
\begin{tabular}{lcccc}
\toprule
\multirow{2}{*}{\textbf{Dataset}} & \multicolumn{2}{c}{\textbf{Kappa}} & \multicolumn{2}{c}{\textbf{mG-Mean}} \\
\cmidrule(lr){2-3} \cmidrule(lr){4-5}\\
 & \textbf{LSH-DynED} & \textbf{DynED} & \textbf{LSH-DynED} & \textbf{DynED} \\
\midrule
Activity      & \textbf{77.49} & 74.35 & \textbf{88.64} & 47.13 \\
Connect-4     & \textbf{26.19} & 22.50 & \textbf{51.05} & 17.04 \\
Covtype       & \textbf{91.13} & 88.64 & \textbf{95.24} & 62.45 \\
Ecoli         & 46.38 & \textbf{61.85} & \textbf{86.50} & 0.00 \\
Gas           & \textbf{92.03} & 89.65 & \textbf{96.74} & 42.31 \\
Poker         & \textbf{95.07} & 93.85 & \textbf{97.36} & 54.51 \\
Shuttle       & \textbf{98.82} & 98.78 & \textbf{99.59} & 54.79 \\
Thyroid       & \textbf{87.73} & 86.07 & \textbf{95.53} & 87.35 \\
Thyroid-s     & \textbf{66.76} & 54.76 & \textbf{91.82} & 31.08 \\
Zoo           & \textbf{93.63} & 93.24 & \textbf{97.19} & 89.03 \\
ACTIVITY-D1   & \textbf{75.80} & 70.73 & \textbf{87.95} & 55.95 \\
CONNECT4-D1   & \textbf{24.10} & 19.60 & \textbf{52.62} & 17.61 \\
COVERTYPE-D1  & \textbf{88.56} & 85.98 & \textbf{91.69} & 69.00 \\
DJ30-D1       & 98.23 & \textbf{98.66} & \textbf{99.06} & 98.49 \\
TAGS-D1       & \textbf{47.21} & 41.09 & \textbf{70.50} & 18.47 \\
\midrule
Avg. Mean     & \textbf{73.94} & 71.98 & \textbf{86.76} & 49.68 \\
\bottomrule
\end{tabular}
\end{table}
Table \ref{table: datasets} displays the characteristics of the datasets we evaluated, which include dynamic imbalance ratios. The last ten (with suffix 'D1') are explicitly made for this purpose \cite{korycki2020online}, and the imbalance ratio over time is presented in Figures \ref{fig:activity-d1}, \ref{fig:connect4-d1}, and \ref{fig:covertype-d1}. We independently compare the top 5 ranked methods to the proposed one to provide a clear image of their performance in the dynamic imbalance ratio challenge. Table \ref{tab: compare-dynamic} and Figures \ref{fig:activity-d1-per}, \ref{fig:poker-d1-per}, and \ref{fig:sensor-d1-per} illustrate the results. 

Interestingly, because of the complexity that this problem brings to the non-stationary data stream classification task, we see different performances across datasets for each method. For instance, SRP outperforms the Kappa measure on certain datasets, including "CONNECT4-D1" and "CRIMES-D1", but only moderately on others. While MicFoal outperforms others in only two datasets in the Kappa measure, "OLYMPIC-D1" and "TAGS-D1", it may obtain a slightly higher Kappa score than SRP. However, due to our undersampling technique, the LSH-DynED outperforms other methods in multiple datasets and achieves the highest performance in terms of the Kappa metric, demonstrating the LSH-DynED's resilience against the dynamic imbalance ratios problem.

\subsection{\textbf{Ablation Analysis}}\label{sec: ablation study}
In this section, we evaluate the effectiveness of our proposed techniques in addressing the imbalance challenges inherent in multi-class non-stationary data streams. To this end, we selected 15 datasets from datasets presented in Table \ref{table: datasets} with varying numbers of features, instance counts, and class sizes to ensure the findings are robust and generalizable. We conducted a comparative analysis between the standard DynED model and our proposed LSH-DynED model, which incorporates multi-sliding windows and an LSH-RHP-based undersampling technique. 


As shown in Table \ref{tab: ablation}, the DynED model achieves an average mG-Mean of 49.68 across the 15 selected datasets. To put this in perspective, a low mG-Mean typically signals that a model is struggling to learn the decision boundaries for one or more minority classes. With LSH-DynED, however, that score jumps dramatically to 86.76. It is a considerable improvement—it is a fundamental shift in how the ensemble behaves, especially regarding its treatment of underrepresented classes. The key here is the LSH-RHP module, which creates balanced training batches for the ensemble’s components. By doing so, it forces the model to pay attention to and learn from those previously overlooked minority classes.

The ablation analysis also highlights a steady, though less dramatic, improvement in the Kappa score. On average, the Kappa score increases from 71.98 for DynED to 73.94 for LSH-DynED. While this gain is more modest compared to the jump in mG-Mean, it is still meaningful. What is important is that these significant improvements for minority classes do not come at the expense of the model’s overall reliability or its accuracy on majority classes.

A common concern with any undersampling method is the risk of losing valuable information \cite{nguyen2012comparative, tarawneh2022stop}. Removing too many instances from the majority class might throw out key data points that help define the decision boundary, which could hurt the model’s ability to correctly classify the majority class. The fact that the Kappa score not only remains stable but actually improves suggests that the LSH-RHP undersampling method is effective. It manages to curb the dominance of the majority class just enough to help the model learn minority classes, all while keeping the essential structure of the majority data intact. This careful balance means the model’s accuracy on the majority class does not suffer, so the overall Kappa score stays strong.

Looking at the results over time, as visualized in Figure \ref{fig: diag}, we see even more evidence of the LSH-RHP module’s impact, especially in a streaming data setting. The figure tracks the Kappa score as it evolves over time for two challenging, large-scale real-world datasets: 'Covtype' and 'Poker'. In both cases, LSH-DynED consistently outperforms DynED throughout the entire data stream. We have similar temporal observations with mG-Mean.


\subsection{Key Findings} \label{sec: overall}
The comprehensive experimental evaluation of various data stream classification models, including our proposed LSH-DynED approach, reveals several key findings.

It demonstrates superior performance across multiple datasets, as evidenced by its higher Kappa and mG-Mean scores, particularly in challenging scenarios such as highly imbalanced datasets and those exhibiting concept drift. Our evaluation includes baseline methods that employ a wide range of imbalance handling techniques, from simpler undersampling or oversampling strategies to more sophisticated adaptive approaches. The comprehensive results, particularly evident in the Kappa and mG-Mean metrics across diverse datasets (Table \ref{tab:comprehensive-metrics}), reveal that LSH-DynED is an effective approach overall in comparison and robustly addresses the multifaceted challenges of multi-class imbalance. While all methods show a negative correlation between performance and the number of classes, LSH-DynED's correlation is not statistically significant, suggesting greater resilience to increasing the number of classes.

Our approach also outperforms other methods across datasets with dynamic imbalance ratios, achieving higher scores in terms of both Kappa and mG-Mean measures. Statistical tests confirm LSH-DynED's superior ranking, with significant improvements over other methods. Ablation analysis highlights the effectiveness of the integrated multi-sliding windows and LSH-RHP-based undersampling technique.

In summary, the capability of maintaining high predictive accuracy, adaptability, resilience, and robust performance in ever-changing data streams positions LSH-DynED as a valuable tool for real-world applications.

\section{Conclusion}\label{sec: Conclusion}

This study addresses the significant challenge of multi-class classification in imbalanced non-stationary data streams by proposing a novel approach that integrates Locality Sensitive Hashing with Random Hyperplane Projections (LSH-RHP) into the Dynamic Ensemble Diversification (DynED) approach.

Our proposed approach, LSH-DynED, represents a major step forward in classifying multi-class imbalanced non-stationary data streams. Notably, this is the first time LSH-RHP has been employed for undersampling in the context of imbalanced non-stationary data streams, underscoring the novelty and significance of our contribution.


Further research could build upon our findings in several interesting ways. We envision potential explorations into adapting our approach for binary imbalanced data streams, as well as studying data streams with a high dimensionality of classes or newly emerging class labels. Additionally, a valuable contribution could be the introduction of novel datasets representing diverse data streams from various fields. Lastly, we are interested in extending our approach to multi-label streams, where each instance may belong to more than one class. These are considered interesting possibilities for future work.

Our study shows that LSH-DynED has the potential to make a significant impact in various domains with imbalanced data streams, including fraud detection, network security, and healthcare monitoring. These steps will help grow this research area and show how our proposed approach can be useful in many fields.

\bibliographystyle{acm}
\bibliography{neurips_2024}

\begin{thebibliography}{10}

\bibitem{abadifard2023dyned}
{\sc Abadifard, S., Bakhshi, S., Gheibuni, S., and Can, F.}
\newblock Dyn{ED}: Dynamic ensemble diversification in data stream classification.
\newblock In {\em Proceedings of the 32nd ACM International Conference on Information and Knowledge Management\/} (New York, NY, USA, 2023), CIKM '23, Association for Computing Machinery, p.~3707–3711.

\bibitem{hamaamin2024java}
{\sc {Abdulkareem Hamaamin, Rebin}, {Mohammed Amin Ali, Omar}, and {Wahhab Kareem, Shahab}}.
\newblock Java programming language: Time permanence comparison with other languages: A review.
\newblock {\em ITM Web Conference 64\/} (2024), 01012.

\bibitem{aguiar2023active}
{\sc Aguiar, G., and Cano, A.}
\newblock An active learning budget-based oversampling approach for partially labeled multi-class imbalanced data streams.
\newblock In {\em Proceedings of the 38th ACM/SIGAPP Symposium on Applied Computing\/} (New York, NY, USA, 2023), SAC '23, Association for Computing Machinery, p.~382–389.

\bibitem{aguiar2023survey}
{\sc Aguiar, G., Krawczyk, B., and Cano, A.}
\newblock A survey on learning from imbalanced data streams: Taxonomy, challenges, empirical study, and reproducible experimental framework.
\newblock {\em Machine Learning 113}, 7 (2023), 4165–4243.

\bibitem{al2024incremental}
{\sc Al-Shammari, A.}
\newblock An incremental ensemble diversification in data stream classification using improved hoeffding trees with thompson sampling.
\newblock {\em Journal of Al-Qadisiyah for Computer Science and Mathematics 16}, 2 (2024), Page--187.

\bibitem{10.1145/3274895.3274943}
{\sc Astefanoaei, M., Cesaretti, P., Katsikouli, P., Goswami, M., and Sarkar, R.}
\newblock Multi-resolution sketches and locality sensitive hashing for fast trajectory processing.
\newblock In {\em Proceedings of the 26th ACM SIGSPATIAL International Conference on Advances in Geographic Information Systems\/} (New York, NY, USA, 2018), SIGSPATIAL '18, Association for Computing Machinery, p.~279–288.

\bibitem{bahri2021data}
{\sc Bahri, M., Bifet, A., Gama, J., Gomes, H.~M., and Maniu, S.}
\newblock Data stream analysis: Foundations, major tasks and tools.
\newblock {\em Wiley Interdisciplinary Reviews: Data Mining and Knowledge Discovery 11}, 3 (2021), e1405.

\bibitem{bakhshi2023broad}
{\sc Bakhshi, S., Ghahramanian, P., Bonab, H., and Can, F.}
\newblock A broad ensemble learning system for drifting stream classification.
\newblock {\em IEEE Access 11\/} (2023), 89315--89330.

\bibitem{barboza2025inca}
{\sc Barboza, E.~V., de~Almeida, P. R.~L., de~Souza~Britto, A., Sabourin, R., and Cruz, R.~M.}
\newblock Inc{A}-{DES}: An incremental and adaptive dynamic ensemble selection approach using online k-d tree neighborhood search for data streams with concept drift.
\newblock {\em Information Fusion 123\/} (2025), 103272.

\bibitem{cohen2009pearson}
{\sc Benesty, J., Chen, J., Huang, Y., and Cohen, I.}
\newblock {\em Pearson correlation coefficient}.
\newblock Springer Berlin Heidelberg, Berlin, Heidelberg, 2009, pp.~1--4.

\bibitem{bernardo2021vfc}
{\sc Bernardo, A., and Della~Valle, E.}
\newblock {VFC-SMOTE}: Very fast continuous synthetic minority oversampling for evolving data streams.
\newblock {\em Data Mining and Knowledge Discovery 35}, 6 (2021), 2679--2713.

\bibitem{bernardo2020incremental}
{\sc Bernardo, A., della Valle, E., and Bifet, A.}
\newblock Incremental rebalancing learning on evolving data streams.
\newblock In {\em 2020 International Conference on Data Mining Workshops (ICDMW)\/} (2020), Institute of Electrical and Electronics Engineers (IEEE), pp.~844--850.

\bibitem{bernardo2020c}
{\sc Bernardo, A., Gomes, H.~M., Montiel, J., Pfahringer, B., Bifet, A., and Valle, E.~D.}
\newblock {C-SMOTE}: Continuous synthetic minority oversampling for evolving data streams.
\newblock In {\em 2020 IEEE International Conference on Big Data (Big Data)\/} (2020), Institute of Electrical and Electronics Engineers (IEEE), pp.~483--492.

\bibitem{bernardo2021smote}
{\sc Bernardo, A., and Valle, E.~D.}
\newblock {SMOTE-OB}: Combining smote and online bagging for continuous rebalancing of evolving data streams.
\newblock In {\em 2021 IEEE International Conference on Big Data (Big Data)\/} (2021), Institute of Electrical and Electronics Engineers (IEEE), pp.~5033--5042.

\bibitem{bifet2007learning}
{\sc Bifet, A., and Gavaldà, R.}
\newblock Learning from time-changing data with adaptive windowing.
\newblock In {\em Proceedings of the 2007 SIAM International Conference on Data Mining (SDM)\/} (2007), Society for Industrial and Applied Mathematics, pp.~443--448.

\bibitem{bifet2010leveraging}
{\sc Bifet, A., Holmes, G., and Pfahringer, B.}
\newblock Leveraging bagging for evolving data streams.
\newblock In {\em Machine Learning and Knowledge Discovery in Databases: European Conference, ECML PKDD 2010, Barcelona, Spain, September 20-24, 2010, Proceedings, Part I 21\/} (2010), Springer Berlin Heidelberg, pp.~135--150.

\bibitem{bifet2009new}
{\sc Bifet, A., Holmes, G., Pfahringer, B., Kirkby, R., and Gavald\`{a}, R.}
\newblock New ensemble methods for evolving data streams.
\newblock In {\em Proceedings of the 15th ACM SIGKDD International Conference on Knowledge Discovery and Data Mining\/} (New York, NY, USA, 2009), KDD '09, Association for Computing Machinery, p.~139–148.

\bibitem{bifet2010moa}
{\sc Bifet, A., Holmes, G., Pfahringer, B., Kranen, P., Kremer, H., Jansen, T., and Seidl, T.}
\newblock {MOA}: Massive online analysis, a framework for stream classification and clustering.
\newblock In {\em Proceedings of the First Workshop on Applications of Pattern Analysis\/} (Cumberland Lodge, Windsor, UK, 01--03 Sep 2010), T.~Diethe, N.~Cristianini, and J.~Shawe-Taylor, Eds., vol.~11 of {\em Proceedings of Machine Learning Research}, PMLR, pp.~44--50.

\bibitem{ferreira2019adaptive}
{\sc Boiko~Ferreira, L.~E., Murilo~Gomes, H., Bifet, A., and Oliveira, L.~S.}
\newblock Adaptive random forests with resampling for imbalanced data streams.
\newblock In {\em 2019 International Joint Conference on Neural Networks (IJCNN)\/} (2019), Institute of Electrical and Electronics Engineers (IEEE), pp.~1--6.

\bibitem{bonab2019less}
{\sc Bonab, H., and Can, F.}
\newblock Less is more: A comprehensive framework for the number of components of ensemble classifiers.
\newblock {\em IEEE Transactions on Neural Networks and Learning Systems 30}, 9 (2019), 2735--2745.

\bibitem{bonab2018goowe}
{\sc Bonab, H.~R., and Can, F.}
\newblock {GOOWE}: Geometrically optimum and online-weighted ensemble classifier for evolving data streams.
\newblock {\em ACM Trans. Knowl. Discov. Data 12}, 2 (jan 2018).

\bibitem{breiman2017classification}
{\sc Breiman, L.}
\newblock {\em Classification and regression trees}.
\newblock Routledge, 2017.

\bibitem{10.1016/j.neunet.2018.07.011}
{\sc Buda, M., Maki, A., and Mazurowski, M.~A.}
\newblock A systematic study of the class imbalance problem in convolutional neural networks.
\newblock {\em Neural Networks 106\/} (2018), 249--259.

\bibitem{cano2020kappa}
{\sc Cano, A., and Krawczyk, B.}
\newblock Kappa updated ensemble for drifting data stream mining.
\newblock {\em Machine Learning 109}, 1 (2020), 175--218.

\bibitem{cano2022rose}
{\sc Cano, A., and Krawczyk, B.}
\newblock {ROSE}: Robust online self-adjusting ensemble for continual learning on imbalanced drifting data streams.
\newblock {\em Machine Learning 111}, 7 (2022), 2561--2599.

\bibitem{carraher2016random}
{\sc Carraher, L.~A., Wilsey, P.~A., Moitra, A., and Dey, S.}
\newblock Random projection clustering on streaming data.
\newblock In {\em 2016 IEEE 16th International Conference on Data Mining Workshops (ICDMW)\/} (2016), Institute of Electrical and Electronics Engineers (IEEE), pp.~708--715.

\bibitem{chawla2002smote}
{\sc Chawla, N.~V., Bowyer, K.~W., Hall, L.~O., and Kegelmeyer, W.~P.}
\newblock {SMOTE}: Synthetic minority over-sampling technique.
\newblock {\em Journal of Artificial Intelligence Research 16\/} (2002), 321--357.

\bibitem{chawla2003smoteboost}
{\sc Chawla, N.~V., Lazarevic, A., Hall, L.~O., and Bowyer, K.~W.}
\newblock {SMOTEB}oost: Improving prediction of the minority class in boosting.
\newblock In {\em Knowledge Discovery in Databases: PKDD 2003: 7th European Conference on Principles and Practice of Knowledge Discovery in Databases, Cavtat-Dubrovnik, Croatia, September 22-26, 2003. Proceedings 7\/} (2003), Springer, pp.~107--119.

\bibitem{chicco2020advantages}
{\sc Chicco, D., and Jurman, G.}
\newblock The advantages of the {M}atthews correlation coefficient ({MCC}) over {F1} score and accuracy in binary classification evaluation.
\newblock {\em BMC Genomics 21\/} (2020), 1--13.

\bibitem{cieslak2008learning}
{\sc Cieslak, D.~A., and Chawla, N.~V.}
\newblock Learning decision trees for unbalanced data.
\newblock In {\em Machine Learning and Knowledge Discovery in Databases: European Conference, ECML PKDD 2008, Antwerp, Belgium, September 15-19, 2008, Proceedings, Part I 19\/} (2008), Springer Berlin Heidelberg, pp.~241--256.

\bibitem{demvsar2006statistical}
{\sc Dem\v{s}ar, J.}
\newblock Statistical comparisons of classifiers over multiple data sets.
\newblock {\em J. Mach. Learn. Res. 7\/} (Dec 2006), 1–30.

\bibitem{derrac2015keel}
{\sc Derrac, J., Garcia, S., Sanchez, L., and Herrera, F.}
\newblock Keel data-mining software tool: Data set repository, integration of algorithms and experimental analysis framework.
\newblock {\em J. Mult. Valued Logic Soft Comput 17\/} (2015), 255--287.

\bibitem{douzas2017geometric}
{\sc Douzas, G., and Bacao, F.}
\newblock Geometric {SMOTE}: Effective oversampling for imbalanced learning through a geometric extension of {SMOTE}, 2017.

\bibitem{douze2024faiss}
{\sc Douze, M., Guzhva, A., Deng, C., Johnson, J., Szilvasy, G., Mazaré, P.-E., Lomeli, M., Hosseini, L., and Jégou, H.}
\newblock The {Faiss} library.
\newblock {\em arXiv preprint arXiv:2401.08281\/} (2024).

\bibitem{espindola2005extending}
{\sc Esp{\'\i}ndola, R.~P., and Ebecken, N.~F.}
\newblock On extending f-measure and g-mean metrics to multi-class problems.
\newblock {\em WIT Transactions on Information and Communication Technologies 35\/} (2005), 25--34.

\bibitem{10.1142/9789811204746_0018}
{\sc Feldman, M.}
\newblock {\em Locality-Sensitive Hashing}.
\newblock WORLD SCIENTIFIC, 2020, ch.~Chapter 18, pp.~425--441.

\bibitem{gama2009issues}
{\sc Gama, J.~a., Sebasti\~{a}o, R., and Rodrigues, P.~P.}
\newblock Issues in evaluation of stream learning algorithms.
\newblock KDD '09, Association for Computing Machinery, p.~329–338.

\bibitem{gomes2017survey}
{\sc Gomes, H.~M., Barddal, J.~P., Enembreck, F., and Bifet, A.}
\newblock A survey on ensemble learning for data stream classification.
\newblock {\em ACM Computing Surveys 50}, 2 (March 2017).

\bibitem{gomes2017adaptive}
{\sc Gomes, H.~M., Bifet, A., Read, J., Barddal, J.~P., Enembreck, F., Pfharinger, B., Holmes, G., and Abdessalem, T.}
\newblock Adaptive random forests for evolving data stream classification.
\newblock {\em Machine Learning 106}, 9–10 (2017), 1469--1495.

\bibitem{gomes2019streaming}
{\sc Gomes, H.~M., Read, J., and Bifet, A.}
\newblock Streaming random patches for evolving data stream classification.
\newblock In {\em 2019 IEEE International Conference on Data Mining (ICDM)\/} (2019), Institute of Electrical and Electronics Engineers (IEEE), pp.~240--249.

\bibitem{gozuaccik2021concept}
{\sc G{\"o}z{\"u}a{\c{c}}{\i}k, {\"O}., and Can, F.}
\newblock Concept learning using one-class classifiers for implicit drift detection in evolving data streams.
\newblock {\em Artificial Intelligence Review 54}, 5 (2021), 3725--3747.

\bibitem{berkay2022multi}
{\sc Gulcan, E.~B., Ecevit, I.~S., and Can, F.}
\newblock Binary transformation method for multi-label stream classification.
\newblock In {\em Proceedings of the 31st ACM International Conference on Information \& Knowledge Management\/} (New York, NY, USA, 2022), CIKM '22, Association for Computing Machinery, p.~3968–3972.

\bibitem{gulowaty2019smote}
{\sc Gulowaty, B., and Ksieniewicz, P.}
\newblock Smote algorithm variations in balancing data streams.
\newblock In {\em Intelligent Data Engineering and Automated Learning -- IDEAL 2019: 20th International Conference, Manchester, UK, November 14--16, 2019, Proceedings, Part II 20\/} (2019), Springer, pp.~305--312.

\bibitem{han2005borderline}
{\sc Han, H., Wang, W.-Y., and Mao, B.-H.}
\newblock Borderline-{SMOTE}: A new over-sampling method in imbalanced data sets learning.
\newblock In {\em Advances in Intelligent Computing\/} (2005), Springer, pp.~878--887.

\bibitem{juan2021dyn}
{\sc Hidalgo, J. I.~G., Santos, S. G. T.~C., and Barros, R. S.~M.}
\newblock Dynamically adjusting diversity in ensembles for the classification of data streams with concept drift.
\newblock {\em ACM Trans. Knowl. Discov. Data 16}, 2 (July 2021).

\bibitem{Jain_2023}
{\sc Jain, A.}
\newblock Comparative analysis of java and python in machine learning: Investigate and compare the suitability and performance of java and python for machine learning tasks.
\newblock {\em International Journal of Research Publication and Reviews 4}, 12 (2023), 1151–1167.

\bibitem{jiang2019incremental}
{\sc Jiang, X., Ng, W.~W., Tian, X., Kwong, S., and Wang, H.}
\newblock Incremental hashing with undersampling.
\newblock In {\em 2019 IEEE International Conference on Systems, Man and Cybernetics (SMC)\/} (2019), Institute of Electrical and Electronics Engineers (IEEE), pp.~616--622.

\bibitem{johnson2019billion}
{\sc Johnson, J., Douze, M., and Jégou, H.}
\newblock Billion-scale similarity search with gpus.
\newblock {\em IEEE Transactions on Big Data 7}, 3 (2021), 535--547.

\bibitem{10.17485/ijst/v16i16.146}
{\sc Karthikeyan, S., and Kathirvalavakumar, T.}
\newblock A hybrid data resampling algorithm combining leader and smote for classifying the high imbalanced datasets.
\newblock {\em Indian Journal of Science and Technology 16\/} (2023), 1214--1220.

\bibitem{kelly2023uci}
{\sc Kelly, M., Longjohn, R., and Nottingham, K.}
\newblock The {UCI} {M}achine {L}earning {R}epository.
\newblock {\em URL https://archive.ics.uci.edu\/} (2023).

\bibitem{kokate2018data}
{\sc Kokate, U., Deshpande, A., Mahalle, P., and Patil, P.}
\newblock Data stream clustering techniques, applications, and models: Comparative analysis and discussion.
\newblock {\em Big Data and Cognitive Computing 2}, 4 (2018), 32.

\bibitem{korycki2020online}
{\sc Korycki, {\L}., and Krawczyk, B.}
\newblock Online oversampling for sparsely labeled imbalanced and non-stationary data streams.
\newblock In {\em 2020 International Joint Conference on Neural Networks (IJCNN)\/} (2020), Institute of Electrical and Electronics Engineers (IEEE), pp.~1--8.

\bibitem{kubat1997addressing}
{\sc Kub{\'a}t, M., and Matwin, S.}
\newblock Addressing the curse of imbalanced training sets: One-sided selection.
\newblock In {\em International Conference on Machine Learning\/} (1997), vol.~97, Citeseer, p.~179.

\bibitem{kuncheva2003measures}
{\sc Kuncheva, L.~I., and Whitaker, C.~J.}
\newblock Measures of diversity in classifier ensembles and their relationship with the ensemble accuracy.
\newblock {\em Machine Learning 51}, 2 (2003), 181.

\bibitem{laurikkala2001improving}
{\sc Laurikkala, J.}
\newblock Improving identification of difficult small classes by balancing class distribution.
\newblock In {\em Artificial Intelligence in Medicine: 8th Conference on Artificial Intelligence in Medicine in Europe, AIME 2001 Cascais, Portugal, July 1--4, 2001, Proceedings 8\/} (2001), Springer, pp.~63--66.

\bibitem{10.4028/www.scientific.net/amm.321-324.804}
{\sc Lee, K.~M.}
\newblock Locality sensitive hashing with extended partitioning boundaries.
\newblock In {\em Mechatronics and Industrial Informatics\/} (7 2013), vol.~321 of {\em Applied Mechanics and Materials}, Trans Tech Publications Ltd, pp.~804--807.

\bibitem{li2020incremental}
{\sc Li, Z., Huang, W., Xiong, Y., Ren, S., and Zhu, T.}
\newblock Incremental learning imbalanced data streams with concept drift: The dynamic updated ensemble algorithm.
\newblock {\em Knowledge-Based Systems 195\/} (2020), 105694.

\bibitem{lipska2022influence}
{\sc Lipska, A., and Stefanowski, J.}
\newblock The influence of multiple classes on learning from imbalanced data streams.
\newblock In {\em Proceedings of the Fourth International Workshop on Learning with Imbalanced Domains: Theory and Applications\/} (2022), vol.~183 of {\em Proceedings of Machine Learning Research}, PMLR, pp.~187--198.

\bibitem{liu2021comprehensiveF}
{\sc Liu, W., Zhang, H., Ding, Z., Liu, Q., and Zhu, C.}
\newblock A comprehensive active learning method for multiclass imbalanced data streams with concept drift.
\newblock {\em Knowledge-Based Systems 215\/} (2021), 106778.

\bibitem{liu2023multiclass}
{\sc Liu, W., Zhu, C., Ding, Z., Zhang, H., and Liu, Q.}
\newblock Multiclass imbalanced and concept drift network traffic classification framework based on online active learning.
\newblock {\em Engineering Applications of Artificial Intelligence 117\/} (2023), 105607.

\bibitem{liu2016multilinear}
{\sc Liu, X., Fan, X., Deng, C., Li, Z., Su, H., and Tao, D.}
\newblock Multilinear hyperplane hashing, 2016.

\bibitem{loezer2020cost}
{\sc Loezer, L., Enembreck, F., Barddal, J.~P., and de~Souza~Britto, A.}
\newblock Cost-sensitive learning for imbalanced data streams.
\newblock In {\em Proceedings of the 35th Annual ACM Symposium on Applied Computing\/} (New York, NY, USA, 2020), SAC '20, Association for Computing Machinery, p.~498–504.

\bibitem{lu2019adaptive}
{\sc Lu, Y., Cheung, Y.-M., and Yan~Tang, Y.}
\newblock Adaptive chunk-based dynamic weighted majority for imbalanced data streams with concept drift.
\newblock {\em IEEE Transactions on Neural Networks and Learning Systems 31}, 8 (2020), 2764--2778.

\bibitem{luo2023multiclass}
{\sc Luo, X., Li, D., Zhang, H., Xu, L., Cai, B., and Deng, J.}
\newblock Multi-classification data stream algorithm based on one-vs-rest strategy.
\newblock In {\em Proceedings of the 2023 3rd International Conference on Artificial Intelligence, Automation and Algorithms\/} (New York, NY, USA, 2023), AI2A '23, Association for Computing Machinery, p.~66–72.

\bibitem{lyon2014hellinger}
{\sc Lyon, R., Brooke, J., Knowles, J., and Stappers, B.}
\newblock Hellinger distance trees for imbalanced streams.
\newblock In {\em 2014 22nd International Conference on Pattern Recognition\/} (2014), Institute of Electrical and Electronics Engineers (IEEE), pp.~1969--1974.

\bibitem{montiel2021river}
{\sc Montiel, J., Halford, M., Mastelini, S.~M., Bolmier, G., Sourty, R., Vaysse, R., Zouitine, A., Gomes, H.~M., Read, J., Abdessalem, T., and Bifet, A.}
\newblock River: Machine learning for streaming data in python.
\newblock {\em Journal of Machine Learning Research 22}, 110 (2021), 1--8.

\bibitem{wing2022hashing}
{\sc Ng, W. W.~Y., Xu, S., Zhang, J., Tian, X., Rong, T., and Kwong, S.}
\newblock Hashing-based undersampling ensemble for imbalanced pattern classification problems.
\newblock {\em IEEE Transactions on Cybernetics 52}, 2 (2022), 1269--1279.

\bibitem{nguyen2012comparative}
{\sc Nguyen, H.~M., Cooper, E.~W., and Kamei, K.}
\newblock A comparative study on sampling techniques for handling class imbalance in streaming data.
\newblock In {\em The 6th International Conference on Soft Computing and Intelligent Systems, and The 13th International Symposium on Advanced Intelligence Systems\/} (2012), Institute of Electrical and Electronics Engineers (IEEE), pp.~1762--1767.

\bibitem{Palli_Jaafar_Gilal_Alsughayyir_Gomes_Alshanqiti_Omar_2024}
{\sc Palli, A.~S., Jaafar, J., Gilal, A.~R., Alsughayyir, A., Gomes, H.~M., Alshanqiti, A., and Omar, M.}
\newblock Online machine learning from non-stationary data streams in the presence of concept drift and class imbalance: A systematic review.
\newblock {\em Journal of Information and Communication Technology 23}, 1 (Jan. 2024), 105–139.

\bibitem{pinto2023imbalance}
{\sc Pinto, J. G.~M.}
\newblock {\em Imbalanced multiclass classification with concept drift}.
\newblock PhD thesis, Universidade do Porto, 2023.
\newblock Copyright - Database copyright ProQuest LLC; ProQuest does not claim copyright in the individual underlying works; Last updated - 2024-12-11.

\bibitem{jinjie2025noise}
{\sc Qiu, J., Zhuo, S., Yu, P.~S., Wang, C., and Huang, S.}
\newblock Online learning for noisy labeled streams.
\newblock {\em ACM Trans. Knowl. Discov. Data\/} (May 2025).
\newblock Just Accepted.

\bibitem{quinlan1986induction}
{\sc Quinlan, J.~R.}
\newblock Induction of decision trees.
\newblock {\em Machine Learning 1\/} (1986), 81--106.

\bibitem{rao1995review}
{\sc Rao, C.~R.}
\newblock A review of canonical coordinates and an alternative to correspondence analysis using hellinger distance.
\newblock {\em Q{\"u}estii{\'o}: Quaderns d'estad{\'\i}stica i investigaci{\'o} operativa\/} (1995).

\bibitem{read2011streaming}
{\sc Read, J., Bifet, A., Holmes, G., and Pfahringer, B.}
\newblock Streaming multi-label classification.
\newblock In {\em Proceedings of the Second Workshop on Applications of Pattern Analysis\/} (CIEM, Castro Urdiales, Spain, 19--21 Oct 2011), T.~Diethe, J.~Balcazar, J.~Shawe-Taylor, and C.~Tirnauca, Eds., vol.~17 of {\em Proceedings of Machine Learning Research}, PMLR, pp.~19--25.

\bibitem{read2023learning}
{\sc Read, J., and Zliobaite, I.}
\newblock Learning from data streams: An overview and update.
\newblock {\em Available at SSRN 4326595\/} (2023).

\bibitem{santhiappan2018novel}
{\sc Santhiappan, S., Chelladurai, J., and Ravindran, B.}
\newblock A novel topic modeling based weighting framework for class imbalance learning.
\newblock In {\em Proceedings of the ACM India Joint International Conference on Data Science and Management of Data\/} (New York, NY, USA, 2018), CODS-COMAD '18, Association for Computing Machinery, p.~20–29.

\bibitem{tarawneh2022stop}
{\sc Tarawneh, A.~S., Hassanat, A.~B., Altarawneh, G.~A., and Almuhaimeed, A.}
\newblock Stop oversampling for class imbalance learning: A review.
\newblock {\em IEEE Access 10\/} (2022), 47643--47660.

\bibitem{tomek1976experiment}
{\sc Tomek, I.}
\newblock An experiment with the edited nearest-neighbor rule.
\newblock {\em IEEE Transactions on Systems, Man, and Cybernetics SMC-6}, 6 (1976), 448--452.

\bibitem{tomek1976two}
{\sc Tomek, I.}
\newblock Two modifications of {CNN}.
\newblock {\em IEEE Transactions on Systems, Man, and Cybernetics SMC-6}, 11 (1976), 769--772.

\bibitem{TSYMBAL200583}
{\sc Tsymbal, A., Pechenizkiy, M., and Cunningham, P.}
\newblock Diversity in search strategies for ensemble feature selection.
\newblock {\em Information Fusion 6}, 1 (2005), 83--98.
\newblock Diversity in Multiple Classifier Systems.

\bibitem{vardi2020efficiency}
{\sc Vardi, M.~Y.}
\newblock Efficiency vs. resilience: What {COVID}-19 teaches computing, April 2020.

\bibitem{wang2012resilience}
{\sc Wang, C.-h., and Pham, L.}
\newblock {\em Resilience and robustness}.
\newblock Engineers Australia Barton, ACT, Barton, A.C.T., 2012, pp.~114--121.

\bibitem{10.1109/cec.2014.6900653}
{\sc Wang, F., Gao, Y., and Zhu, Z.}
\newblock Locality-sensitive hashing based multiobjective memetic algorithm for dynamic pickup and delivery problems.
\newblock In {\em 2014 IEEE Congress on Evolutionary Computation (CEC)\/} (2014), Institute of Electrical and Electronics Engineers (IEEE), pp.~661--666.

\bibitem{10.4028/www.scientific.net/amm.556-562.3804}
{\sc Wang, P., Yin, D., and Sun, T.}
\newblock Bi-level locality sensitive hashing index based on clustering.
\newblock In {\em Mechatronics Engineering, Computing and Information Technology\/} (7 2014), vol.~556 of {\em Applied Mechanics and Materials}, Trans Tech Publications Ltd, pp.~3804--3808.

\bibitem{wang2020auc}
{\sc Wang, S., and Minku, L.~L.}
\newblock Auc estimation and concept drift detection for imbalanced data streams with multiple classes.
\newblock In {\em 2020 International Joint Conference on Neural Networks (IJCNN)\/} (2020), Institute of Electrical and Electronics Engineers (IEEE), pp.~1--8.

\bibitem{Wang_Minku_Yao_2015}
{\sc Wang, S., Minku, L.~L., and Yao, X.}
\newblock Resampling-based ensemble methods for online class imbalance learning.
\newblock {\em IEEE Transactions on Knowledge and Data Engineering 27}, 5 (May 2015), 1356–1368.

\bibitem{wang2016dealing}
{\sc Wang, S., Minku, L.~L., and Yao, X.}
\newblock Dealing with multiple classes in online class imbalance learning.
\newblock In {\em Proceedings of the Twenty-Fifth International Joint Conference on Artificial Intelligence\/} (2016), IJCAI'16, AAAI Press, p.~2118–2124.

\bibitem{wang2018systematic}
{\sc Wang, S., Minku, L.~L., and Yao, X.}
\newblock A systematic study of online class imbalance learning with concept drift.
\newblock {\em IEEE Transactions on Neural Networks and Learning Systems 29}, 10 (2018), 4802--4821.

\bibitem{yen2009cluster}
{\sc Yen, S.-J., and Lee, Y.-S.}
\newblock Cluster-based under-sampling approaches for imbalanced data distributions.
\newblock {\em Expert Systems with Applications 36}, 3, Part 1 (2009), 5718--5727.

\bibitem{zhu2017synthetic}
{\sc Zhu, T., Lin, Y., and Liu, Y.}
\newblock Synthetic minority oversampling technique for multiclass imbalance problems.
\newblock {\em Pattern Recognition 72\/} (2017), 327--340.

\end{thebibliography}
\end{document}